\documentclass[review=false]{acmart}
\usepackage{amsmath}
\usepackage{amsfonts}
\usepackage{multirow}
\usepackage{tikz}
\usepackage{longtable}
\usepackage{tikz-qtree}
\usepackage{geometry}
\usepackage[T1]{fontenc}
\usepackage[edges]{forest}
\usetikzlibrary{trees} 
\usepackage[figuresright]{rotating}
\AtBeginDocument{%
  \providecommand\BibTeX{{%
    \normalfont B\kern-0.5em{\scshape i\kern-0.25em b}\kern-0.8em\TeX}}}

\setcopyright{acmcopyright}
\copyrightyear{2018}
\acmYear{2022}
\acmDOI{XXXXXXX.XXXXXXX}

\begin{document}
\title{Non-Imaging Medical Data Synthesis for Trustworthy AI: A Comprehensive Survey}

\author{Xiaodan Xing}
\authornote{These authors contributed equally to this research.}
\authornote{Send correspondence to x.xing@imperial.ac.uk and g.yang@imperial.ac.uk}
\email{x.xing@imperial.ac.uk}
\orcid{0000-0002-2468-9266}
\author{Huanjun Wu}
\authornotemark[1]
\email{hw1019@imperial.ac.uk}
\author{Lichao Wang}
\authornotemark[1]
\email{lw422@imperial.ac.uk}
\affiliation{%
  \institution{Imperial College London}
  \city{London}
  \country{UK}
}

\author{Iain Stenson}
\email{istenson@turing.ac.uk}
\author{May Yong}
\email{myong@turing.ac.uk}
\affiliation{%
  \institution{Alan Turing Institute}
  \city{London}
  \country{UK}}
\author{Javier Del Ser}
\email{javier.delser@tecnalia.com}
\affiliation{%
  \institution{TECNALIA, Basque Research \& Technology Alliance (BRTA)}
  \city{Derio}
  \country{Spain}}
\author{Simon Walsh}
\authornote{Co-last senior authors}
\email{s.walsh@imperial.ac.uk}
\author{Guang Yang}
\authornotemark[2]
\authornotemark[3]
\affiliation{%
  \institution{Imperial College London}
  \city{London}
  \country{UK}}
\email{g.yang@imperial.ac.uk}

\renewcommand{\shortauthors}{Xing, et al.}

\begin{abstract}
Data quality is the key factor for the development of trustworthy AI in healthcare. A large volume of curated datasets with controlled confounding factors can help improve the accuracy, robustness and privacy of downstream AI algorithms. However, access to good quality datasets is limited by the technical difficulty of data acquisition and large-scale sharing of healthcare data is hindered by strict ethical restrictions. Data synthesis algorithms, which generate data with a similar distribution as real clinical data, can serve as a potential solution to address the scarcity of good quality data during the development of trustworthy AI. However, state-of-the-art data synthesis algorithms, especially deep learning algorithms, focus more on imaging data while neglecting the synthesis of non-imaging healthcare data, including clinical measurements, medical signals and waveforms, and electronic healthcare records (EHRs). Thus, in this paper, we will review the synthesis algorithms, particularly for non-imaging medical data, with the aim of providing trustworthy AI in this domain. This tutorial-styled review paper will provide comprehensive descriptions of non-imaging medical data synthesis on aspects including algorithms, evaluations, limitations and future research directions. 
\end{abstract}



\maketitle

\section{Introduction}

The use of Artificial Intelligence (AI) on health data is creating promising tools for assisting clinicians in application fields such as automatic evaluation of diseases and prognosis management \cite{healthcareAI}. However, the AI algorithms can be biased, unfair, or unethical with a high risk of privacy breach \cite{trustworthyAI}. These AI algorithms, failing to win the trust of human \cite{humanandAI}, hinder the development and large-scale applications in healthcare scenarios. Over the past decades, researchers have been working on developing {\itshape trustworthy AI} \cite{trustworthyAI,trustworthyAI2} by improving robustness, variety, and transparency through the AI life cycle, where the training data of AI algorithms \cite{trustworthyAI2} is identified as a key factor for trustworthy AI.

A trustworthy training data should have a set of (overlapping) properties: 1) having a large number; 2) having an unbiased variety; 3) being collected under strict ethical regulations \cite{voigt2017gdpr,hipaa} and 4) being used with a low risk of the privacy breach. Considering the complex procedure and strict protocols of medical data acquisition, practitioners face enormous difficulty acquiring a large quantity of high-quality medical data in the real-world. In addition to the data acquisition challenges, healthcare data is sensitive, sharing/working with which can easily lead to a violation of patients' privacies. 

To address these problems, data synthesis algorithms have been developed. By synthesizing medical data instead of acquiring data from the real world, researchers can improve the size and variety of training datasets, impute the missing values, and protect patients' privacy. These synthetic data can serve as a qualified training set for the trustworthy AI algorithms \cite{liang2022advances,nightingale2022ai}. Conventional data synthesis algorithms are based on sampling, i.e., modeling the data distribution and generating data by sampling from this distribution. However, this statistical sampling requires prior information on the distributional functions. Recently, deep learning models were developed in data synthesis because they do not rely on an explicit selection of distributional functions, and their performance has been widely validated, especially on medical images \cite{gan2014,gainwithtricks}. These algorithms can generate different types of synthetic medical data, which can be used in various AI algorithms \cite{yang2021gan}. 

Synthetic medical data has two forms: imaging data and non-imaging data. Medical imaging data, including X-ray, CT, and MR images, are a major type of healthcare data because of their lesion observation and quantification ability. However, medical images cannot serve as a single standard for diagnosis and prognosis. Analyses of non-imaging medical data are indispensable and beneficial in large-scale in silico clinical trials. Non-imaging medical data varies in acquisition methods, data structures, and analysis algorithms. One of the most commonly-used non-imaging medical data is electronic healthcare record (EHR) \cite{EHRofficial1,EHRofficial2}, which contains patients' demographics, symptoms, medications, and disease histories. Laboratory test results, sometimes included in each patient's EHR, are another essential subtype of non-imaging medical data. Blood tests, for example, play a decisive role in diagnosing diabetes and cancer \cite{bloodtestsNHS}; and automatic diagnosis models can be developed on the blood test results \cite{gunvcar2018application,podnar2019diagnosing}. Even though imaging and non-imaging data are both crucial in clinical settings, we discovered that \textbf{non-imaging data had rarely been investigated} in data synthesis algorithms and computer-aided medical applications in recent decades. In addition, we also discovered that the development of synthesis algorithms is overwhelmed by deep learning models, and detailed \textbf{explanations of non-deep learning methods and their relations are not well-addressed} in the literature.    

To address these research gaps and to provide entry-level guidance toward non-imaging medical data synthesis in trustworthy AI, this paper provides a comprehensive survey on the algorithms, evaluation metrics, and related datasets on non-imaging medical data synthesis in technical aspects. During the literature review, we discovered three review papers investigating the synthesis algorithms of non-imaging medical data. Wang et al. \cite{ehrReview1}, and Goncalves et al. \cite{ehrReview1} performed comparative studies of EHR synthesis on a number of algorithms. Hernandez et al. \cite{ehrReview3} provided a comprehensive literature review about non-imaging data synthesis algorithms. Still, these reviews did not include the mathematical foundations of these data synthesis algorithms nor the relationships among these data synthesis algorithms. In this review, we will provide explanations and mathematical definitions of non-imaging medical data synthesis algorithms to provide a tutorial-style survey for entry-level researchers and a complete storyline for developing synthesis algorithms. We will also discuss current limitations and dilemmas to provide future research directions in this field. The literature related to non-imaging healthcare data synthesis from 2000 to 2022 is reviewed, systematically categorizing it by the methods in use and applications. We also investigate open-source datasets and toolkits for synthetic data generation. The contributions of this survey are four-fold:
\begin{itemize}
\item To provide a comprehensive, up-to-date survey about non-imaging medical data synthesis algorithms;
\item To highlight the limitations and research issues regarding non-imaging medical data synthesis and to provide guidance for the development of trustworthy AI;
\item To provide a clear definition and description of non-imaging medical data, as well as the open-source datasets and pre-processing methods;
\item To introduce the mathematical and statistical foundation of non-imaging data synthesis and to provide a tutorial for the non-imaging synthesis algorithms.
\end{itemize}

\textcolor{black}{The rest of this review paper is organized as follows: first, Section \ref{sec:literature} poses the criteria adopted to collect and filter the literature reviewed in the survey, and presents the taxonomy used to organize it in a systematic and coherent fashion. In the following Sections \ref{sec:simulation}, \ref{sec:statistics} and \ref{sec:gan}, we will introduce three major types of data generation algorithms. In section \ref{sec:metrics}, we will introduce three major aspects of trustworthy synthetic data quality and corresponding measurements of these qualities. In section \ref{sec:datasets}, we will introduce several datasets according to their data types, and we will also explain commonly practiced pre-processing procedures of these data types. In section \ref{sec:discussion}, we will discuss the limitations of non-imaging medical data synthesis and provide potential research directions for non-imaging data synthesis with trustworthy AI considerations.}

\section{Literature collection and taxonomy} \label{sec:literature}
\setlength\rotFPtop{0pt plus 1fil}
\begin{figure}
\centering
\tikzset{
    my node/.style={
        font=\small,
        rectangle,
        draw=#1!75,
        align=justify,
    }
}
\forestset{
    my tree style/.style={
        for tree={grow=east,
            parent anchor=east, 
            child anchor=west,  
        where level=0{my node=black,text width=2em}{},
        where level=1{my node=black,text width=8em}{},
        where level=2{my node=black,text width=15em}{},
       where level=3{my node=black,text width=12em}{},
            l sep=1.5em,
            forked edge,                
            fork sep=1em,               
            edge={draw=black!50, thick},                
            if n children=3{for children={
                    if n=2{calign with current}{}}
            }{},
            tier/.option=level,
        }
    }
}
    \begin{forest}
      my tree style
[All
    [Review papers
            [Data synthesis review \cite{ehrReview1,ehrReview2,ehrReview3}]
            [SMOTE review \cite{SmoteReview}]
            [fMRI synthesis review \cite{fmrisynthesisreview}]
        ]
        [Simulation based methods (14)
                [Medical signal simulation \cite{erhardt2012simtb,afshin2011misimulation,barzegaran2019eegsourcesim,smith2013identifying,abowd2004new,lindquist2007modeling}]
                [Sequential events simulation
                    [EMERGE \cite{emerge2009,emerge2010}]
                    [Knowledge-driven methods \cite{EMRBot,knowledge2013,knowledge2016}]
                    [PADARSER \cite{PADARSER,walonoski2018synthea}]
                    [CorMESR\cite{CorMESR}]
                ]
        ] 
                [Statistical modeling (24)
            [Single- and multi-variate distributions
                    [Single distributions \cite{cocoa}]
                    [Multi distributions \cite{machanavajjhala2008privacy}  ]
                    [Copula functions \cite{jeong2016copula,dpcopula,li2014dpsynthesizer,patki2016syntheticdatavault} ]
            ]
            [SMOTE \cite{Smote,smotepd} 
                    [Borderline-SMOTE \cite{Borderline-SMOTE} ]
                    [Sequential SMOTE \cite{INOS}  ]
                    [Adaptive SMOTE \cite{ADASYN,adasynad} ]]
            [Multiple imputation  
                    [GADP \cite{GADP} ]
                    [IPSO \cite{IPSO,caiola2010random,fuzzyMI,MISVM}  ]
                    [MICE \cite{goncalves2020generation} ]
                    [Multiple imputation with differential privacy \cite{PGS2013,PGSapplication} ]
                    ]
            [Conditional distributions with (PGM)  
                    [PGM \cite{sun2015bayesian,deeva2020bayesian,luo2021oversampling,tucker2020generating} ]
                    [PGM with differential privacy \cite{Privbayes,ping2017datasynthesizer} ]
            ]
        ]
        [	Deep learning based methods (34)
            [Tabular data 
                    [Half synthesis networks \cite{imputationAE2017,gain,gainwithtricks,improvedgain} 
                    ]
                    [Fully synthesis networks \cite{AE-ELM,OVAE,medGAN,ehrGAN,EMRGAN,heterogeneousGAN,dash2019healthGAN,rashidian2020smooth,TGAN,CTGAN,SPRINTGAN,table-gan,bcgan, obgan,SMOGAN,SAGAN,cwganbasedoversampling} 
                    ]
            ]
            [Sequential data \cite{eva,wang2019scGAN,timegan,DAAE,rcgan,longgan,SynTEG,CorGAN} 
            ]
            [Deep neural networks with differential privacy \cite{dpgan,pategan,chen2018dpvae} 
            ]
          [Deep neural networks with fairness \cite{fairgan,fairganP} 
            ]
        ] 
]
    \end{forest}
\caption{All papers included in this review. These papers are categorized by their synthesis algorithms.}
\label{fig:overall}
\end{figure}
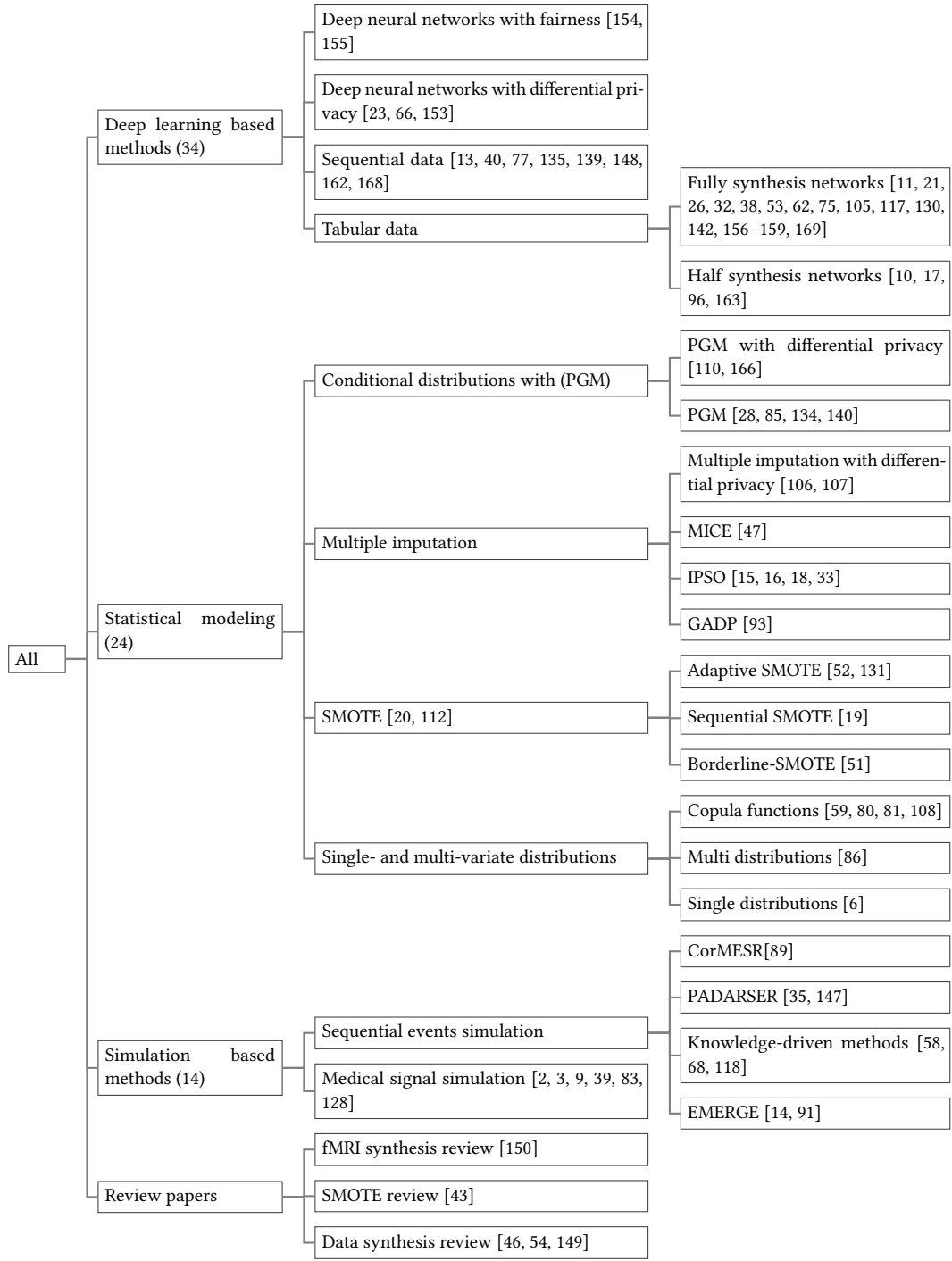
All papers included in this review were obtained by a three-stage searching strategy. 

First, we selected the papers regarding non-imaging healthcare data synthesis from January 1st, 2000 to July 1st, 2022 with the keywords “data synthesis”, “synthetic data”, “data generation”,  “data augmentation”, and “oversampling”. They were concatenated in a “or” logic relation. We confined our search to computer sciences area and deleted the papers that are not related according to their abstracts. \textcolor{black}{We focus on two types of non-imaging medical data during our searching process: tabular data and sequential data. Other non-imaging data medical types, although provide crucial information in healthcare analysis. are not discussed. It is because 1) the synthesis and applications of these data types are scarce, i.e., social networks for infectious \cite{medicalNetwork} or family inherited diseases; or 2) the synthetic data does not provide new information for downstream tasks, e.g., medical reports \cite{medicalReport}}. At this stage, we used Scopus (\url{https://www.scopus.com}) as our search engine because it is the largest database for peer-reviewed literature. After the first screening, 988 papers were selected. 

During the first stage of the search for references, we did not especially consider healthcare data synthesis algorithms. For example, we included SMOTE \cite{Smote}, which proposes an algorithm from an application-agnostic perspective; hence no specific use cases are identified. Thus, after the coarse search based on abstracts, we further read the papers and deleted those papers that focus on non-healthcare applications and do not propose novel algorithms, leaving 67 papers. 

In the last stage of the literature collection process, we re-read the papers and summarized their methods and applications. We then checked the reference lists of these papers to complete the paper searching. In stage 1 and stage 2, papers in {\itshape Arxiv} (\url{https://arxiv.org/}) are not included because they are not peer-reviewed. However, in the reference stage, highly-cited {\itshape Arxiv} papers (citation >20) are included. After the last stage, 77 papers were selected and analyzed, shown in Fig. \ref{fig:overall}.

\section{Simulation-based algorithms}\label{sec:simulation}
In this section, we will introduce simulation-based algorithms for data synthesis. We identified two sets of simulation-based algorithms according to their target non-imaging data types. Simulation-based methods generate synthetic data by trying to simulate underlying real-world mechanisms. 

\subsection{Medical signal simulation}
Simulation-based algorithms have been widely used, particularly for generating medical signals, such as MEG, EEG, and fMRI \cite{fmrisynthesisreview}. For medical signal simulation, simulation-based methods use the summation of three basic components of the signals: a baseline signal, a signal of interest (\textcolor{black}{in fMRI simulation, this signal of interest is the BOLD signal; in EEG simulation, the signal of interest is the electrical signals produced by the brain.}) and noises, and the final simulated signals are the summation of these three basic components. 

\textcolor{black}{The baseline signal represents the basic numerical level of simulated signals, and the values of baselines can be tissue-specific \cite{erhardt2012simtb}; while some algorithms simply set the baseline value to zero \cite{afshin2011misimulation}. For the synthesis of signals of interest, most medical signal methods considered the correlations among signals from different brain regions. Multivariate autoregressive modeling (MAR) \cite{barzegaran2019eegsourcesim,smith2013identifying} provides spatial-related correlations toward medical signal simulation. The noises, however, can be modeled by Gaussian distributions or mixture Gaussian distributions \cite{lindquist2007modeling}, and motion noise can also be added to simulate the patient motion during scanning. }


\subsection{EHR simulation}\label{subsec:ehrsimulation}
The most well-known EHR simulation algorithm is the Synthetic Electronic Medical Records Generator (EMERGE) project \cite{emerge2009,emerge2010}. They synthesized time series of events, which were addressed as "patient care models", "care flow" or "caremaps", for both background populations and populations with a specific kind of disease. To be specific, the patient care models contain a series of care-related tasks in managing a patient trajectory, and provide a workflow guidance for patients with specific diseases \cite{gooch2011careflow}. 

EMERGE started by synthesizing basic demographic information and symptoms reported for the first visits, and then a series of timestamps were synthesized for each synthetic patient. For each time stamp from each synthetic patient, EMERGE selected the closest health care record from the real dataset according to weighted Euclidean distance and Jaccard distance. Finally, a human expert was invited to modify the care models for each synthetic patient.

Here, we would like to elaborate more on the timestamp synthesis for the EMERGE project. It is worth noting that the timestamp synthesis did not directly synthesize time series with dates and symptoms for each patient; instead, it synthesized the population frequency information. This population-frequency strategy has been widely applied to EHR simulation to simulate time stamps. Let's consider a specific example and select patients that first visit the hospital for Viral Enteritis (ICD 008), and their visits were all before Jan. 1st. The EMERGE first obtained the number of patients $n$ who re-visit the hospital on Jan. 2nd. Then for Jan. 2nd, $n$ visiting records were generated, and were randomly assigned to the patient population.   

\begin{figure}[h]
  \centering
  \includegraphics[width=\linewidth]{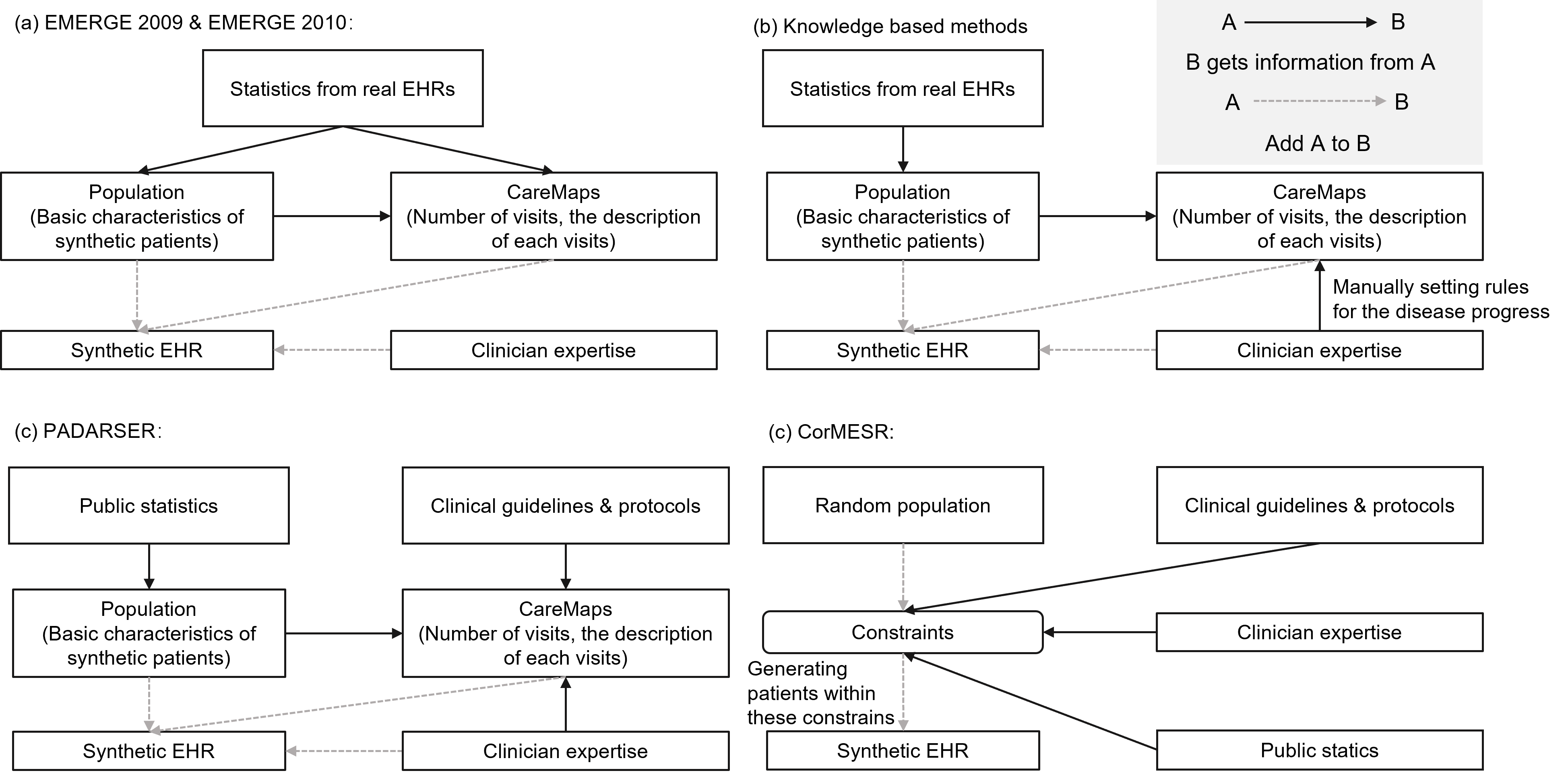}
  \caption{A detailed development of the care model synthesis.}

  \label{fig:simulation}
\end{figure}

The heavily data-driven nature of EMERGE has been criticized for the risk of patient re-identifying \cite{CriticEmerge}. Thus, knowledge-based algorithms were then proposed \cite{EMRBot,knowledge2013,knowledge2016}. The PADARSER \cite{PADARSER} and CorMESR methods \cite{CorMESR} combined expert knowledge and data-driven approaches by obtaining information from public statistics, Clinical Practice Guidelines (CPG), and Health Incidence Statistics (HIS). A toolkit named Synthea \cite{walonoski2018synthea} provided a well-engineered implementation of PADARSER. Detailed development of the care model synthesis can be found in Fig. \ref{fig:simulation}. 

Unlike methods in the following sections, section \ref{sec:statistics} and section \ref{sec:gan}, simulation-based algorithms do not necessarily require a real-world dataset for reference. The unnecessity of reference real-world datasets lowers the risk of a potential privacy breach. However, these algorithms require expert knowledge during the simulation of the data generation mechanisms, thus increasing the human workload to a great extent. In the following section, we will introduce statistical modeling-based algorithms that do not acquire large-scale manual guidance.

\section{Statistical modeling}\label{sec:statistics}
This section will introduce data synthesis algorithms, as in Table \ref{tab:single} that use statistical modeling and sampling strategies. A common ground of these algorithms is that they approximate the distributions of attributes and synthesize the data by sampling. The distributions of attributes can be an independent distribution (section \ref{subsec:multi-variate}), joint attributions of all attributes (include Copula functions in section \ref{subsec:multi-variate} and SMOTE in section \ref{subsec:smote}), conditional distributions on selected attributes (section \ref{subsec:multipleimputation}), or conditioned on attribute relations (section \ref{subsec:bayesianNetworks}). \textcolor{black}{It should be noted that the EHR simulation methods mentioned in section \ref{subsec:ehrsimulation} may use statistical modeling in the synthesis pipeline, but the performances of those methods rely more on prior knowledge, such as the population information and the treatment workflow of diseases, not sophisticated modeling of attribute distributions. }

We will denote the motivation as synthesized data $Y =\{y_1^T, y_2^T,...,y_N^T|y_i^T \in R^M, i\in [1,2,...,N]\}$ from real data $X=\{x_1^T, x_2^T,...,x_N^T|x_i \in R^M, i\in [1,2,...,N]\}$. Here $x_i^T$ and $y_i^T$ are attributes for both datasets. For tabular data, the indexing of all $N$ attributes is permutable. For a series of events, the indexing follows a chronological order, where $i$ indicates the $i$-th events.

\subsection{Sampling from single- and multi-variate distributions}\label{subsec:multi-variate}
The simplest method for data generation is to generate each variable independently from the corresponding pre-defined distributions. These distributions can be denoted as $Pr(X_i;\theta)$, where $\theta$ is the parameter estimated from the occurrence of values in the real-world datasets, and $X_i$ is the $i$-th attribute.

\textcolor{black}{This independent attribute distribution modeling has been widely used in many data-driven clinical applications, such as the EMERGE project \cite{emerge2009,emerge2010} we mentioned before. Gaussian distributions are used for continuous variables, and for discrete variables, Binomial distributions are used. However, the selections of distributions can be more varied. } COCOA \cite{cocoa} is a framework for generating relational tables. and it has 11 common data distributions, including normal, beta, chi, chi-square, exponential (exp), gamma, geometric, logarithmic (log), Poisson, t-student (Tstu), and uniform (uni) \cite{distributions}. 


The major limitation of this independent variable synthesis is that the intrinsic pattern between variables is discarded during training, leading to unmatched variables for synthetic populations. Thus, multivariate distributions of all attributes were proposed \cite{machanavajjhala2008privacy}. However, the multivariate distributions rely heavily on the type of distributions chosen, and for high-dimensional datasets, the computation efficiency is low, and the multivariate distributions might be sparse. 

Copula functions provide a way to model the correlations between features, as well as avoid intensive parameter searching. For a two dimensional dataset $X=\{x_1^T,x_2^T\}$, and their marginal cumulative distribution function (CDF) of each attribute $F_1(i)=Pr(x_1^T\leq i)$ and $F_2(i)=Pr(x_2^T\leq i)$. There exists a two-dimension Copula function $C$ such that $F(i_1,i_2) = C(F_1(i_1),F_2(i_2))$\cite{sklar1959fonctions}. Thus, the multivariate distribution modeling can then be simplified into Copula function estimating and marginal computation. The DPCopula \cite{dpcopula} and its extension DPSynthesizer \cite{li2014dpsynthesizer} used Gaussian copula functions, which further disentangled the multivariate Gaussian distributions into the product of the Gaussian dependence and margins. 

The Copula functions have been proved to be efficient in population synthesis \cite{jeong2016copula}, i.e., generating basic demographics for the target population. An open-sourced Copula modeling implementation can be found in \cite{patki2016syntheticdatavault}. However, since the Copula is defined on the CDF, the Copula function-based modeling can only be applied to continuous variables.  

\begin{table}[]
\caption{Papers using single- and multi-variate distribution sampling. *Although these papers did not report synthesis performance on medical data, they are open-sourced and easily implemented. }
\centering
\begin{tabular}{llll}
\toprule
Paper   reference & Year & Distributions & Medical data applications \\ \midrule
\cite{machanavajjhala2008privacy} & 2008 & Multinomial sampling with a dirichlet prior & Demongraphics (Census data) \\ \midrule
DPCopula \cite{dpcopula} & 2014 & Copula functions with differential privacy &  \\ \midrule
DPSynthesizer \cite{li2014dpsynthesizer} & 2014 & Copula functions with differential privacy & Demongraphics (Census data) \\ \midrule
COCOA \cite{cocoa} & 2016 & 11 common data distributions & NaN* \\ \midrule
\cite{jeong2016copula} & 2016 & Copula functions & Hospital emergency population \\ \midrule
SyntheticDataVault   \cite{patki2016syntheticdatavault} & 2016 & Copula functions &  NaN* \\ \bottomrule
\end{tabular}
\label{tab:single}
\end{table}

\subsection{A special multi-variate distribution modeling: SMOTE}\label{subsec:smote} 

The statistical methods mentioned in subsection \ref{subsec:multi-variate} explicitly compute the joint distributions of all variables. To avoid the tricky parameter selection procedure of these statistical methods in section  \ref{subsec:multi-variate}, Synthetic Minority Over-sampling Technique (SMOTE) \cite{Smote} was proposed to generate samples by interpolation. Each piece of data from an individual was treated as a point in the data space. The distributions of data were not modeled explicitly in SMOTE; SMOTE approximated the distribution by assuming that the data space can be spanned by all existing data points and sampled from the distribution by interpolating existing data points. SMOTE focused on data imbalance problems and created data points that belong to the minority class. Extensions of SMOTE algorithms \cite{SmoteReview} mainly focus on three aspects:
\begin{enumerate}
    \item The initial selections of minority samples. Borderline-SMOTE \cite{Borderline-SMOTE} presumed that samples away from the classification borderline may contribute little to the classification. Thus they selected minority samples that were close to the classification borderline. 
    \item Strategies for data synthesizing. For time-series data, INOS \cite{INOS} was proposed to maintain the inner structure among time series. 
    \item Adaptive generation of synthetic samples. ADASYN \cite{ADASYN} generated synthetic samples adaptively: they used a distribution for all minority class examples weighted by the classification difficulty of these samples, and for a fixed number of synthetic samples, the samples which were difficult to classify will be generated in larger numbers. 
\end{enumerate}

 A detailed SMOTE family review can be found in \cite{SmoteReview}. \textcolor{black}{Being well-engineered and implemented in many toolkits \cite{imbalancedlearn}, SMOTE and its variants has proved their efficiency in the medical data analysis domain, especially for disease classification whereas the number of patients is much less than the number of normal controls. Applications include Parkinson Disease classification \cite{smotepd}, Alzheimer's Disease Classification \cite{adasynad}, etc. }
 
 The SMOTE strategy considers all attributes together, while it fails to model the relations between attributes. Mathematically, considering $x_1^T$ as the {\itshape label} vector for the combination of all other attributes, the synthetic samples in SMOTE only consider the conditional distribution $Pr(x_{\lnot 1}^T|x_1^T)$, while rigidly inheriting the marginal of non-label attributes from real datasets by interpolation. Moreover, the nature of oversampling is also cursed with the mode collapse problem for samples in the minority class, where the synthetic data lacks diversity. 

\subsection{Sampling from conditional distributions: multiple imputation}\label{subsec:multipleimputation} 
A further improvement for the attribute relation modeling is multiple imputation, shown in Table \ref{tab:multipleimputation} . Initially proposed for missing data imputation, the concept of multiple imputation proposed by Rubin \cite{rubin93} has also been widely used in privacy protection data releases. The main concept of multiple imputation is to produce partially synthetic datasets, where the missing attributes are predicted by other non-missing attributes. In the scenarios of privacy protection, the sensitive attributes are treated as "missing attributes": sensitive attributes are replaced by the values conditionally synthesized from non-sensitive attributes.

For example, let's consider dataset $X=\{x_1^T,x_2^T,...,x_N^T\}$ with $N$ attributes, and the subset $X_C=\{x_1^T,x_2^T,...,x_C^T|C < N\}$ contains missing values (or sensitive values in privacy protection scenarios). The basic idea of multiple imputation is to sample each attribute $x_i^T$ from conditional distributions $\{x_i^T\sim P(x_i^T|x_{1}^T,x_{2}^T,...x_N^T), i\in [1,C]$\}, respectively. We will address all non-confidential (or non-missing) attributes as $X_U=\{x_{C+1},x_{C+2},...x_N\}$ for a clear statement. 

The conditional distributions can be modeled explicitly with a specific mean and an variance. For example, General Additive Data Perturbation (GADP) \cite{GADP} was proposed to define means and variances for the conditional distributions. Later in 2003, Information Preserving Statistical Obfuscation (IPSO) \cite{IPSO} was proposed. In the IPSO, the synthetic confidential attributes were obtained by multiple regression of $X_C$ on non-confidential attributes $X_U$; and this multiple regression model, also known as general linear model (GLM) has then been improved by regression trees \cite{caiola2010random} and fuzzy c-regression \cite{fuzzyMI}. Thus, after IPSO, the multiple imputation algorithm using SVM \cite{MISVM} allowed the synthesis of discrete variables (but it can only be applied to discrete variables). 


Multiple Imputation by Chained Equations (MICE) \cite{wulff2017multiple} also used regression models to synthesize sensitive attributes from nonsensitive attributes, but it is also featured by an iterative synthesis strategy, and the final synthesis results were pooled by all results synthesized during iterations. Applications can be found in breast cancer data synthesis \cite{goncalves2020generation}. 

Another extension of multiple imputation methods is to improve the privacy-protection during data release. The multiple imputation does not protect data privacy by nature. PeGS \cite{PGS2013} then introduced the differential privacy concept in multiple imputation algorithms, and an application of their algorithm on healthcare data can be found in \cite{PGSapplication}. 

Multiple imputation algorithms modelled the condition of missing attributes (or sensitive attributes) on existing attributes (or all attributes). However, for fully synthetic data, one needs to traverse all variables to investigate the cross-conditional distributions, which is tedious and time-consuming. 


\begin{table}[]
\caption{Papers using multiple imputation. }
\centering
\begin{tabular}{p{3cm}p{1cm}p{4cm}p{4cm}}
\toprule
Paper   reference & Year & Methods & Medical data applications \\ \midrule
GADP \cite{GADP} & 1999 & Defining mean and variances for the distributions of $X_C$   conditioned on $X_U$ & NaN \\ \midrule
IPSO \cite{IPSO} & 2003 & General linear models for $X_C$ from $X_U$ & NaN \\ \midrule
CART \cite{caiola2010random} & 2010 & Random forests for $X_C$ on $X_U$ (only applicable to discrete   sensitive attributes) & Demongraphics (Census data) \\ \midrule
\cite{fuzzyMI} & 2009 & Fuzzy c-means for $X_C$ on $X_U$ & Demongraphics (Census data) \\ \midrule
\cite{MISVM} & 2010 & Support vector machines for $X_C$ on $X_U$ & Health insurances data \\ \midrule
\cite{goncalves2020generation} & 2020 & MICE & Cancer registry data from the Surveillance Epidemiology and   End Results (SEER) program \\ \midrule
PeGS \cite{PGS2013} & 2013 & General linear models with differential privacy for $X_C$ from   $X_U$ (only applicable to discrete sensitive attributes) & Public Patient Discharge Data from California Office of Statewide Health Planning and Development \\ \midrule
PeGS applications \cite{PGSapplication} & 2013 & General linear models with differential privacy for $X_C$ from   $X_U$ (only applicable to discrete sensitive attributes) & Public-use data files from Centers for Medicare and Medicaid Services \\ \bottomrule
\end{tabular}
\label{tab:multipleimputation} 
\end{table}





\subsection{Sampling from conditional distributions with attribute relationships: Probabilistic Graphical Model (PGM)} \label{subsec:bayesianNetworks} 
To model the relations among attributes, a probabilistic graphical model (PGM) can be used. The edges for the PGM are relationships between attributes, while each node represents a conditional distribution of one attribute. 
\begin{figure}[h]
  \centering
  \includegraphics[width=8cm]{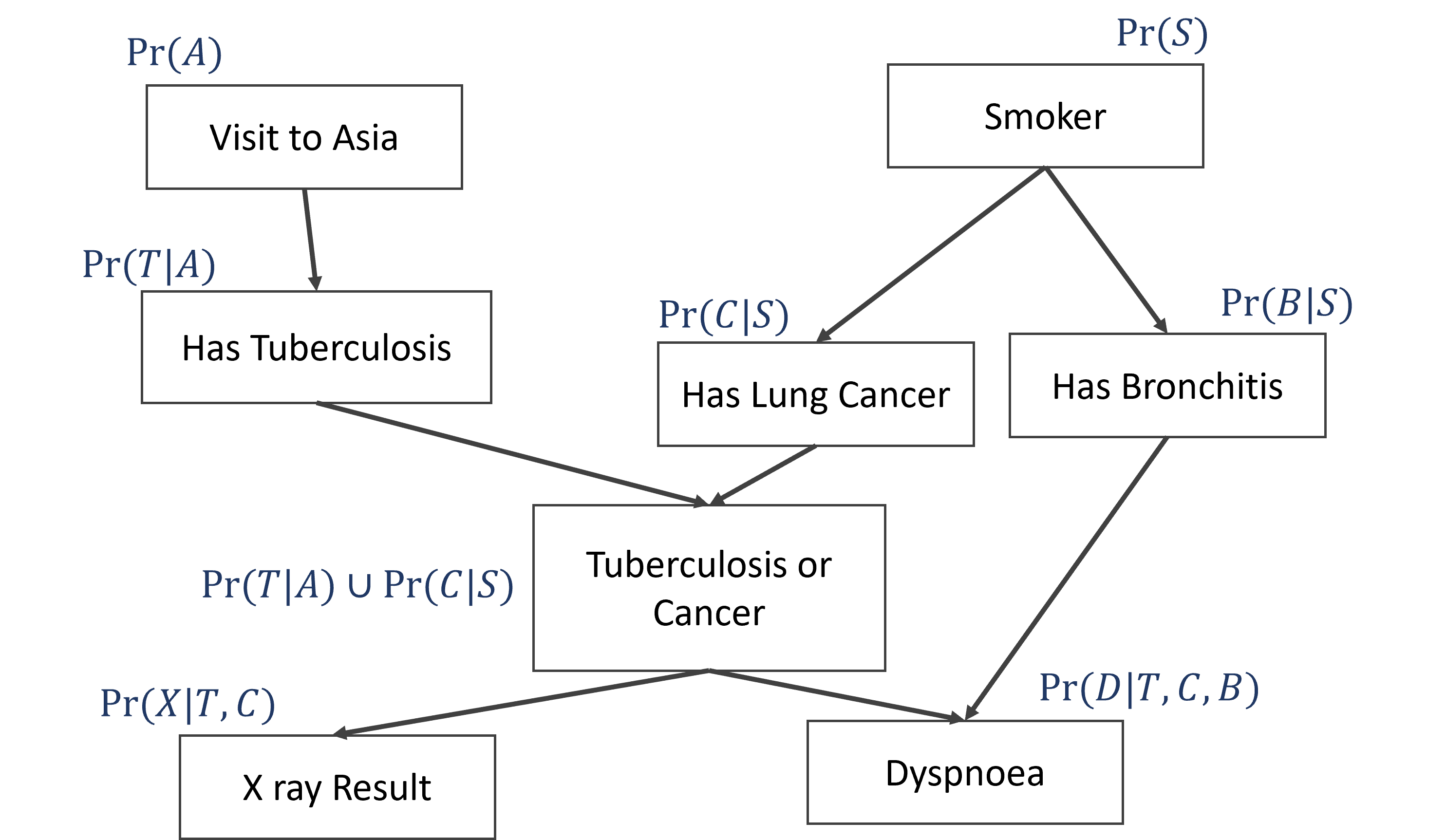}
  \caption{An example for bayesian networks.}
  \label{fig:bayesian}
\end{figure}
A Bayesian Network (also known as Bayesian belief network) describes the joint probability distribution, represented in a directional acyclic graph (DAG) structure among attributes of interests. The directions of the edges show casual relations between attributes. In Fig. \ref{fig:bayesian}, we present a popular showcase for healthcare Bayesian Networks, the Asia Network, defined in 1988 \cite{lauritzen1988local}. The Asia Network presumes a Bernoulli distribution for all attributes. To synthesize a piece of patient data from the Asia Network, commonly, one should first sample from $Pr(S)$ and $Pr(A)$ as a start node, then obtain the following attributes from their conditional distributions, shown in Fig. \ref{fig:bayesian}. The inference of Bayesian Networks does not always start with independent variables. In summary, the Bayesian Networks provide a graphical causal relation for attributes and allow an efficient calculation of the conditional distributions of each attribute \cite{lauritzen1988local}.   


Three steps are required in order to use Bayesian Network to generate synthetic samples:
\begin{itemize}
    \item First is structure learning, i.e., identifying the causal relations between attributes. 
    \item Second is parameter learning, i.e., learning the conditional distribution of each variable. 
    \item The third step is the inference: values of each attribute are first sampled from sets of initial attributes and propagated to values in other attributes according to conditional distributions. 
\end{itemize}
 It should be noted that the structure and parameter learning of Bayesian Networks can always be knowledge-driven, i.e., human experts can pre-define the relations between each attribute by their experiences. For data-driven algorithms, the structure and parameters are learned simultaneously. Data-driven methods for constructing a Bayesian Network include
 \begin{itemize}
    \item Constraint-based algorithms. In these algorithms, conditional independence tests were be used to evaluate the dependency between each pair of attributes, and a BN was constructed using related attribute pairs. 
    \item Score-based algorithms. The score-based algorithms first searched all possible structures, and then used a score function \cite{BIC,MDL} to evaluate these graphs. 
    \item Hybrid algorithms. These algorithms used constraint-based algorithms to generate a subspace of all possible structures and then used score-based algorithms to evaluate and select these graphs.      
\end{itemize}
We summarized all Bayesian Network-based algorithms in Table \ref{tab:bayesian}. Bayesian Networks have been widely used in medical data synthesis \textcolor{black}{because the inherent suitability of Bayesian networks for knowledge-driven structure design and parameter learning enable improved trustworthiness of medical experts in this kind of models. }

\begin{table}[]
\caption{Bayesian Network-based algorithms for data synthesis. *Although these papers did not report synthesis performance on medical data, they are open-sourced and easily implemented. }
\centering
\begin{tabular}{@{}p{2cm}p{1cm}p{4cm}p{4cm}p{2cm}@{}}
\toprule
Paper reference & Year & Structural and parameter learning & Inference & Medical data applications \\ \midrule
\cite{sun2015bayesian} & 2015 &  Score-based (tabu search by Python Package bnlearn \cite{bnlearn}) & Global sampling  & Demographics \\\midrule
PrivBayes \cite{Privbayes} & 2017 & Constraint-based (Mutual Information and differential privacy) & Global sampling &  NaN* \\\midrule
DataSynthesizer \cite{ping2017datasynthesizer} & 2017 & PrivBayes & Global sampling &  NaN* \\\midrule
 \cite{deeva2020bayesian} & 2020 & Score-based (AIC by Python Package pomegranate \cite{schreiber2017pomegranate}) & Global sampling & Demographics\\\midrule
\cite{tucker2020generating} & 2020 & Constraint-based (FCI with EM for missing data) & Global sampling & CPRD Aurum data synthesis \\\midrule
 \cite{kaur2021application} & 2021 & Score-based (by Python Package bnlearn \cite{bnlearn}) & Heart Disease (UCI), Diabetes datasets (UCI), MIMIC-III\\\midrule
\cite{luo2021oversampling}& 2021 & Constraint-based ($G^2$-test) & Global sampling from the label attribute  & Breast cancer (UCI), Diabetes (UCI)\\\midrule
PrivSyn \cite{zhang2021privsyn} & 2021 & Constraint-based (Independent Difference (InDif for short)) & Gradually Update Method (GUM) &  NaN*\\ \bottomrule
\end{tabular}
\label{tab:bayesian}
\end{table}

\section{Deep learning}\label{sec:gan}

Two types of deep neural networks (also referred to as Deep Learning) have been widely used in data synthesis: Auto-Encoder (AE) \cite{AE2013} and generative adversarial networks (GAN) \cite{gan2014}. They are all composed of stacked linear or non-linear functions, and the major difference between these two types of methods is their target functions. An AE usually has two basic components, an encoder $\mathcal{E}$ that maps vectors in the data space into a latent space, and a decoder $\mathcal{D}$ that maps the latent space features into the data space. Mathematically, the objective function for an AE with an input real data $x$ is defined as 
\begin{equation}\label{eq:aeloss}
    L(X,\mathcal{E},\mathcal{D})=||\mathcal{D}(\mathcal{E}(x)) - x||_p,
\end{equation}
where $||\cdot||_p$ is the p-norm. Synthetic data can be generated from AE by first sampling vectors from the latent space and then mapping the sampled vectors into the data space. 

The GAN method, however, uses an additional network, discriminator, to optimize the performance of the data synthesizer. A GAN is also composed of two components: a generator $\mathcal{G}$ and a discriminator $\mathcal{D}$. The inputs for generators are usually noises $z$, which are transformed into meaningful data vectors by deep learning models and can improve the variety of synthetic data. The inputs for the discriminator are both synthetic data from generators $\mathcal{G}(z)$ and real data $x$ for reference. The objective function of a GAN is 
\begin{equation}\label{eq:ganloss}
    L(X,\mathcal{G},\mathcal{D})= E_x[\mathrm{log}(\mathcal{D}(x))] + E_z[\mathrm{log}(1-\mathcal{D}(\mathcal{G}(z)))],
\end{equation}
where $E_x$ and $E_z$ are expectations over all data instances. The generator can have many variants \cite{srivastava2017veegan,arjovsky2017wgan}, and some researches even use AE as an generator \cite{CorGAN,longgan}, and the loss functions are then a combined value of both Eq. \ref{eq:aeloss} and Eq. \ref{eq:ganloss}. 

Considering the overlapping design of AEs and GANs, in this section, we group the deep learning-based algorithms according to their target data types. The detailed architectures of deep learning algorithms are plotted as well. Although the DP and fairness techniques focus on synthesizing tabular data (whereas they can always be used on sequential data), we would like to mention it in a subsection to point them out particularly. 

\subsection{Deep neural networks for tabular data}
According to the output of deep neural networks, we further divided the tabular data deep learning models into two categories: half synthesis networks whose target is to impute values, and fully synthesis networks.

\subsubsection{Half synthesis networks} 
\begin{figure}[h]
  \centering
  \includegraphics[width=\linewidth]{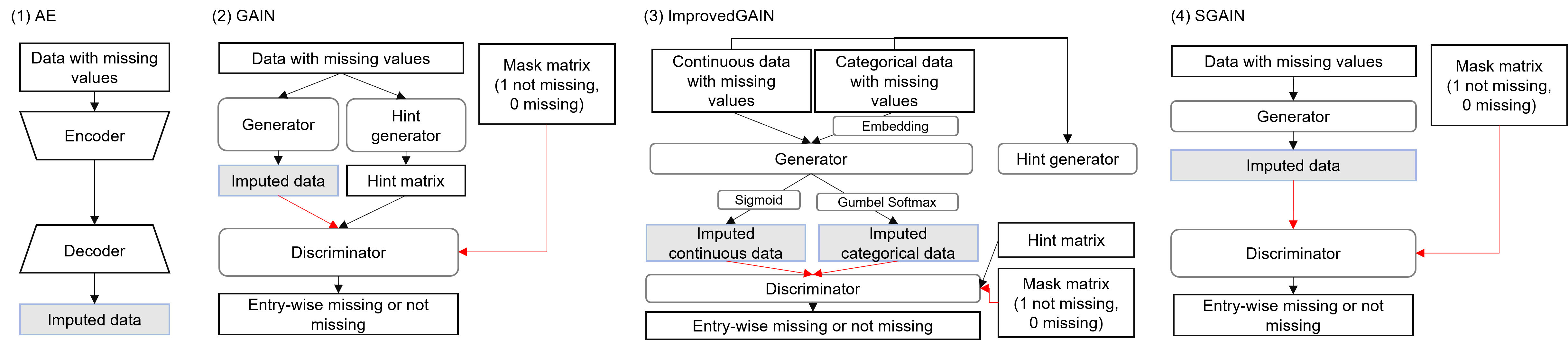}
  \caption{Architectures for tabular data imputation.}
  \label{fig:imputationgan}
\end{figure}

We plotted the structures of half synthesis networks in Fig. \ref{fig:imputationgan}.
A use case of AE in clinical data imputation can be found in \cite{imputationAE2017}. Denoising autoencoders (DAE) \cite{imputationAE2008} was proposed to extract robust feature representations, but its structure has also been used in missing data imputation. During training, the DAE first set several elements of inputs to zero randomly and then was trained to reconstruct the values of these elements. Once well-trained, the DAE could be used for missing data imputation. The objective function for DAE is the MSE between the corrupted input and the corresponding ground truth. Thus, the training of DAE is fully supervised and requires a large scale of complete ground truths. 
To improve the fully supervised training of DAE, Generative Adversarial Imputation Nets (GAIN) \cite{gain} used a discriminator to optimize the synthesis performance during training. In data imputation tasks, the discriminators identify the real and fake elements in the imputed data. This imputation discriminator is similar to the patch discriminator \cite{patchgan} in image synthesis, i.e., instead of outputting a numeric result reporting real or fake (1 or 0), the discriminator produces a map locating the real and imputed values. This "patch-wise" discriminator has been widely used in imputation GAN models.

The GAIN models were further improved by GAN training tricks such as Gradient Penalty and Wasserstein Loss in SGAIN \cite{gainwithtricks}. To further specify the imputation of categorical and numerical missing variables, improved GAIN \cite{improvedgain} split these two kinds into variables and imputes them separately. 


\subsubsection{Fully synthesis networks}\label{subsubsec:tabgan}
For the fully synthesis networks, we will discuss these networks according to their architectures. 

\noindent\textbf{Encoder-decoder or not: A debate on tabular data synthesis}.
AE models, which are featured by the encoder-decoder structure, can be applied to tabular data synthesis. Synthetic data can be generated by manipulating the hidden feature vectors derived from real data. Examples can be found in AE-ELM \cite{AE-ELM} and OVAE \cite{OVAE}, which used VAE to model the distributions of latent feature vectors. 

\begin{figure}[h]
  \centering
  \includegraphics[width=\linewidth]{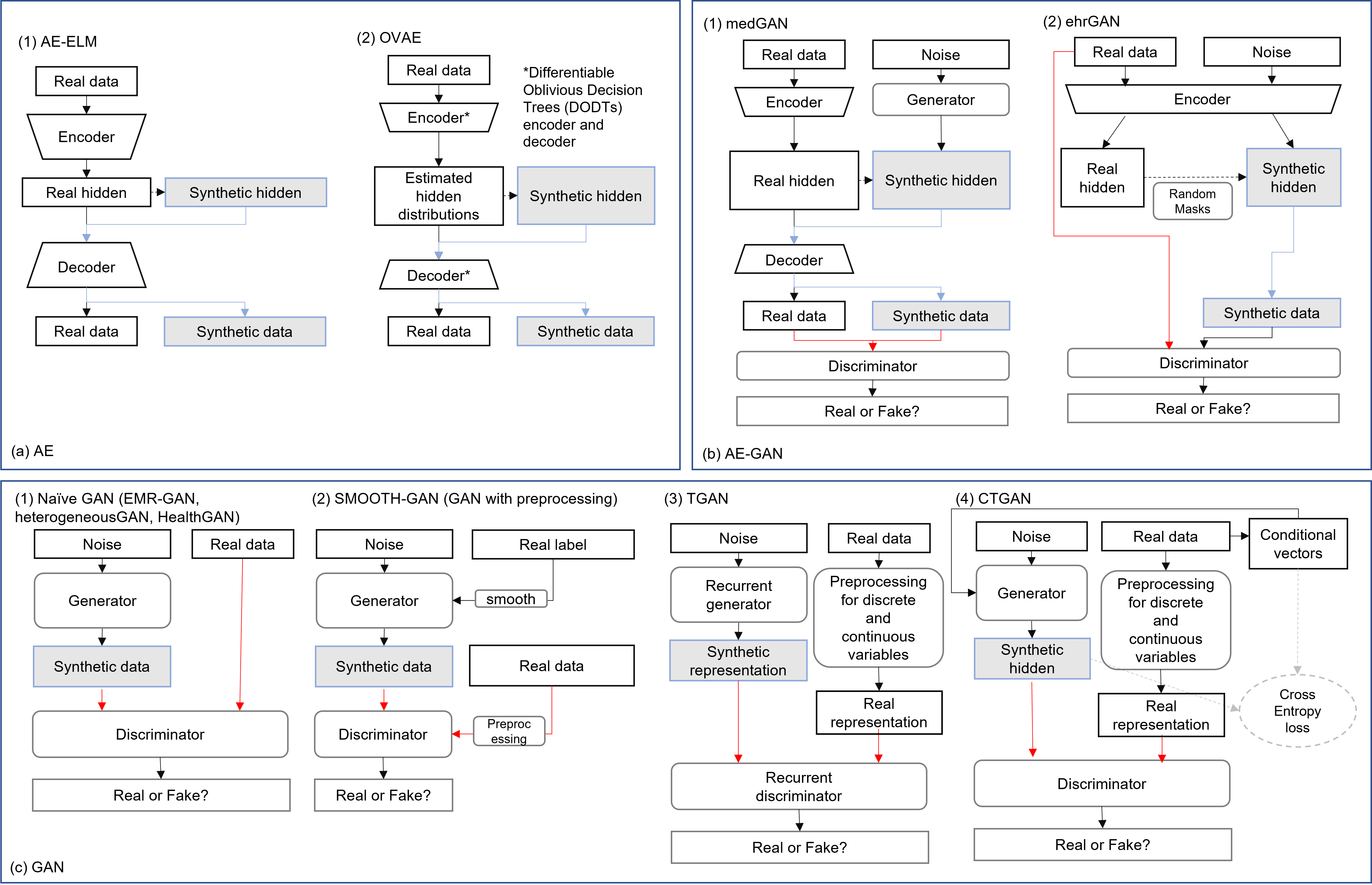}
  \caption{Architectures for tabular data synthesis, grouped by the usage of an encoder-decoder structure.}
  \label{fig:aeornotae}
\end{figure}

As for the GAN models, we discovered a long-lasting debate over the object to be generated in the tabular GAN generators: some algorithms \cite{medGAN,ehrGAN} used an encoder-decoder structure to map the original data into a latent space, while others generated the vectors of data directly \cite{EMRGAN}. We plotted their architectures in Fig. \ref{fig:aeornotae}. 

Algorithms that used encoder-decoders to map feature vectors into a latent space tried to avoid synthesizing tabular data directly \cite{medGAN,ehrGAN}. For example, medGAN \cite{medGAN} used an encoder network that mapped the original values into a hidden space, and the generator generated values in the hidden space instead of in the original space. The encoder-decoder in medGAN was also pre-trained on real datasets, and the target for the pre-training was to reconstruct the input real datasets. EhrGAN \cite{ehrGAN} used the encoder-decoder structure to map a transition distribution of the form $P(\Tilde{x}|x)$, whereas $\Tilde{x}$ and $x$ are synthetic and real data. 

However, EMR-GAN \cite{EMRGAN} has a conflicting conclusion claiming that the direct synthesis of values works better than the synthesis of hidden feature maps. They argue that because "these GANs (medGAN \cite{medGAN}) rely on an autoencoder, they may be led to a biased model because noise is introduced into the learning process". SPRINT-GAN \cite{SPRINTGAN}, heterogeneousGAN \cite{heterogeneousGAN}, healthGAN \cite{dash2019healthGAN} and SMOOTH-GAN \cite{rashidian2020smooth} also synthesized data values directly.  

TGAN \cite{TGAN} also discovered that “Simply normalizing numerical feature to$[-1, 1]$ and using tanh activation to generate these (numerical) features does not work well”. However, TGAN did not turn to latent space synthesis and did not use encoder-decoder structures. TGAN introduced a statistical representation synthesis strategy known as mode-specific normalization. Instead of synthesizing numerical values directly, the authors of TGAN used a Gaussian Mixture Model with $m$ Gaussian distributions to model each feature, and they synthesized the parameters for GMMs. In their improved CTGAN \cite{CTGAN}, they replaced the LSTM generator in TGAN and used a conditional synthesis strategy to model the categorical features. Other pre-processing algorithms can be found at smoothGAN \cite{rashidian2020smooth}, where continuous data were pre-processed by deleting outliers and scaling, and discrete data were mapped into continuous scores.

\begin{figure}[h]
  \centering
  \includegraphics[width=\linewidth]{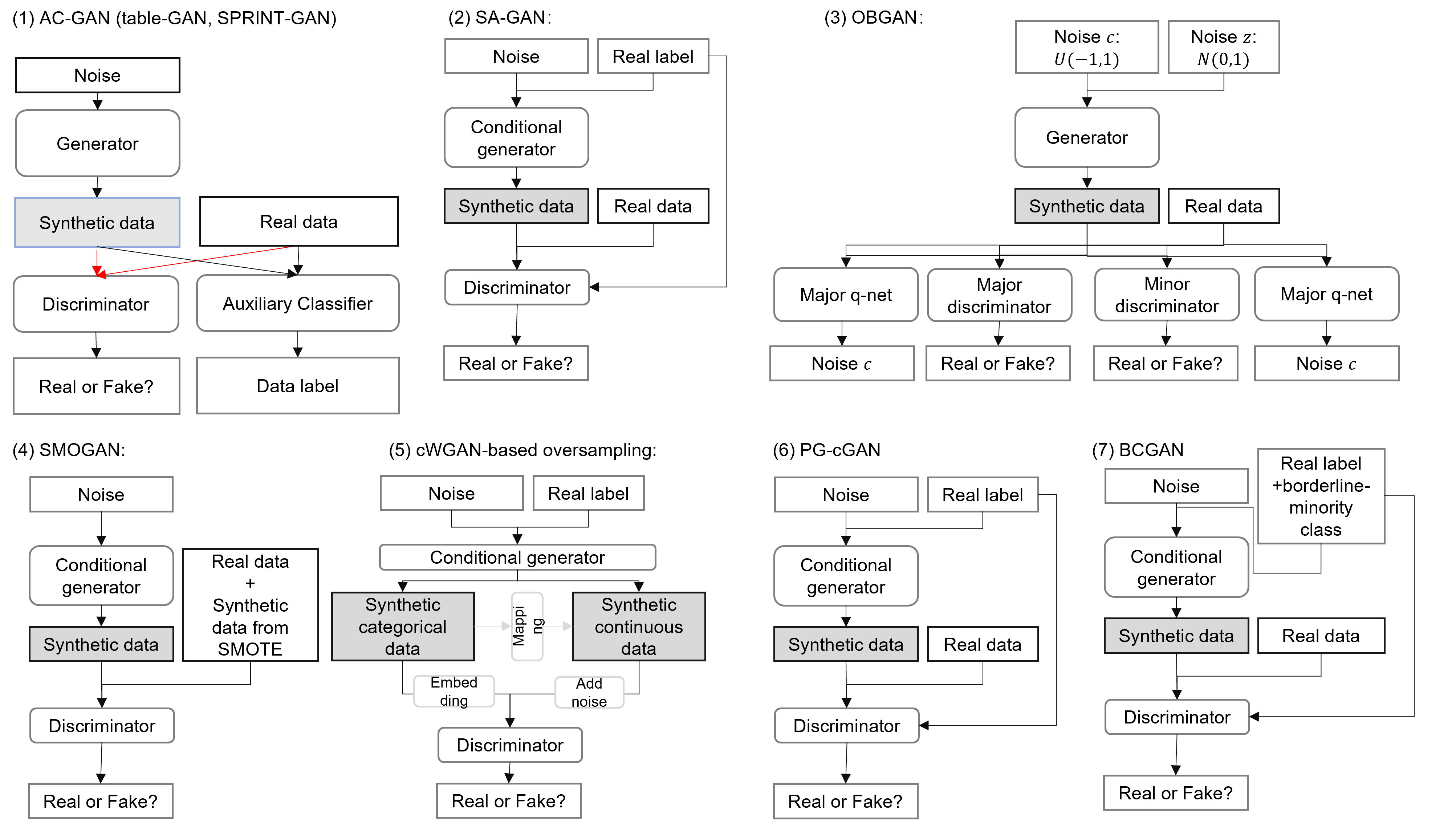}
  \caption{Architectures of tabular data synthesis with labels.}
  \label{fig:conditional}
\end{figure}

\noindent\textbf{Synthesis with labels}. For downstream prediction tasks, the labels of data should be synthesized alongside the data. ACGAN \cite{odena2017acgan} used an auxiliary classifier to synthesize data labels, and has successful implementations as SPRINT-GAN \cite{SPRINTGAN} and table-GAN \cite{table-gan}.

Some medical data synthesis applications aim to synthesize data within a target group, e.g., synthesizing data with the minority class. For minority class data synthesis, the conditional generator has been widely used. BCGAN \cite{bcgan} and OBGAN \cite{obgan} synthesized data points that were close to the decision boundary, and the former algorithm achieved borderline synthesis by introducing an additional borderline minority class; the latter used the Q-Net concept from InfoGAN \cite{infogan}, to allow output editing in GAN models. SMOGAN \cite{SMOGAN} used a SMOTE algorithm before GAN to augment the data points for GAN training. SAGAN \cite{SAGAN} brought the relation between single attributes and data labels to GAN models. To better analyze discrete variables, cWGAN-based oversampling \cite{cwganbasedoversampling} adjusted the GAN model, and embedded the discrete attributes. 

In addition to network architectures, different loss functions have also been introduced in deep generative models. ADS-GAN \cite{yoon2020adsgan} introduced the indetififiability loss, which maximizes the Euclidean distance between real and synthetic data. HealthGAN adopted the Wasserstein loss \cite{arjovsky2017wgan} during the training of GAN models.

\subsection{Deep neural networks for sequential data}

\begin{figure}[h]
  \centering
  \includegraphics[width=12cm]{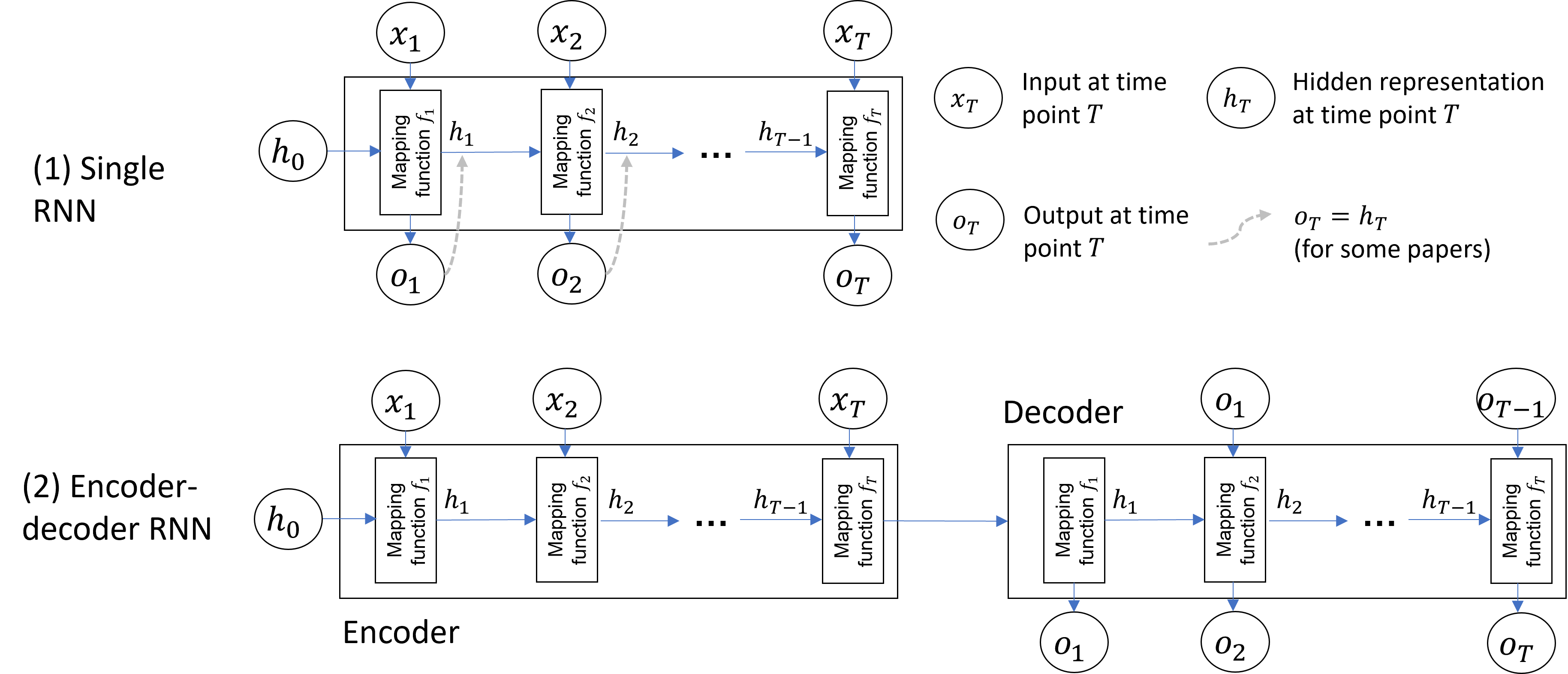}
  \caption{Architecture of a recurrent network (1) and a encoder-decoder recurrent network (2).}
  \label{fig:recurrent}
\end{figure}

For sequential data, recurrent structures have been utilized. The recurrent structures have many variants, such as Recurrent Neural Networks (RNN) \cite{rumelhart1986rnn}, Gated Recurrent Unit \cite{cho2014gru} and Long Short-Term Memory (LSTM) \cite{hochreiter1997lstm}, but they share a same basic structure shown in Fig. \ref{fig:recurrent}. Here, this recurrent mapping maps input $\{x_1,x_2,...,x_T\}$ into an output $\{y_1,y_2,...,y_T\}$. Each cell in the recurrent mapping receives two inputs: one from the present $x_T$, and one from the latest past $h_{T-1}$. For the first cell, the $h_0$ is addressed as the initial state, which, for many RNN implementations \cite{RNN}, are assigned by a vector of zeros. Each cell in the recurrent mapping outputs two outputs, but in some algorithms such as RNN and GRU, $h_T=o_T$. 

\begin{figure}[h]
  \centering
  \includegraphics[width=\linewidth]{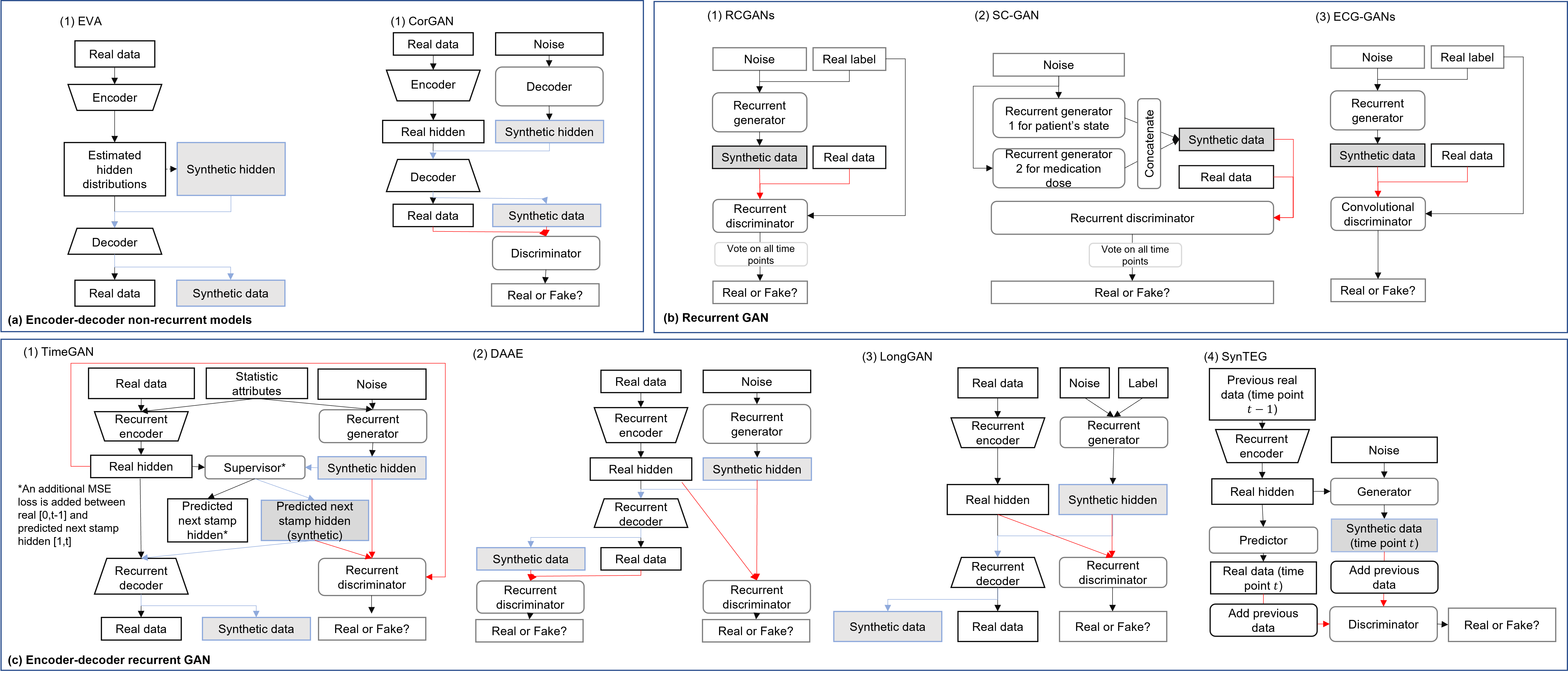}
  \caption{Architectures of deep neural networks for sequential data, grouped by the use of encoder-decoder structure and recurrent structure.}
  \label{fig:sequencegan}
\end{figure}

\textbf{There is an implied convention of using encoder-decoder structures in sequential synthesis models \cite{seq2seq}.} As shown in Fig. \ref{fig:recurrent}, the encoder-decoder structure in sequential synthesis allows the network to read all time points before producing outputs. The decoupling of data reading and data generating has been commonly practiced in sequential data synthesis. Thus, we can easily find successful applications of encoder-decoder recurrent structures for sequential EHR generation, including TimeGAN \cite{timegan}, DAAE \cite{DAAE}, LongGAN \cite{longgan}, and SynTEG \cite{SynTEG}, as is shown in Fig. \ref{fig:sequencegan}. 

However, some algorithms break this convention. We noticed that, in sequential EHR synthesis, RCGAN \cite{rcgan} and SC-GAN \cite{wang2019scGAN} generate data without the encoder-decoder structures. In addition, for ECG synthesis, we found another recurrent synthesis application without encoder-decoder structures \cite{ECG-GAN}. These algorithms generate sequences directly from noises. However, these papers did not claim that their direct synthesis is better than latent synthesis. 

Besides of recurrent networks, convolutional operations have also been used to investigate and preserve the inner correlations among time stamps, and papers including EVA \cite{eva} and CorGAN \cite{CorGAN} used 1-D convolutions on temporal datasets. The experiments in ECG-GAN \cite{ECG-GAN} also demonstrated the best synthesis performance using a recurrent generator and a non-recurrent discriminator. Despite the fact that they do not use recurrent structures, the EVA and CorGAN also decouple data reading and data generating with encoder-decoder structures. 

Here, we would like to elaborate on loss functions improvements for sequential data synthesis. In addition to the network architectures, TimeGAN introduced a supervised loss in the training of recurrent GAN models, where the temporal relationships between time stamps were used for supervised criteria. \textcolor{black}{The recurrent supervisor in TimeGAN was proposed to ``explicitly encouraging the model to capture the stepwise conditional distributions in the data.''} Practically, TimeGAN used an additional recurrent network named supervisor. The recurrent supervisor received the latent feature extracted from the real data $\{h_1,h_2,...h_t\}$(real hidden in Fig. \ref{fig:sequencegan}), and outputted the latent features in the  next time stamps $\{h'_2,h'_3,...h'_{t+1}\}$. The loss function of the supervisor is thus
\begin{equation}
    L_{\mathrm{MSE}} = \sum_{i=0}^t||h_i-h'_i||^2.
\end{equation}
\textcolor{black}{This inner sequential supervision proved its efficacy in medical data synthesis using a large private lung cancer pathways dataset.}

\subsection{Deep neural networks with additional targets}

\textcolor{black}{In this section, we will introduce deep neural networks with additional targets, such as differential privacy and fairness.} The concept of differential privacy provides a well-defined solution to data privacy protection, and detailed definitions of differential privacy will be elaborated on section \ref{subsubsec:DP}. In 2016, Differential Privacy Stochastic Gradient Descent (DP-SGD) \cite{dpsgd} was proposed to incorporate the concept of DP in deep learning algorithms by adding noises in the gradient during training stages. DPGAN \cite{dpgan} first introduced the concept of differential privacy (DP) into GAN models in 2018, and adopted the noisy gradient strategy as in DP-SGD. DP-AuGM and DP-VAEGM \cite{chen2018dpvae} also used the DP-gradient descent strategy during training and proposed to use AE- and VAE-based strategies for data synthesis. 

PATE-GAN \cite{pategan} used a private aggregation of teacher ensembles (PATE) mechanism during the training of discriminators to synthesize data according to the level of DP. PATE-GAN used $n$ teacher discriminators, and split the real data into $n$ subsets. Each teacher discriminator only discriminated between synthetic data and its corresponding subset of real data. PATE-GAN also implemented a student discriminator, which does not rely on any public data. The student discriminator only received synthetic data as inputs, and the labels of these data were assigned by the teacher discriminators. Successful applications of PATEGAN include implementations on Kaggle cervical cancer dataset \cite{kagglecervical}, UCI ISOLET dataset  \cite{uci2019} and UCI
Epileptic Seizure Recognition dataset \cite{uci2019}.

\begin{figure}[h]
  \centering
  \includegraphics[width=\linewidth]{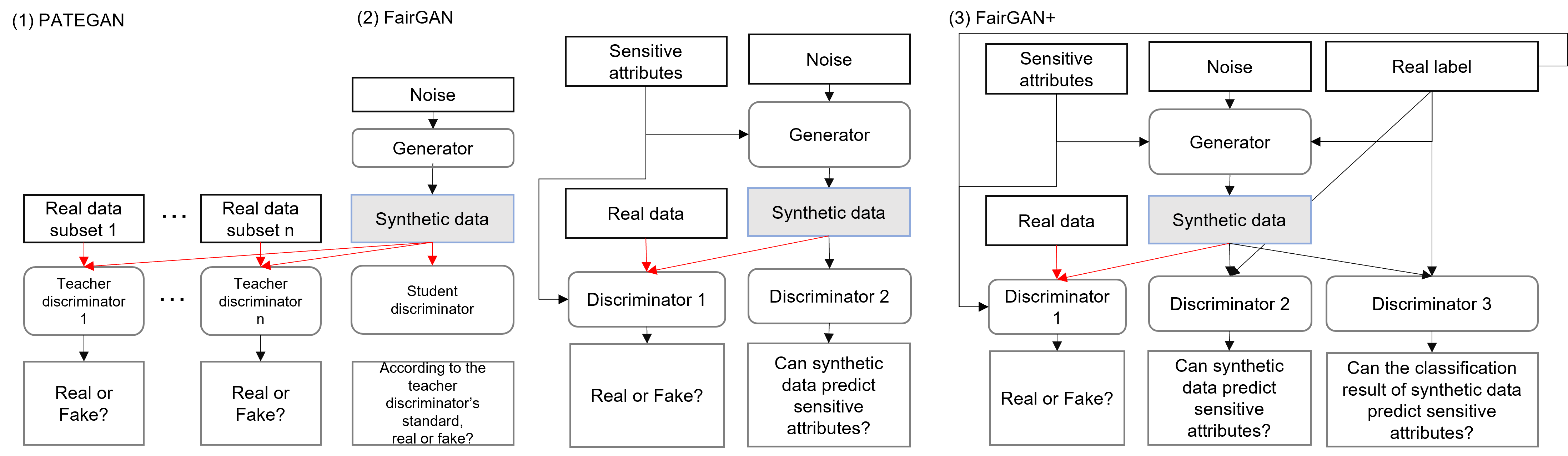}
  \caption{Architectures of deep neural networks for differential privacy and fairness.}
  \label{fig:othergan}
\end{figure}

FairGAN \cite{fairgan} and FairGAN+ \cite{fairganP} aim to improve data fairness by using conditional GANs. In FairGAN, an additional discriminator was used to minimize the attribute disclosure, i.e., to minimize the predictive ability of nonsensitive attributes on sensitive attributes. The FairGAN+ then introduced classification fairness. We include the architecture mentioned in this subsection in Fig. \ref{fig:othergan}. Although fairness is an important quality required in a trustworthy training dataset, these fairness generative models have not been applied to healthcare data. 

\section{Metrics}\label{sec:metrics}
\begin{figure}
    \centering
\tikzset{
    my node/.style={
        font=\small,
        rectangle,
        draw=#1!75,
        align=justify,
    }
}
\forestset{
    my tree style/.style={
        for tree={grow=east,
            parent anchor=east, 
            child anchor=west,  
        where level=0{my node=black,text width=5em}{},
        where level=1{my node=black,text width=5em}{},
        where level=2{my node=black,text width=30em}{},
            l sep=1.5em,
            forked edge,                
            fork sep=1em,               
            edge={draw=black!50, thick},                
            if n children=3{for children={
                    if n=2{calign with current}{}}
            }{},
            tier/.option=level,
        }
    }
}
    \begin{forest}
      my tree style
      [Metrics
          [Fidelity
            [Manual validation (e.g. \cite{SPRINTGAN,emerge2009,timegan})]
              [Statistical significance (e.g. \cite{tucker2020generating})]
              [Dimension-wise Probability (e.g. \cite{medGAN,CorGAN,EMRGAN})]
              [Marginal Probability (e.g.\cite{Privbayes})]
              [Attribute Correlation difference (e.g.\cite{CorGAN})]
              [Dimension-wise Prediction (e.g. tabular prediction \cite{EMRGAN}; sequential prediction: \cite{SynTEG})]
              [Additional discriminator (e.g. \cite{timegan})]
          ]
        [Utility
          [Train on synthetic test on real (e.g. \cite{dpgan,heterogeneousGAN}]
            [Train on hybrid test on real (e.g. \cite{SAGAN,obgan})]
        ]
        [Privacy
          [Attribute inference attack (e.g. \cite{emerge2009,medGAN,EMRGAN})]
            [Membership inference attack (e.g. \cite{medGAN})]
            [Differential Privacy* (e.g. \cite{dpgan})]
        ]
     [Fairness**
                [Data releasing fairness (e.g. \cite{fairganP})]
                [Data modeling fairness e.g. \cite{fairganP})]
      ]
    ]
    \end{forest}
    \caption{Tree diagram of evaluation metrics. *It should be noted that differential privacy is not strictly a metric by definition. **Fairness is not commonly evaluated in medical data synthesis, while we still introduce this metric due to its high correlation with trustworthiness. }
    \label{fig:metrics}
\end{figure}
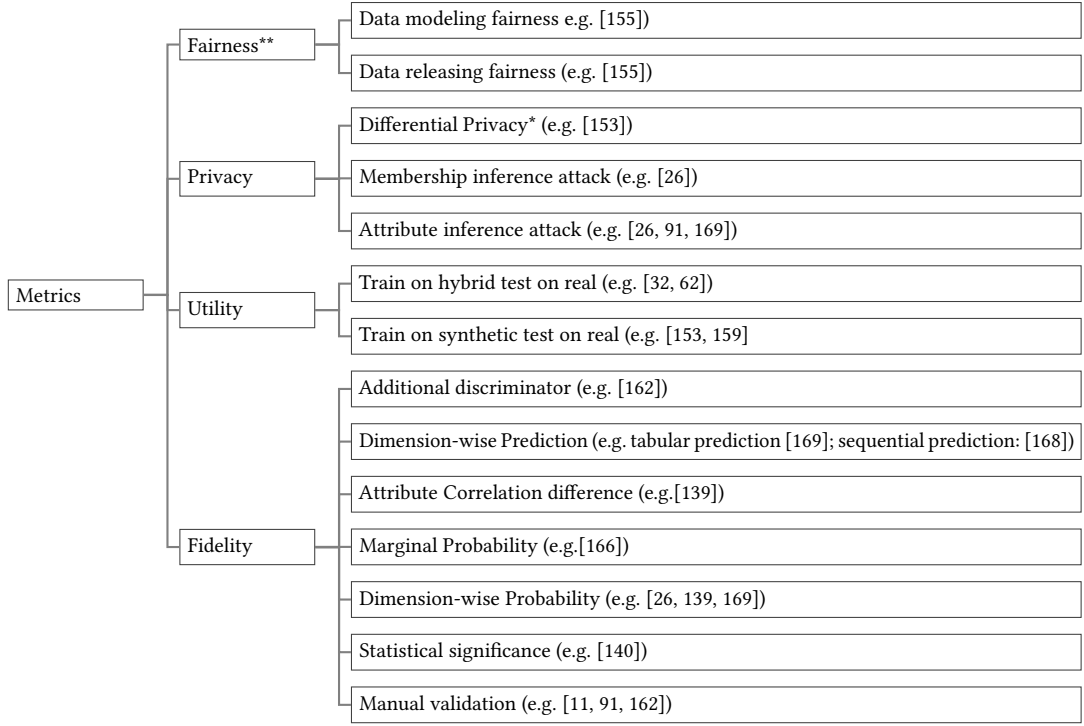

From the selected papers, we categorize their evaluation metrics into four classes, shown in Fig. \ref{fig:metrics}: 1) fidelity in section \ref{subsec:fidelity}: how real are the synthetic data; 2) utility in section \ref{subsec:utility}: how much can synthetic data helps in downstream tasks; and 3) privacy in section \ref{subsec:privacy}: how private the synthetic data is and 4) fairness in section \ref{subsec:fairness}.

\subsection{Fidelity}\label{subsec:fidelity}
The fidelity of synthetic data can be measured by a board of experts, i.e., by inviting clinicians who work closely with the real data to inspect the synthetic data. For example, SPRINT-GAN \cite{SPRINTGAN} used ACGAN to generate a group of synthesized sequential health data, including blood pressure and medication counts. They then mixed up both real and synthetic patients and invited three experienced physicians to determine whether the sequential data is real. EMERGE \cite{emerge2009} invited a medical expert to review the synthetic healthcare records and to identify some records that had content problems or inconsistencies, such as the disease category did not match the symptoms of synthetic patients. To further help clinicians investigate data fidelity, dimensional reduction algorithms were used to visualize the data distribution. TimeGAN \cite{timegan} used T-SNE \cite{tsne} algorithm to visualize the distributions of real and synthetic data points. 

In addition to the manual validation, the statistical closeness between real and synthetic data can also be measured, and can provide quantitative result for the synthesis performance. The statistical closeness can be measured in several aspects:

\noindent\textbf{Statistical significance.} Two of the most commonly used textbook measurement of the similarity between two datasets in medical data synthesis are the Chi-squared test and the two-sample Kolmogorov–Smirnov (KS) test. The chi-squared test is commonly used in discrete variables. In the Chi-squared test, one will obtain a $p$ value indicating the significance of this test, and commonly, a $p>0.05$ indicates there is no significant difference between real and synthetic variables \cite{swinscow2002statistics}. In the KS test, the significance is measured under a specific confidence level, and each confidence level has a unique threshold for the statistic. Ones can claim that the synthetic data has similar distributions as the real data \cite{massey1951kolmogorov} if the statistic is lower than the threshold.

\noindent\textbf{Dimension-wise Probability.} The Kullback–Leibler (KL) divergence \cite{kullback1951kldivergence} can be used measure the distribution difference between real attributes and synthetic attributes. For discrete variables, the KL divergence from the synthetic distribution $Q$ to the real distribution $P$ is defined as 
\begin{equation}
    D_{KL}(P||Q) = \sum_{x\in \mathcal{X}} P(x)\mathrm{log} (\frac{P(x)}{Q(x)}),
\end{equation}
where the $\mathcal{X}$ denotes all possible values in both datasets. For continuous variables, the KL divergence from $Q$ to $P$ is defined by a integral:
\begin{equation}
    D_{KL}(P||Q) = \int_{-\infty}^{\infty} p(x)\mathrm{log} (\frac{p(x)}{q(x)}) dx,
\end{equation}
where $p$ and $q$ are the probability densities of $P$ and $Q$.

Specially for the ICD9 embedded EHR data, Bernoulli success possibility of each dimension can be used for a dimension-wise probability metric (an example can be found in CorGAN \cite{CorGAN}). For each ICD9 code, the “success” in the Bernoulli trail is defined by the frequency of this code in all time points from all patients. The frequency for each ICD9 code should be similar between the real and the synthetic dataset. 

\noindent\textbf{Marginal Probability.} Dimension-wise probability measures the distribution difference on single attributes, and the marginal probability measures the joint probabilities among attributes. A $k$-way marginal is associated with a subset of all $k$ attributes. For example, $Pr(x_1,x_2)$ is a 2-way marginal. The KL divergence can also be used to measure the $k$-way marginal difference. In PrivBayes \cite{Privbayes}, the total variation distance between itself and the real marginal is used to evaluate the marginal distribution. Mathematically, the total variation distance between the synthetic distribution $Q$ and the real distribution $P$ is
\begin{equation}
   \mu(P,Q) = \mathrm{sup}_\mathcal{X} |P - Q|,
\end{equation}
where the supremum runs over all possible values or finite grids (for continuous variables) \cite{pollard2005total}. As with KL divergence, a lower distance indicates a better synthesis performance.

\noindent\textbf{Attribute Correlation difference.} The correlation between attributes is also an important feature in datasets. Preserving the correlation among attributes assures the clinical usage of the synthetic dataset. To prove that feature correlation are preserved, TGAN \cite{TGAN} computed the Mutual Information between each pair of feature columns for single tables. Pearson’s correlation \cite{CorGAN} is another correlation function measuring the intrinsic pattern among features. However, they do not qualitatively compare the correlation differences. 

\noindent\textbf{Dimension-wise Prediction.} This evaluation also measures how well the synthesis model captures the inner correlations in real datasets. A random attribute is $x_{(k,real)}$ selected in the real dataset, and the rest of attributes $x_{(\lnot k,real)}$ in the real dataset and $x_{(\lnot k,syn)}$ in the synthetic dataset are used to train two classifiers to predict the value of this selected attribute. The performance of these two classifiers should be similar. For sequential EHR data, the value for the last time point is often chosen as the target attribute \cite{DAAE}.

\noindent\textbf{Additional discriminator}. Some articles \cite{timegan,DAAE} introduced an additional discriminator to classify synthetic data from real data. Different from the discriminator in GAN models, this additional discriminator is not optimized alongside the generator. A lower accuracy of the additional discriminator indicates a better synthesis performance.

\subsection{Utility}\label{subsec:utility}
The utility of synthetic datasets is measured by the performance of synthetic dataset on downstream tasks. Usually, a label is assigned to each record in the datasets. For privacy preserved data synthesis, the Train on Synthetic Test on Real (TSTR) strategy has been widely used. A good synthetic dataset should achieve similar performance as the real dataset, and this classification similarity can be measured by KS tests. For data imputation and data augmentation, the utility is measured by classification improvement. The imputed data and augmented data should be able to improve the classification accuracy compared to the original dataset.  

\subsection{Privacy} \label{subsec:privacy}
Rubin in 1993 \cite{rubin93} first proposed to use fully synthetic data for privacy preserved data sharing. However, synthetic data is not automatically private \cite{jordan2022synthetic}.  If a “privacy preserved” synthesis algorithm is proposed, either an empirical analysis of privacy breach risks or an analytic proof of privacy is recommended. Of all papers reviewed, three types of metrics evaluating the ability of privacy preserving have been used. They are: 1) attribute inference attack, including getting the specific values of data and the statistical properties of the dataset; 2) membership inference attack, which is to identify the presence a specific record; and 3) differential privacy, which provide analytic proofs of privacy. 

\subsubsection{Attribute inference attack}
The attribute inference assumes the attacker have identities of some non-sensitive attributes, i.e., partially compromised records, and they will use these attributes, as well as the synthetic dataset, to speculate the sensitive attributes \cite{attributeReview}. Most algorithms used a similarity attack strategy \cite{emerge2009,medGAN,EMRGAN}. First, a subset of the real records is randomly sampled as the compromised records. For each individual in the compromised records, $n$ attributes are randomly chosen and provided to the attacker. Next, for each record in the compromised dataset, the attacker will compute the similarity of its publicized $n$ attributes with all records in the synthetic dataset. The similarity measurements include mean absolute value difference and the Euclidean distance \cite{IPSO}. Finally, the attacker can obtain the sensitive attributes of the most similar synthetic record as the sensitive attributes of this real individual.

Specifically, since most EHR synthesis used ICD9 codes, thus their data only contains discrete values. They \cite{emerge2009,medGAN,EMRGAN} performed k-nearest neighbor classifications on the synthetic dataset for each sensitive attribute, and by inferencing the classifiers on the compromised dataset, they could obtain the discrete values for the sensitive attributes. Finally, for each sensitive attribute, the classification accuracy metrics were reported. Lower accuracies indicates a higher privacy preserving ability. 

\subsubsection{Membership inference attack}
The membership inference attack assumes that the attacker only have the access to data, while do not have the access to the generative models. The attacker can obtain a set of complete records, i.e., all attributes are publicized, and by observing the synthetic dataset $S$, the attacker will determining whether a given data record in $P$ was part of the synthetic model' training dataset \cite{hu2021membership}.  \textbf{Most membership inference attacks can be seen as a classification task}: the target is to classify whether each record in $P$ is 0 (not presented in the training dataset) or 1 (presented in the training dataset). 


Most data generative models used a metric-based membership inference attack and are performed in an unsupervised manner. These metric-based membership inference presumes that, the synthetic records must bear similarity with the records that were used to generate them \cite{chen2020ganleaks}. Using different distance metrics, the similarities between the target record $p_i$ and all records from $S$ can be measured. If the mean similarity between this record and all synthetic records is under a specific threshold, then this record is considered as 1 (presented in the training dataset). The metrics include Hamming distance (for discrete variables) \cite{medGAN} or the Euclidean distance (for continuous variables).

In 2017, a shadow model based membership inference attack \cite{shokri2017membership} was proposed, and then this shadow model attack was further implemented for generative models \cite{stadler2021synthetic}. For each record $p_i$, the shadow model attack train two generative models, one with $p_i$ and one without $p_i$. The synthetic records from the generative model trained with $p_i$ are assigned with label 1, and the synthetic data from the model without $p_i$ are assigned with label 0. Then, a classifier is trained on the synthetic records and the presence labels, and by inferencing this classifier with the publicized synthetic records $S$, the attacker can identify the presence of $p_i$ in the training dataset. 

\subsubsection{Differential Privacy (DP)}\label{subsubsec:DP}
Differential privacy (DP) is an important solution in the context of membership inference. The "differential" here indicates obtaining a patient privacy by tracking the difference of a released dataset.  

Considering adding a new individual into the dataset $D$, and the dataset with this newly added individual is $D'$. Attackers can easily identify the information of this individual by comparing the information between $D$ and $D'$. For example, by investigating the contingency tables of both datasets, the attackers find that $D$ has 4 males and $D'$ has 5 males so they can easily obtain the gender information of the newly added individual. Thus, the target of DP is to blurring the values of the data. Considering a function $K$ whereas attackers are only allowed to obtain the information by $K(D)$, the target of DP is to minimize the difference between the distributions of $K(D)$ and $K(D')$.

As mentioned previously, the KL-Divergence is used in measuring the distribution difference. Since the concept of DP is to measure the maximum privacy breach, the max divergence is used in \cite{dp}. Then we can easily obtain the target of DP algorithms mathematically: An algorithm $K$ gives $\epsilon$-differential privacy if the maximum divergence between the distributions of $K(D)$ and $K(D')$ is bounded by $\epsilon$, i.e.,
\begin{equation}
    Div_\infty(K(D)||K(D')) = \mathrm{max}_{d\in D} [ln\frac{Pr[K(D)\in S]}{Pr[K(D')\in S]}] \leq \epsilon.
\end{equation}
More commonly, this definition is written as
\begin{equation}
    Pr[K(D)\in S]\leq e^\epsilon Pr[K(D')\in S]
\end{equation}
if $K$ gives $\epsilon$-differential privacy. The DP can be achieved by adding specific patterns of noises when releasing the dataset, i.e.,
\begin{equation}
    K(D) = D + \mathrm{Noise}.
\end{equation}
The distributions of noises include Laplacian mechanism \cite{laplaciandp} and the exponential mechanism \cite{expdp}. For numerical data, the Laplacian mechanism outputs synthetic data with noises from a Laplacian distribution. For categorical data, the exponential mechanism introduces a scoring function, and produces the possibility of each value. 

Unlike the two types of inference attacks mentioned before, differential privacy does not provide a typical “evaluation metric”, while it provides analytic proof of privacy in models. Thus, DP is, at most of the time, achieved by introducing the DP mechanisms into the synthesis algorithm, and introducing DP provides a guarantee for the privacy protection. For example, Differentially Private Stochastic Gradient Descent (DP-SGD) \cite{dpsgd} introduced the concept of DP in the SGD optimizer in deep learning; and \cite{Privbayes} introduced DP in Bayesian Networks.

\subsection{Fairness}\label{subsec:fairness}
\textcolor{black}{Fairness is an essential quality for trustworthy AI. A fair dataset should avoid increasing the discrete impact among different groups, especially for "protected groups, or a category of people protected by law, policy or similar authority" \cite{medicalfairness}. In the medical context, the fairness of dataset assures  health equities among different races and sexes.} 

Considering the patients' identities as sensitive attributes, a fair dataset \cite{fairganP} should 1) not reveal the sensitive attributes through insensitive attributes (data releasing fairness) and 2) not lead to a biased downstream prediction with respect to the sensitive attributes  (data modeling fairness). Considering a dataset composed of three components $D=\{X_U,X_C,Y\}$, whereas $X_U,X_C,Y$ are insensitive attributes, sensitive attributes (race or sex), and the data labeling, respectively.

For data releasing fairness, two measurements can be used:

\noindent\textbf{Risk difference (RD) \cite{fairganP}. } The RD for data releasing is defined by
\begin{equation}
   RD_r = Pr(Y = 1|X_C = 1)-Pr(Y = 1|X_C = 0). 
\end{equation}

\noindent\textbf{Balanced error rate (BER) \cite{epsilonFairness}. }A trust model $f: X_U \to X_C$ is built to predict sensitive variables $X_C$ from insensitive variables $X_U$ \cite{epsilonFairness}. The balanced error rate (BER) of $f$ is defined as 
\begin{equation}
    BER(f(X_U),X_C) = Pr[f(X_U) = 0|X_C = 1] + Pr[f(X_U) = 1|X_C = 0].
\end{equation}
According to BER, a synthetic dataset $(X_U,X_C)$ is $\epsilon$-fair if for any trust models, 
\begin{equation}
    BER(f(X_U),X_C) > \epsilon
\end{equation}

For the measurement of data modeling fairness, a classification model $\eta: X_U \to Y$ is built to predict data labels $Y$ from insensitive variables $X_U$. The data modeling fairness requires that the prediction of $\eta$ is unbiased with respect to $X_C$. Mathematically, three metrics are defined to measure the data modeling fairness

\noindent\textbf{Risk difference (RD) \cite{fairganP}. } The RD for modeling, which is also known as demographic parity, is defined by
\begin{equation}
   RD_m = Pr(\eta(X_U) = 1|X_C = 1)-Pr((\eta(X_U) = 1|X_C = 0). 
\end{equation}

\noindent\textbf{Odds difference (OD) \cite{fairganP}.} The equality of odds requires the classifier to have equal true positive rates (TPR) and equal false positive rates (FPR) between two subgroups $X_C=1$ and $X_C=0$. Mathematically, the odds difference is defined by 
\begin{equation}
   OD_m = \sum_{y\in \{0,1\}} Pr(\eta(X_U) = 1|Y = y,X_C = 1)-Pr((\eta(X_U) = 1|Y=y,X_C = 0). 
\end{equation}



\section{Common pre-processing practices and datasets} \label{sec:datasets}
In this section, we will provide common pre-processing practices for both tabular and sequential data, and we will also provide open-sourced datasets available for the synthesis algorithms development. The datasets we included in this survey are open-access for research purposes. However, due to the high sensitiveness of healthcare data, accesses to these dataset may require formal inquiries. In addition, we will also share three released synthetic datasets whose synthesis procedures produce hands-on experiences for data synthesis practitioners. 

\begin{longtable}{p{2cm}p{1cm}p{2cm}p{4cm}p{4cm}}
\caption{Open-sourced datasets used for non-imaging medical data synthesis. \label{tab:datasets}}\\
\toprule
Dataset name & Patient number & Data type & Data information & Disease Category \\ \midrule
MIMIC-I (or MIMIC) \cite{moody1996mimic1} & 100 & Medical signals and sequential EHR & Patient monitor data,  patient-descriptive data (gender, age, record duration), symptoms, fluid  balance, diagnoses, progress notes, medications, and laboratory results & Potential hemodynamically  unstable \\ \midrule
MIMIC-II \cite{saeed2002mimic2}& 33,000 & Medical signals and sequential  EHR & Patient monitor data,  patient-descriptive data (demographics, admissions, transfers, discharge  times, dates of death), diagnoses, notes, reports, procedure data,  medications, fluid balances, and laboratory test data & diseases of the circulatory  system; trauma; diseases of the digestive system; pulmonary diseases;  infectious diseases; and neoplasms \\ \midrule
MIMIC-III \cite{johnson2016mimic3}& 46,520 & Medical signals and sequential  EHR & Patient monitor data,  patient-descriptive data, diagnoses, reports, notes, interventions,  medications, and laboratory tests data. & Diseases of the circulatory  system, pulmonary diseases, infectious and parasitic diseases, diseases of  the digestive system, diseases of the genitourinary system, neoplasms,  diseases of the genitourinary system, and trauma \\ \midrule
MIMIC-IV \cite{johnson2020mimic4}& 383,220 & Medical signals and sequential  EHR & Hosp module contains  patient-descriptive data, basic health data (blood pressure, height,  weight…), medication, procedure data, and diagnoses. Icu module contains  timing information data, patient monitor data, fluid balance, and procedure  data. & Diseases of the circulatory  system, pulmonary diseases, infectious and parasitic diseases, diseases of  the digestive system, diseases of the genitourinary system, neoplasms,  diseases of the genitourinary system, and trauma \\ \midrule
eICU-CRD \cite{pollard2019icu}& 139,367 & Sequential EHR & Vital signs, laboratory measurements, medications, APACHE  components, care plan information, admission diagnosis, patient history, and  time-stamped diagnoses. & pulmonary sepsis, acute  myocardial infarction, cerebrovascular accident, congestive heart failure,  renal sepsis, diabetic ketoacidosis, coronary artery bypass graft, atrial  rhythm disturbance, cardiac arrest, and emphysema \\ \midrule
Amsterdam UMCdb \cite{thoral2021umcdb} & 20,109 & Medical signals and sequential  EHR & Patient monitor and life support  device data, laboratory measurements, clinical observation and scores,  medical procedures and tasks, medication, fluid balance, diagnosis groups and  clinical patient outcomes & Not specified\\ \midrule
UT Physicians clinical database (UTP) \cite{UTPdata} & 5,501,776 & Sequential EHR & Demographic data, vital signs, immunization data (body site,  dose), laboratory data, transaction data (evaluation and management,  radiology, medicine, surgery, anethesia), appointment data, medications, and  invoices & diabetes mellitus, hyperlipidemia, hypertension, and  unspecified chest pain \\ \midrule
Breast Cancer Wisconsin dataset (UCI) \cite{uci2019} & 569 & Tabular data & Diagnoses, radiuses, texture data, perimeters, areas,  smoothness data, compactness data, concavity data, concave points data,  symmetry data, and fractal dimensions. & Breast cancer\\ \midrule
Heart Disease dataset (UCI) \cite{uci2019} & 303 & Tabular data & Demographic data, smoking status data, disease history data,  exercise protocols, chart data (blood pressure, heart rate, ECG), pain status  data, and diagnoses & Heart disease\\ \midrule
Diabete dataset (UCI) \cite{uci2019} & 70 & Sequential data & Iinsulin dose, blood glucose measurement, hypoglycemic symptoms, meal  ingestion, exercise activity & Diabete\\
\bottomrule
\end{longtable}


\subsection{Pre-processing methods for tabular data}
For tabular data, each row of the table represents a patient, and columns of the table are features describing the patient. Tabular data is straightforward for data analysis, and the statistical properties, such as mean values and standard deviations, of the population can be derived. A sub-type of tabular data is multiple-tables, or relational tables, where information from different sites and under different levels is linked with one column from the tables. It should be noted that these relational tables can be merged into a meta table by a unique combination of linkage variables.

Tabular data are composed of two sets of variables, continuous variables and categorical variables. Continuous variables include age, blood pressure and temperature. In many statistical modeling-based algorithm, continuous variables are often pre-processed into discrete variables. \textcolor{black}{It is because the finding a suitable prior distribution for continuous variables can be complex, and the joint or conditional distributions among multiple continuous variables are difficult to derive from data-driven methods. }Thus, for most statistical modeling synthesis algorithms, continuous variables are classified into categorical variables according to their value. For example, in PrivBayes \cite{Privbayes}, continuous variables were first discretized into a fixed number $l$ of equi-width bins and then binarized into $log l$ classes.

\subsection{Pre-processing methods for sequential data}
The sequential data majorly has two forms, the medical signals and the EHRs. Many datasets provide both data, while during synthesis and analysis, these two data forms have different pre-processing steps.

Medical signals include neurological signals such as EEG and fMRI, and physiological signals such as continuous blood pressure waveforms or continuous heart rates. For each time point, these signals only contain one value. These medical signals are often periodic, so pre-processing methods of these signals can focus on time domain \cite{fmrianalysisTemporal,pardey1996review} and frequency domain \cite{fmrianalysisSpectral}; and quantities such as amplitude \cite{pessoa2002neural} and frequency \cite{nuwer1988quantitative} for these signals are considered for the synthesis and analysis. To medical signals, pre-processing \cite{bigdely2015prep,litvak2011eeg} include de-noising, artifact removing, and normalizations. Specific procedures depend on the modality of medical signals. 

\textcolor{black}{Another sequential data form in the medical context is sequential events that collected in EHR. Although sequential events data could be technically merged into one meta table, with date stamps as an attribute, the tabular structure of the meta table fails to investigate the chronological order of events.} Some sequential event synthesis algorithms, particularly the simulation-based algorithms \cite{emerge2009,emerge2010,PADARSER,CorMESR} in section \ref{sec:simulation}, preserve the complex multi-table structure of EHR and the data structure is addressed as "patient care maps" or "careflow" in their algorithms. They would first synthesize several time points for each synthetic patient and then would add random tables for each time point. Each time point would have different numbers and types of tables, representing different events happened at this time point.

Other algorithms, however, normalize the events at each time point into a fix-size vector, because the fix-size input for each time point is required for algorithms, particularly the deep learning-based algorithms \cite{SynTEG,CorGAN,medGAN}. In these algorithms, the number of time points may vary, but the length of vectors at each time point must be fixed. Since all events contained in the datasets can span an event space, the fix-size vector is then a one-hot vector, and each entry of this vector represents the presence of the corresponding event at this specific time point. For example, considering an event space contains white cell abnormality, FVC abnormality, X-ray abnormality and medication usage, The fixed size vector [1, 1, 1, 0], indicates a presence of white cell count abnormality, FVC abnormality and X-ray abnormality, while no presence of medication usage. 

To standardize the event space and provide a standardized dictionary for different symptoms, many algorithms will first use International Classification of Diseases (ICD) format to describe the events, and then use the space of ICD codes to generate fix-size vectors for each time point. The ICD format \cite{ICD9} is a set of diagnostic encoding rules for clinical signs, symptoms, abnormal findings, complaints, social circumstances, and external causes of injury or diseases. Under this format, two versions, ICD9 \cite{ICD9} and ICD10 \cite{ICD10}, are commonly used in synthesis algorithms. By encoding each symptom into the ICD format, an attribute space containing all ICD codes in the dataset is constructed, and for each time stamp, a fix-size, one-hot encoded vector representing the coordinators in the ICD space is derived for each time stamp of EHR.

\subsection{Synthetic datasets for EHR}
 We also discovered three released synthetic datasets, including Vanderbilt Synthetic Derivative (SD) dataset, Data Entrepreneurs’ Synthetic Public Use File (DE-SynPUF) dataset and EMR Bots. These datasets, using different synthesis algorithms, provide hands-on practices for data synthesis. 
 
\subsubsection{Vanderbilt Synthetic Derivative (SD) dataset} The Vanderbilt SD dataset \cite{danciu2014vanderbiltSD} is a synthetic dataset derived from a real database containing over 2.2 million patients. In SD dataset, demographic information, ICD-9 codes (diagnoses), CPT procedure codes, medications, vital signs, registry, patient histories, and lab values is included. The dataset was de-identified by altering the records with their closest neighbors.

\subsubsection{CMS 2008-2010 Data Entrepreneurs’ Synthetic Public Use File (DE-SynPUF)} The DE-SynPUF dataset \cite{DE-SynPUF} is also a synthetic EHR dataset containing data from over 2 millions synthetic patients in five domains: Beneficiary Summary, which includes the demographic information and hospital enrollment reasons; Inpatient Claims, such as the presence of a surgery and other clinical measurements for patients admitted in hospitals; Outpatient Claims, which is the procedure of examinations happened outside of hospitals; Carrier Claims, which are derived of the bill information of all medical services, and include the name and date of the billed services, as well as the reimbursement amount related to this bill; Prescription Drug Events (PDE), which contains the medication information for each patient. 

To derive this large synthetic dataset, a combination of simulations and multiple imputation algorithms were used. Five steps of data synthesis were used: 1) variable reduction, where only clinical useful attributes were selected to be released; 2) suppression, where the rare data that had disclosure risk were removed; 3) substitution, where, similar as SD dataset \cite{danciu2014vanderbiltSD}, the attribute values for each patient were replaced by its nearest neighbors; 4) imputation, where the collected values of single variables were replaced by values synthesized from conditional distributions on key variables; 5) perturbation, where timelines of patient records were altered by changing dates; 6) coarsening, where the continuous variables were coarsened into discrete variables. 

\subsubsection{EMRBots} The EMRBots \cite{EMRBot} dataset is also an artificially generated EHR dataset contains three sub-datasets with 100, 10,000 and 100,000 synthetic patients. Unlike Vanderbilt SD and De-SynPUF dataset, the data from EMRBots are generated by a set of pre-defined criteria, which was set by an experienced clinician.



\section{Takeaway messages}\label{sec:discussion}
In this section, we will then turn to two limitations we discovered during the literature review and discuss on the fact that researchers are still facing the open issues regarding the utility of synthetic data. We will also provide some potential solutions towards these limitations in this section, and provide several potential research directions for non-imaging data synthesis. 

\subsection{A dilemma between data-driven and knowledge driven approaches}

Deep neural networks have been providing efficient tools in medical image synthesis. In the field of non-imaging data synthesis, it is also on the top of our recommendation list. Deep neural networks, such as CTGAN \cite{CTGAN}, medGAN \cite{medGAN} and temporal gans such as TimeGAN \cite{timegan}, are easy to implement, and they does not require any prior knowledge during the training. However, the data-driven nature of deep generative models makes these methods suffer from overfitting (or mode collapse in data synthesis field). Methods such as Wasserstein loss \cite{arjovsky2017wgan}, ease the pain of overfitting. Moreover, the deep generative models are notorious for their difficulty in interpretation. If there are any unreal output values, it is nearly impossible for ones to adjust accordingly in deep generative models. Attention mechanisms \cite{attention1,vaswani2017transformer} have been widely used to improve the network explainability, while they have been rarely discussed in the non-imaging data synthesis field. 

The EMERGE family \cite{emerge2009,PADARSER} and Bayesian Networks \cite{Privbayes} allow ones to bring experts' prior knowledge in data synthesis. However, both of these algorithms are not good at handling high-dimensional data (tabular data with >1000 attributes) \cite{Privbayes}. For EMERGE-based methods, building patients' caremap models from scratch is laborious and time-consuming. Synthea \cite{walonoski2018synthea} provides over 35 modules for different patient caremap modeling regarding different diseases, while for customized caremap modeling, customized caremap building is still difficult. \textcolor{black}{ A new type of data modeling named theory-driven modeling \cite{karpatne2017theory} is proposed to find a balance between knowledge-driven and data-driven methods. In addition, incorporating prior knowledge into deep generative networks \cite{raissi2019physics} shall be a potential solution towards the synthesis efficiency and network explainability. }

\subsection{A dilemma between data utility and data privacy}
In 2021, a comparative study \cite{stadler2021synthetic} was performed on different data synthesis algorithms, and the authors evaluate the risk of privacy violation and the utility of synthetic data from these algorithms. They discover that synthetic data does not protect the patients' privacy naturally. The synthetic data with high utility are vulnerable to the membership inference attack, thus have a high risk of individual re-identification. Meanwhile, they also perform a utility assessment on DP-based generative models and the results further support the dilemma between data utility and data privacy. 

This dilemma is also reported in many articles in this review, especially those with DP mechanism \cite{dpgan,Privbayes,chen2018dpvae}. With more noise added (lower $\epsilon$ in DP), the statistical closeness and prediction accuracy for downstream tasks get lower. Thus, it is important to re-think the rationale of ONLY using data synthesis to protect privacy during data release. In 2004, a paper \cite{abowd2004new} was proposed to combine three approaches for confidential data release: (1) synthetic data releasing; (2) real analytical data releasing and (3) restricted access to aforementioned data. This complex method combining techniques and policy might be a solution to this dilemma, yet research investigating this dilemma is still ongoing. 

\subsection{Beyond the debate over encoder-decoder: uncertainty and diversity}
\textcolor{black}{For tabular data, whose rows are individuals and columns are attributes, different deep learning architectures have been proposed. In section \ref{subsubsec:tabgan}, we elaborated a debate over whether or not to use encoder-decoder structures in tabular data synthesis. Some algorithms \cite{EMRGAN,dash2019healthGAN,SPRINTGAN} synthesize data directly, while others \cite{medGAN,ehrGAN} synthesize features in a latent space, and use a decoder to generate target data. Although in this review, we cannot determine the winner regarding to synthesis performance, but we can provide three perspectives which allows the readers to evaluate these two synthesis strategies. }

First is the uncertainty of the synthetic data. The target function of encoder-decoder is to reconstruct the input. No matter how well-trained the encoder-decoder is, the MSE loss between input and the reconstruction cannot reach 0. This additional noise increases the uncertainty of the synthetic data, and might introduce a shift in the synthetic data domain and the real data domain. 

Second is the diversity of the synthetic data. Mode collapse is a common GAN failure whereas the diversity of synthetic data is limited. Most deep learning generators just verify a posteriori probability on the diversity, such as the Wasserstein loss \cite{arjovsky2017wgan}. However, if the synthesis is latent space synthesis, the diversity of synthetic samples might not be easily controlled. 

Of course, the qualities of latent space synthesis can not be guaranteed due to its indirectness. However, this does not mean that the latent space synthesis is not worthy investigating at all: the energy-based models \cite{energymodel} introduces a new perspective for us to understand the data diversities in GAN models, and provides theoretical proofs for the diversities in the latent space synthesis. 

Considering a dataset $X$, the energy-based models approximate the distributions of $Pr(X)$ with a density function of $p(X)$ using Boltzmann model
\begin{equation}\label{eq:energyprob}
  p_\theta(X) = - \frac{e^{-E_\theta(X)}}{Z_\theta},
\end{equation}
where $Z_\theta = \int e^{-E_\theta(X)} dX$ is a normalizing constant. The generative process is to sampling from the distribution $p_\theta(X)$; however, sampling from $p_\theta(X)$ is a hard task \cite{energymodel}. Considering the sampling difficulty, a neural network $G$ is proposed to perform fast approximated sampling for the energy model. With the neural network, the model distribution $p_\theta$ in Eq. \ref{eq:energyprob} by the distributions of the output $p_G$. To achieve this goal, these two distributions must be close. Mathematically, we use the KL divergence to measure the distance between $p_\theta$ and $p_G$
\begin{equation}\label{eq:energykl}
KL(p_G(X)||p_\theta(X))= \int p_G(X)log\frac{p_G(X)}{p_\theta(X)}dX
 = - H_G(X)+ \mathbb{E}_{X\sim p_G(X)}[E_\theta(X)]+logZ_\theta.
\end{equation}
\textcolor{black}{Here, $H_G(X)= \int p_G(X)logp_G(X)dX$ is the entropy of the distribution $p_G(X)$, and the diversity of a GAN model maximize the $H_G(X)$ (i.e., minimize the $ - H_G(X)$ in Eq. \ref{eq:energykl}). However, the entropy approximation can be cumbersome. Thus, instead of calculating $H_G(X)$ directly during the optimization of generation models, \textbf{the introduction of encoder-decoder can estimate the variety $H_G(X)$ by approximating its lower bounds \cite{aeapproximation}}. The pre-trained encoders in AE+GAN models allow an approximation of $H_G(X)$ by the entropy $H(P(X)$ of the real reference data and this approximation is particularly guaranteed when synthetic data is derived from the real hidden feature maps, such as the architecture presented in ehrGAN \cite{ehrGAN}. }

\subsection{Other potential research directions}
In addition to investigating the solutions to two aforementioned dilemmas, we will provide other potential research directions for non-imaging data synthesis in this subsection. 

\subsubsection{New models and new metrics}
 GANs have been widely used in non-imaging data synthesis. Diffusion models, which synthesize data by reversing a Gaussian noising process, have proved to have a better synthesis performance than GANs in image synthesis \cite{dhariwal2021diffusion}, but have been scarcely investigated in non-imaging data synthesis. This new method and its applications on non-imaging data synthesis might be worth investigating. 

As for the metrics, we introduced differential privacy as an important metric in evaluating the privacy of synthetic data. In addition to differential privacy, there are also techniques analyzing the extent of privacy preserving quantitatively, including $k$-anonymity, $l$-diversity, $t$-closeness and $p$-indistinguishability \cite{attributeReview}. However, they are not commonly practiced in the data synthesis field, and implementations of these metrics have not been investigated yet.





\subsubsection{Multi-modality synthesis}
The concept of multi-modality \cite{multimodal} is widely used in medical image synthesis, whereas different image modalities such as CT, MRI and X-ray images are synthesized together for a comprehensive representation of patients. Since non-imaging medical data is also an important clinical modality, the hybrid synthesis of imaging and non-imaging data should be considered. However, many algorithms only investigate the independent synthesis of imaging data and non-imaging data and coupled synthesis is still scarce in the literature.




\section{Conclusion}\label{sec:conclusion}
\textcolor{black}{
The trustworthiness represents a set of essential qualities required for an AI algorithm: privacy, robustness, explainability, and fairness. In order to develop the trustworthy AI on non-imaging medical data, data synthesis algorithms have been proposed. By improving the number, variety and privacy of training samples, data synthesis algorithms are able to help AI models with a better accuracy at a lower cost. Trustworthy AI algorithms should cover all tasks in the AI field, including prediction and generation. However. most works so far concentrate on trustworthy “predictive modeling”, whereas the AI generation model raises many concerns as those exposed in this manuscript. Thus. this survey aims to be a referential point of discussion and a motivating catalyst of research around trustworthy synthetic data generation.}
\textcolor{black}{
In this paper, we identified three major types of non-imaging medical data synthesis algorithms and provided a comprehensive literature review about them and their evaluations. We also identified two challenges faced by all data synthesis algorithms: finding the balance between the utilization of data and knowledge and finding the balance between data utility and data privacy. We revealed some limitations existing in non-imaging data synthesis, and called for new architectures, new evaluation metrics and multi-modality strategies to drive future efforts in this exciting research area.}


\begin{acks}
This study was supported in part by the ERC IMI (101005122), the H2020 (952172), the MRC (MC/PC/21013), the Royal Society (IEC\textbackslash NSFC\textbackslash211235), the NVIDIA Academic Hardware Grant Program, the SABER project supported by Boehringer Ingelheim Ltd, and the UKRI Future Leaders Fellowship (MR/V023799/1). J. Del Ser also acknowledges support from the Department of Education of the Basque Government via the Consolidated Research Group MATHMODE (IT1456-22).
\end{acks}

\bibliographystyle{ACM-Reference-Format}
\bibliography{reference}


\begin{thebibliography}{169}


\ifx \showCODEN    \undefined \def \showCODEN     #1{\unskip}     \fi
\ifx \showDOI      \undefined \def \showDOI       #1{#1}\fi
\ifx \showISBNx    \undefined \def \showISBNx     #1{\unskip}     \fi
\ifx \showISBNxiii \undefined \def \showISBNxiii  #1{\unskip}     \fi
\ifx \showISSN     \undefined \def \showISSN      #1{\unskip}     \fi
\ifx \showLCCN     \undefined \def \showLCCN      #1{\unskip}     \fi
\ifx \shownote     \undefined \def \shownote      #1{#1}          \fi
\ifx \showarticletitle \undefined \def \showarticletitle #1{#1}   \fi
\ifx \showURL      \undefined \def \showURL       {\relax}        \fi
\providecommand\bibfield[2]{#2}
\providecommand\bibinfo[2]{#2}
\providecommand\natexlab[1]{#1}
\providecommand\showeprint[2][]{arXiv:#2}

\bibitem[Abadi et~al\mbox{.}(2016)]%
        {dpsgd}
\bibfield{author}{\bibinfo{person}{Martin Abadi}, \bibinfo{person}{Andy Chu},
  \bibinfo{person}{Ian Goodfellow}, \bibinfo{person}{H~Brendan McMahan},
  \bibinfo{person}{Ilya Mironov}, \bibinfo{person}{Kunal Talwar}, {and}
  \bibinfo{person}{Li Zhang}.} \bibinfo{year}{2016}\natexlab{}.
\newblock \showarticletitle{Deep learning with differential privacy}. In
  \bibinfo{booktitle}{\emph{Proceedings of the 2016 ACM SIGSAC conference on
  computer and communications security}}. \bibinfo{publisher}{Association for
  Computing Machinery}, \bibinfo{address}{Vienna, Austria},
  \bibinfo{pages}{308--318}.
\newblock


\bibitem[Abowd and Lane(2004)]%
        {abowd2004new}
\bibfield{author}{\bibinfo{person}{John~M. Abowd} {and} \bibinfo{person}{Julia
  Lane}.} \bibinfo{year}{2004}\natexlab{}.
\newblock \showarticletitle{New Approaches to Confidentiality Protection:
  Synthetic Data, Remote Access and Research Data Centers}. In
  \bibinfo{booktitle}{\emph{Privacy in Statistical Databases}},
  \bibfield{editor}{\bibinfo{person}{Josep Domingo-Ferrer} {and}
  \bibinfo{person}{Vicen{\c{c}} Torra}} (Eds.). \bibinfo{publisher}{Springer
  Berlin Heidelberg}, \bibinfo{address}{Berlin, Heidelberg},
  \bibinfo{pages}{282--289}.
\newblock
\showISBNx{978-3-540-25955-8}


\bibitem[Afshin-Pour et~al\mbox{.}(2011)]%
        {afshin2011misimulation}
\bibfield{author}{\bibinfo{person}{Babak Afshin-Pour}, \bibinfo{person}{Hamid
  Soltanian-Zadeh}, \bibinfo{person}{Gholam-Ali Hossein-Zadeh},
  \bibinfo{person}{Cheryl~L Grady}, {and} \bibinfo{person}{Stephen~C
  Strother}.} \bibinfo{year}{2011}\natexlab{}.
\newblock \showarticletitle{A mutual information-based metric for evaluation of
  fMRI data-processing approaches}.
\newblock \bibinfo{journal}{\emph{Human brain mapping}} \bibinfo{volume}{32},
  \bibinfo{number}{5} (\bibinfo{year}{2011}), \bibinfo{pages}{699--715}.
\newblock


\bibitem[Arjovsky et~al\mbox{.}(2017)]%
        {arjovsky2017wgan}
\bibfield{author}{\bibinfo{person}{Martin Arjovsky}, \bibinfo{person}{Soumith
  Chintala}, {and} \bibinfo{person}{L{\'e}on Bottou}.}
  \bibinfo{year}{2017}\natexlab{}.
\newblock \showarticletitle{{W}asserstein Generative Adversarial Networks}. In
  \bibinfo{booktitle}{\emph{Proceedings of the 34th International Conference on
  Machine Learning}} \emph{(\bibinfo{series}{Proceedings of Machine Learning
  Research}, Vol.~\bibinfo{volume}{70})},
  \bibfield{editor}{\bibinfo{person}{Doina Precup} {and}
  \bibinfo{person}{Yee~Whye Teh}} (Eds.). \bibinfo{publisher}{PMLR},
  \bibinfo{address}{Sydney, Australia}, \bibinfo{pages}{214--223}.
\newblock
\urldef\tempurl%
\url{https://proceedings.mlr.press/v70/arjovsky17a.html}
\showURL{%
\tempurl}


\bibitem[Atreya et~al\mbox{.}(2013)]%
        {CriticEmerge}
\bibfield{author}{\bibinfo{person}{Ravi~V Atreya}, \bibinfo{person}{Joshua~C
  Smith}, \bibinfo{person}{Allison~B McCoy}, \bibinfo{person}{Bradley Malin},
  {and} \bibinfo{person}{Randolph~A Miller}.} \bibinfo{year}{2013}\natexlab{}.
\newblock \showarticletitle{Reducing patient re-identification risk for
  laboratory results within research datasets}.
\newblock \bibinfo{journal}{\emph{Journal of the American Medical Informatics
  Association}} \bibinfo{volume}{20}, \bibinfo{number}{1}
  (\bibinfo{year}{2013}), \bibinfo{pages}{95--101}.
\newblock
\showISSN{1527-974X}


\bibitem[Ayala-Rivera et~al\mbox{.}(2016)]%
        {cocoa}
\bibfield{author}{\bibinfo{person}{Vanessa Ayala-Rivera},
  \bibinfo{person}{A.~Omar Portillo-Dominguez}, \bibinfo{person}{Liam Murphy},
  {and} \bibinfo{person}{Christina Thorpe}.} \bibinfo{year}{2016}\natexlab{}.
\newblock \showarticletitle{COCOA: A Synthetic Data Generator for Testing
  Anonymization Techniques}. In \bibinfo{booktitle}{\emph{Privacy in
  Statistical Databases}}, \bibfield{editor}{\bibinfo{person}{Josep
  Domingo-Ferrer} {and} \bibinfo{person}{Mirjana Peji{\'{c}}-Bach}} (Eds.).
  \bibinfo{publisher}{Springer International Publishing},
  \bibinfo{address}{Cham, Switzerland}, \bibinfo{pages}{163--177}.
\newblock
\showISBNx{978-3-319-45381-1}


\bibitem[Bahdanau et~al\mbox{.}(2014)]%
        {attention1}
\bibfield{author}{\bibinfo{person}{Dzmitry Bahdanau},
  \bibinfo{person}{Kyunghyun Cho}, {and} \bibinfo{person}{Yoshua Bengio}.}
  \bibinfo{year}{2014}\natexlab{}.
\newblock \bibinfo{title}{Neural Machine Translation by Jointly Learning to
  Align and Translate}.
\newblock
\newblock
\urldef\tempurl%
\url{https://doi.org/10.48550/ARXIV.1409.0473}
\showDOI{\tempurl}


\bibitem[Barrett et~al\mbox{.}(2009)]%
        {medicalNetwork}
\bibfield{author}{\bibinfo{person}{Christopher~L. Barrett},
  \bibinfo{person}{Richard~J. Beckman}, \bibinfo{person}{Maleq Khan},
  \bibinfo{person}{V.~S.~Anil Kumar}, \bibinfo{person}{Madhav~V. Marathe},
  \bibinfo{person}{Paula~E. Stretz}, \bibinfo{person}{Tridib Dutta}, {and}
  \bibinfo{person}{Bryan Lewis}.} \bibinfo{year}{2009}\natexlab{}.
\newblock \showarticletitle{Generation and analysis of large synthetic social
  contact networks}. In \bibinfo{booktitle}{\emph{Proceedings of the 2009
  Winter Simulation Conference (WSC)}}. \bibinfo{publisher}{IEEE},
  \bibinfo{address}{Austin, Texas, USA}, \bibinfo{pages}{1003--1014}.
\newblock
\urldef\tempurl%
\url{https://doi.org/10.1109/WSC.2009.5429425}
\showDOI{\tempurl}


\bibitem[Barzegaran et~al\mbox{.}(2019)]%
        {barzegaran2019eegsourcesim}
\bibfield{author}{\bibinfo{person}{Elham Barzegaran},
  \bibinfo{person}{Sebastian Bosse}, \bibinfo{person}{Peter~J Kohler}, {and}
  \bibinfo{person}{Anthony~M Norcia}.} \bibinfo{year}{2019}\natexlab{}.
\newblock \showarticletitle{EEGSourceSim: A framework for realistic simulation
  of EEG scalp data using MRI-based forward models and biologically plausible
  signals and noise}.
\newblock \bibinfo{journal}{\emph{Journal of neuroscience methods}}
  \bibinfo{volume}{328} (\bibinfo{year}{2019}), \bibinfo{pages}{108377}.
\newblock


\bibitem[Beaulieu-Jones et~al\mbox{.}(2017)]%
        {imputationAE2017}
\bibfield{author}{\bibinfo{person}{Brett~K Beaulieu-Jones},
  \bibinfo{person}{Jason~H Moore}, {and} \bibinfo{person}{POOLED RESOURCE
  OPEN-ACCESS ALS CLINICAL~TRIALS CONSORTIUM}.}
  \bibinfo{year}{2017}\natexlab{}.
\newblock \showarticletitle{Missing data imputation in the electronic health
  record using deeply learned autoencoders}. In
  \bibinfo{booktitle}{\emph{Pacific symposium on biocomputing 2017}}.
  \bibinfo{publisher}{World Scientific}, \bibinfo{address}{Hawaii, USA},
  \bibinfo{pages}{207--218}.
\newblock


\bibitem[Beaulieu-Jones et~al\mbox{.}(2019)]%
        {SPRINTGAN}
\bibfield{author}{\bibinfo{person}{Brett~K Beaulieu-Jones},
  \bibinfo{person}{Zhiwei~Steven Wu}, \bibinfo{person}{Chris Williams},
  \bibinfo{person}{Ran Lee}, \bibinfo{person}{Sanjeev~P Bhavnani},
  \bibinfo{person}{James~Brian Byrd}, {and} \bibinfo{person}{Casey~S Greene}.}
  \bibinfo{year}{2019}\natexlab{}.
\newblock \showarticletitle{Privacy-preserving generative deep neural networks
  support clinical data sharing}.
\newblock \bibinfo{journal}{\emph{Circulation: Cardiovascular Quality and
  Outcomes}} \bibinfo{volume}{12}, \bibinfo{number}{7} (\bibinfo{year}{2019}),
  \bibinfo{pages}{e005122}.
\newblock
\showISSN{1941-7705}


\bibitem[Bigdely-Shamlo et~al\mbox{.}(2015)]%
        {bigdely2015prep}
\bibfield{author}{\bibinfo{person}{Nima Bigdely-Shamlo}, \bibinfo{person}{Tim
  Mullen}, \bibinfo{person}{Christian Kothe}, \bibinfo{person}{Kyung-Min Su},
  {and} \bibinfo{person}{Kay~A Robbins}.} \bibinfo{year}{2015}\natexlab{}.
\newblock \showarticletitle{The PREP pipeline: standardized preprocessing for
  large-scale EEG analysis}.
\newblock \bibinfo{journal}{\emph{Frontiers in neuroinformatics}}
  \bibinfo{volume}{9} (\bibinfo{year}{2015}), \bibinfo{pages}{16}.
\newblock


\bibitem[Biswal et~al\mbox{.}(2021)]%
        {eva}
\bibfield{author}{\bibinfo{person}{Siddharth Biswal}, \bibinfo{person}{Soumya
  Ghosh}, \bibinfo{person}{Jon Duke}, \bibinfo{person}{Bradley Malin},
  \bibinfo{person}{Walter Stewart}, \bibinfo{person}{Cao Xiao}, {and}
  \bibinfo{person}{Jimeng Sun}.} \bibinfo{year}{2021}\natexlab{}.
\newblock \showarticletitle{EVA: Generating Longitudinal Electronic Health
  Records Using Conditional Variational Autoencoders}. In
  \bibinfo{booktitle}{\emph{Proceedings of the 6th Machine Learning for
  Healthcare Conference}} \emph{(\bibinfo{series}{Proceedings of Machine
  Learning Research}, Vol.~\bibinfo{volume}{149})},
  \bibfield{editor}{\bibinfo{person}{Ken Jung}, \bibinfo{person}{Serena Yeung},
  \bibinfo{person}{Mark Sendak}, \bibinfo{person}{Michael Sjoding}, {and}
  \bibinfo{person}{Rajesh Ranganath}} (Eds.). \bibinfo{publisher}{PMLR},
  \bibinfo{address}{Virtual}, \bibinfo{pages}{260--282}.
\newblock
\urldef\tempurl%
\url{https://proceedings.mlr.press/v149/biswal21a.html}
\showURL{%
\tempurl}


\bibitem[Buczak et~al\mbox{.}(2010)]%
        {emerge2010}
\bibfield{author}{\bibinfo{person}{Anna~L Buczak}, \bibinfo{person}{Steven
  Babin}, {and} \bibinfo{person}{Linda Moniz}.}
  \bibinfo{year}{2010}\natexlab{}.
\newblock \showarticletitle{Data-driven approach for creating synthetic
  electronic medical records}.
\newblock \bibinfo{journal}{\emph{BMC medical informatics and decision making}}
  \bibinfo{volume}{10}, \bibinfo{number}{1} (\bibinfo{year}{2010}),
  \bibinfo{pages}{1--28}.
\newblock
\showISSN{1472-6947}


\bibitem[Burridge(2003)]%
        {IPSO}
\bibfield{author}{\bibinfo{person}{Jim Burridge}.}
  \bibinfo{year}{2003}\natexlab{}.
\newblock \showarticletitle{Information preserving statistical obfuscation}.
\newblock \bibinfo{journal}{\emph{Statistics and Computing}}
  \bibinfo{volume}{13}, \bibinfo{number}{4} (\bibinfo{year}{2003}),
  \bibinfo{pages}{321--327}.
\newblock
\showISSN{1573-1375}


\bibitem[Caiola and Reiter(2010)]%
        {caiola2010random}
\bibfield{author}{\bibinfo{person}{Gregory Caiola} {and}
  \bibinfo{person}{Jerome~P Reiter}.} \bibinfo{year}{2010}\natexlab{}.
\newblock \showarticletitle{Random Forests for Generating Partially Synthetic,
  Categorical Data.}
\newblock \bibinfo{journal}{\emph{Trans. Data Priv.}} \bibinfo{volume}{3},
  \bibinfo{number}{1} (\bibinfo{year}{2010}), \bibinfo{pages}{27--42}.
\newblock


\bibitem[Camino et~al\mbox{.}(2019)]%
        {improvedgain}
\bibfield{author}{\bibinfo{person}{Ramiro~D. Camino},
  \bibinfo{person}{Christian~A. Hammerschmidt}, {and} \bibinfo{person}{Radu
  State}.} \bibinfo{year}{2019}\natexlab{}.
\newblock \bibinfo{title}{Improving Missing Data Imputation with Deep
  Generative Models}.
\newblock
\newblock
\urldef\tempurl%
\url{https://doi.org/10.48550/ARXIV.1902.10666}
\showDOI{\tempurl}


\bibitem[Cano and Torra(2009)]%
        {fuzzyMI}
\bibfield{author}{\bibinfo{person}{Isaac Cano} {and} \bibinfo{person}{Vicenç
  Torra}.} \bibinfo{year}{2009}\natexlab{}.
\newblock \showarticletitle{Generation of synthetic data by means of fuzzy
  c-Regression}. In \bibinfo{booktitle}{\emph{2009 IEEE International
  Conference on Fuzzy Systems}}. \bibinfo{publisher}{IEEE},
  \bibinfo{address}{Jeju Island, Korea}, \bibinfo{pages}{1145--1150}.
\newblock
\showISBNx{142443596X}


\bibitem[Cao et~al\mbox{.}(2013)]%
        {INOS}
\bibfield{author}{\bibinfo{person}{Hong Cao}, \bibinfo{person}{Xiao-Li Li},
  \bibinfo{person}{David Yew-Kwong Woon}, {and} \bibinfo{person}{See-Kiong
  Ng}.} \bibinfo{year}{2013}\natexlab{}.
\newblock \showarticletitle{Integrated oversampling for imbalanced time series
  classification}.
\newblock \bibinfo{journal}{\emph{IEEE Transactions on Knowledge and Data
  Engineering}} \bibinfo{volume}{25}, \bibinfo{number}{12}
  (\bibinfo{year}{2013}), \bibinfo{pages}{2809--2822}.
\newblock
\showISSN{1041-4347}


\bibitem[Chawla et~al\mbox{.}(2002)]%
        {Smote}
\bibfield{author}{\bibinfo{person}{Nitesh~V Chawla}, \bibinfo{person}{Kevin~W
  Bowyer}, \bibinfo{person}{Lawrence~O Hall}, {and} \bibinfo{person}{W~Philip
  Kegelmeyer}.} \bibinfo{year}{2002}\natexlab{}.
\newblock \showarticletitle{SMOTE: synthetic minority over-sampling technique}.
\newblock \bibinfo{journal}{\emph{Journal of artificial intelligence research}}
   \bibinfo{volume}{16} (\bibinfo{year}{2002}), \bibinfo{pages}{321--357}.
\newblock
\showISSN{1076-9757}


\bibitem[Che et~al\mbox{.}(2017)]%
        {ehrGAN}
\bibfield{author}{\bibinfo{person}{Zhengping Che}, \bibinfo{person}{Yu Cheng},
  \bibinfo{person}{Shuangfei Zhai}, \bibinfo{person}{Zhaonan Sun}, {and}
  \bibinfo{person}{Yan Liu}.} \bibinfo{year}{2017}\natexlab{}.
\newblock \showarticletitle{Boosting Deep Learning Risk Prediction with
  Generative Adversarial Networks for Electronic Health Records}. In
  \bibinfo{booktitle}{\emph{2017 IEEE International Conference on Data Mining
  (ICDM)}}. \bibinfo{publisher}{IEEE}, \bibinfo{address}{New Orleans, LA, USA},
  \bibinfo{pages}{787--792}.
\newblock
\urldef\tempurl%
\url{https://doi.org/10.1109/ICDM.2017.93}
\showDOI{\tempurl}


\bibitem[Chen et~al\mbox{.}(2020)]%
        {chen2020ganleaks}
\bibfield{author}{\bibinfo{person}{Dingfan Chen}, \bibinfo{person}{Ning Yu},
  \bibinfo{person}{Yang Zhang}, {and} \bibinfo{person}{Mario Fritz}.}
  \bibinfo{year}{2020}\natexlab{}.
\newblock \showarticletitle{GAN-Leaks: A Taxonomy of Membership Inference
  Attacks against Generative Models}. In \bibinfo{booktitle}{\emph{Proceedings
  of the 2020 ACM SIGSAC Conference on Computer and Communications Security}}
  (Virtual) \emph{(\bibinfo{series}{CCS '20})}. \bibinfo{publisher}{Association
  for Computing Machinery}, \bibinfo{address}{New York, NY, USA},
  \bibinfo{pages}{343–362}.
\newblock
\showISBNx{9781450370899}
\urldef\tempurl%
\url{https://doi.org/10.1145/3372297.3417238}
\showDOI{\tempurl}


\bibitem[Chen et~al\mbox{.}(2018)]%
        {chen2018dpvae}
\bibfield{author}{\bibinfo{person}{Qingrong Chen}, \bibinfo{person}{Chong
  Xiang}, \bibinfo{person}{Minhui Xue}, \bibinfo{person}{Bo Li},
  \bibinfo{person}{Nikita Borisov}, \bibinfo{person}{Dali Kaarfar}, {and}
  \bibinfo{person}{Haojin Zhu}.} \bibinfo{year}{2018}\natexlab{}.
\newblock \bibinfo{title}{Differentially Private Data Generative Models}.
\newblock
\newblock
\urldef\tempurl%
\url{https://doi.org/10.48550/ARXIV.1812.02274}
\showDOI{\tempurl}


\bibitem[Chen et~al\mbox{.}(2016)]%
        {infogan}
\bibfield{author}{\bibinfo{person}{Xi Chen}, \bibinfo{person}{Yan Duan},
  \bibinfo{person}{Rein Houthooft}, \bibinfo{person}{John Schulman},
  \bibinfo{person}{Ilya Sutskever}, {and} \bibinfo{person}{Pieter Abbeel}.}
  \bibinfo{year}{2016}\natexlab{}.
\newblock \showarticletitle{InfoGAN: Interpretable Representation Learning by
  Information Maximizing Generative Adversarial Nets}. In
  \bibinfo{booktitle}{\emph{Advances in Neural Information Processing
  Systems}}, \bibfield{editor}{\bibinfo{person}{D.~Lee},
  \bibinfo{person}{M.~Sugiyama}, \bibinfo{person}{U.~Luxburg},
  \bibinfo{person}{I.~Guyon}, {and} \bibinfo{person}{R.~Garnett}} (Eds.),
  Vol.~\bibinfo{volume}{29}. \bibinfo{publisher}{Curran Associates, Inc.},
  \bibinfo{address}{Barcelona, Spain}.
\newblock
\urldef\tempurl%
\url{https://proceedings.neurips.cc/paper/2016/file/7c9d0b1f96aebd7b5eca8c3edaa19ebb-Paper.pdf}
\showURL{%
\tempurl}


\bibitem[Cho et~al\mbox{.}(2014)]%
        {cho2014gru}
\bibfield{author}{\bibinfo{person}{Kyunghyun Cho}, \bibinfo{person}{Bart van
  Merrienboer}, \bibinfo{person}{Dzmitry Bahdanau}, {and}
  \bibinfo{person}{Yoshua Bengio}.} \bibinfo{year}{2014}\natexlab{}.
\newblock \bibinfo{title}{On the Properties of Neural Machine Translation:
  Encoder-Decoder Approaches}.
\newblock
\newblock
\urldef\tempurl%
\url{https://doi.org/10.48550/ARXIV.1409.1259}
\showDOI{\tempurl}


\bibitem[Choi et~al\mbox{.}(2017)]%
        {medGAN}
\bibfield{author}{\bibinfo{person}{Edward Choi}, \bibinfo{person}{Siddharth
  Biswal}, \bibinfo{person}{Bradley Malin}, \bibinfo{person}{Jon Duke},
  \bibinfo{person}{Walter~F. Stewart}, {and} \bibinfo{person}{Jimeng Sun}.}
  \bibinfo{year}{2017}\natexlab{}.
\newblock \showarticletitle{Generating Multi-label Discrete Patient Records
  using Generative Adversarial Networks}. In
  \bibinfo{booktitle}{\emph{Proceedings of the 2nd Machine Learning for
  Healthcare Conference}} \emph{(\bibinfo{series}{Proceedings of Machine
  Learning Research}, Vol.~\bibinfo{volume}{68})},
  \bibfield{editor}{\bibinfo{person}{Finale Doshi-Velez}, \bibinfo{person}{Jim
  Fackler}, \bibinfo{person}{David Kale}, \bibinfo{person}{Rajesh Ranganath},
  \bibinfo{person}{Byron Wallace}, {and} \bibinfo{person}{Jenna Wiens}} (Eds.).
  \bibinfo{publisher}{PMLR}, \bibinfo{address}{Boston, Massachusetts, USA},
  \bibinfo{pages}{286--305}.
\newblock
\urldef\tempurl%
\url{https://proceedings.mlr.press/v68/choi17a.html}
\showURL{%
\tempurl}


\bibitem[Danciu et~al\mbox{.}(2014)]%
        {danciu2014vanderbiltSD}
\bibfield{author}{\bibinfo{person}{Ioana Danciu}, \bibinfo{person}{James~D
  Cowan}, \bibinfo{person}{Melissa Basford}, \bibinfo{person}{Xiaoming Wang},
  \bibinfo{person}{Alexander Saip}, \bibinfo{person}{Susan Osgood},
  \bibinfo{person}{Jana Shirey-Rice}, \bibinfo{person}{Jacqueline Kirby}, {and}
  \bibinfo{person}{Paul~A Harris}.} \bibinfo{year}{2014}\natexlab{}.
\newblock \showarticletitle{Secondary use of clinical data: the Vanderbilt
  approach}.
\newblock \bibinfo{journal}{\emph{Journal of biomedical informatics}}
  \bibinfo{volume}{52} (\bibinfo{year}{2014}), \bibinfo{pages}{28--35}.
\newblock


\bibitem[Deeva et~al\mbox{.}(2020)]%
        {deeva2020bayesian}
\bibfield{author}{\bibinfo{person}{Irina Deeva}, \bibinfo{person}{Petr~D.
  Andriushchenko}, \bibinfo{person}{Anna~V. Kalyuzhnaya}, {and}
  \bibinfo{person}{Alexander~V. Boukhanovsky}.}
  \bibinfo{year}{2020}\natexlab{}.
\newblock \showarticletitle{Bayesian Networks-Based Personal Data Synthesis}.
  In \bibinfo{booktitle}{\emph{Proceedings of the 6th EAI International
  Conference on Smart Objects and Technologies for Social Good}} (Antwerp,
  Belgium) \emph{(\bibinfo{series}{GoodTechs '20})}.
  \bibinfo{publisher}{Association for Computing Machinery},
  \bibinfo{address}{New York, NY, USA}, \bibinfo{pages}{6–11}.
\newblock
\showISBNx{9781450375597}
\urldef\tempurl%
\url{https://doi.org/10.1145/3411170.3411243}
\showDOI{\tempurl}


\bibitem[Delaney et~al\mbox{.}(2019)]%
        {ECG-GAN}
\bibfield{author}{\bibinfo{person}{Anne~Marie Delaney}, \bibinfo{person}{Eoin
  Brophy}, {and} \bibinfo{person}{Tomas~E. Ward}.}
  \bibinfo{year}{2019}\natexlab{}.
\newblock \bibinfo{title}{Synthesis of Realistic ECG using Generative
  Adversarial Networks}.
\newblock
\newblock
\urldef\tempurl%
\url{https://doi.org/10.48550/ARXIV.1909.09150}
\showDOI{\tempurl}


\bibitem[Demir and Unal(2018)]%
        {patchgan}
\bibfield{author}{\bibinfo{person}{Ugur Demir} {and} \bibinfo{person}{Gozde
  Unal}.} \bibinfo{year}{2018}\natexlab{}.
\newblock \bibinfo{title}{Patch-Based Image Inpainting with Generative
  Adversarial Networks}.
\newblock
\newblock
\urldef\tempurl%
\url{https://doi.org/10.48550/ARXIV.1803.07422}
\showDOI{\tempurl}


\bibitem[Dhariwal and Nichol(2021)]%
        {dhariwal2021diffusion}
\bibfield{author}{\bibinfo{person}{Prafulla Dhariwal} {and}
  \bibinfo{person}{Alexander Nichol}.} \bibinfo{year}{2021}\natexlab{}.
\newblock \showarticletitle{Diffusion models beat gans on image synthesis}.
\newblock \bibinfo{journal}{\emph{Advances in Neural Information Processing
  Systems}}  \bibinfo{volume}{34} (\bibinfo{year}{2021}),
  \bibinfo{pages}{8780--8794}.
\newblock


\bibitem[Dong et~al\mbox{.}(2022)]%
        {SAGAN}
\bibfield{author}{\bibinfo{person}{Yongfeng Dong}, \bibinfo{person}{Huaxin
  Xiao}, {and} \bibinfo{person}{Yao Dong}.} \bibinfo{year}{2022}\natexlab{}.
\newblock \showarticletitle{SA-CGAN: An oversampling method based on single
  attribute guided conditional GAN for multi-class imbalanced learning}.
\newblock \bibinfo{journal}{\emph{Neurocomputing}}  \bibinfo{volume}{472}
  (\bibinfo{year}{2022}), \bibinfo{pages}{326--337}.
\newblock
\showISSN{0925-2312}


\bibitem[Drechsler(2010)]%
        {MISVM}
\bibfield{author}{\bibinfo{person}{J{\"o}rg Drechsler}.}
  \bibinfo{year}{2010}\natexlab{}.
\newblock \showarticletitle{Using Support Vector Machines for Generating
  Synthetic Datasets}. In \bibinfo{booktitle}{\emph{Privacy in Statistical
  Databases}}, \bibfield{editor}{\bibinfo{person}{Josep Domingo-Ferrer} {and}
  \bibinfo{person}{Emmanouil Magkos}} (Eds.). \bibinfo{publisher}{Springer
  Berlin Heidelberg}, \bibinfo{address}{Berlin, Heidelberg},
  \bibinfo{pages}{148--161}.
\newblock
\showISBNx{978-3-642-15838-4}


\bibitem[Dua and Graff(2017)]%
        {uci2019}
\bibfield{author}{\bibinfo{person}{Dheeru Dua} {and} \bibinfo{person}{Casey
  Graff}.} \bibinfo{year}{2017}\natexlab{}.
\newblock \bibinfo{title}{{UCI} Machine Learning Repository}.
\newblock
\newblock
\urldef\tempurl%
\url{http://archive.ics.uci.edu/ml}
\showURL{%
\tempurl}


\bibitem[Dube and Gallagher(2014)]%
        {PADARSER}
\bibfield{author}{\bibinfo{person}{Kudakwashe Dube} {and}
  \bibinfo{person}{Thomas Gallagher}.} \bibinfo{year}{2014}\natexlab{}.
\newblock \showarticletitle{Approach and Method for Generating Realistic
  Synthetic Electronic Healthcare Records for Secondary Use}. In
  \bibinfo{booktitle}{\emph{Foundations of Health Information Engineering and
  Systems}}, \bibfield{editor}{\bibinfo{person}{Jeremy Gibbons} {and}
  \bibinfo{person}{Wendy MacCaull}} (Eds.). \bibinfo{publisher}{Springer Berlin
  Heidelberg}, \bibinfo{address}{Berlin, Heidelberg}, \bibinfo{pages}{69--86}.
\newblock
\showISBNx{978-3-642-53956-5}


\bibitem[Dwork et~al\mbox{.}(2006)]%
        {laplaciandp}
\bibfield{author}{\bibinfo{person}{Cynthia Dwork}, \bibinfo{person}{Frank
  McSherry}, \bibinfo{person}{Kobbi Nissim}, {and} \bibinfo{person}{Adam
  Smith}.} \bibinfo{year}{2006}\natexlab{}.
\newblock \showarticletitle{Calibrating Noise to Sensitivity in Private Data
  Analysis}. In \bibinfo{booktitle}{\emph{Theory of Cryptography}},
  \bibfield{editor}{\bibinfo{person}{Shai Halevi} {and} \bibinfo{person}{Tal
  Rabin}} (Eds.). \bibinfo{publisher}{Springer Berlin Heidelberg},
  \bibinfo{address}{Berlin, Heidelberg}, \bibinfo{pages}{265--284}.
\newblock
\showISBNx{978-3-540-32732-5}


\bibitem[Dwork and Roth(2014)]%
        {dp}
\bibfield{author}{\bibinfo{person}{Cynthia Dwork} {and} \bibinfo{person}{Aaron
  Roth}.} \bibinfo{year}{2014}\natexlab{}.
\newblock \showarticletitle{The algorithmic foundations of differential
  privacy}.
\newblock \bibinfo{journal}{\emph{Foundations and Trends® in Theoretical
  Computer Science}} \bibinfo{volume}{9}, \bibinfo{number}{3–4}
  (\bibinfo{year}{2014}), \bibinfo{pages}{211--407}.
\newblock
\showISSN{1551-305X}


\bibitem[Engelmann and Lessmann(2021)]%
        {cwganbasedoversampling}
\bibfield{author}{\bibinfo{person}{Justin Engelmann} {and}
  \bibinfo{person}{Stefan Lessmann}.} \bibinfo{year}{2021}\natexlab{}.
\newblock \showarticletitle{Conditional Wasserstein GAN-based oversampling of
  tabular data for imbalanced learning}.
\newblock \bibinfo{journal}{\emph{Expert Systems with Applications}}
  \bibinfo{volume}{174} (\bibinfo{year}{2021}), \bibinfo{pages}{114582}.
\newblock
\showISSN{0957-4174}


\bibitem[Erhardt et~al\mbox{.}(2012)]%
        {erhardt2012simtb}
\bibfield{author}{\bibinfo{person}{Erik~B Erhardt}, \bibinfo{person}{Elena~A
  Allen}, \bibinfo{person}{Yonghua Wei}, \bibinfo{person}{Tom Eichele}, {and}
  \bibinfo{person}{Vince~D Calhoun}.} \bibinfo{year}{2012}\natexlab{}.
\newblock \showarticletitle{SimTB, a simulation toolbox for fMRI data under a
  model of spatiotemporal separability}.
\newblock \bibinfo{journal}{\emph{Neuroimage}} \bibinfo{volume}{59},
  \bibinfo{number}{4} (\bibinfo{year}{2012}), \bibinfo{pages}{4160--4167}.
\newblock


\bibitem[Esteban et~al\mbox{.}(2017)]%
        {rcgan}
\bibfield{author}{\bibinfo{person}{Cristóbal Esteban},
  \bibinfo{person}{Stephanie~L. Hyland}, {and} \bibinfo{person}{Gunnar
  Rätsch}.} \bibinfo{year}{2017}\natexlab{}.
\newblock \bibinfo{title}{Real-valued (Medical) Time Series Generation with
  Recurrent Conditional GANs}.
\newblock
\newblock
\urldef\tempurl%
\url{https://doi.org/10.48550/ARXIV.1706.02633}
\showDOI{\tempurl}


\bibitem[Feldman et~al\mbox{.}(2015)]%
        {epsilonFairness}
\bibfield{author}{\bibinfo{person}{Michael Feldman},
  \bibinfo{person}{Sorelle~A. Friedler}, \bibinfo{person}{John Moeller},
  \bibinfo{person}{Carlos Scheidegger}, {and} \bibinfo{person}{Suresh
  Venkatasubramanian}.} \bibinfo{year}{2015}\natexlab{}.
\newblock \showarticletitle{Certifying and Removing Disparate Impact}. In
  \bibinfo{booktitle}{\emph{Proceedings of the 21th ACM SIGKDD International
  Conference on Knowledge Discovery and Data Mining}} (Sydney, NSW, Australia)
  \emph{(\bibinfo{series}{KDD '15})}. \bibinfo{publisher}{Association for
  Computing Machinery}, \bibinfo{address}{New York, NY, USA},
  \bibinfo{pages}{259–268}.
\newblock
\showISBNx{9781450336642}
\urldef\tempurl%
\url{https://doi.org/10.1145/2783258.2783311}
\showDOI{\tempurl}


\bibitem[Fernandes et~al\mbox{.}(2017)]%
        {kagglecervical}
\bibfield{author}{\bibinfo{person}{Kelwin Fernandes}, \bibinfo{person}{Jaime~S.
  Cardoso}, {and} \bibinfo{person}{Jessica Fernandes}.}
  \bibinfo{year}{2017}\natexlab{}.
\newblock \showarticletitle{Transfer Learning with Partial Observability
  Applied to Cervical Cancer Screening}. In \bibinfo{booktitle}{\emph{Pattern
  Recognition and Image Analysis}},
  \bibfield{editor}{\bibinfo{person}{Lu{\'i}s~A. Alexandre},
  \bibinfo{person}{Jos{\'e} Salvador~S{\'a}nchez}, {and}
  \bibinfo{person}{Jo{\~a}o M.~F. Rodrigues}} (Eds.).
  \bibinfo{publisher}{Springer International Publishing},
  \bibinfo{address}{Cham}, \bibinfo{pages}{243--250}.
\newblock
\showISBNx{978-3-319-58838-4}


\bibitem[Fernández et~al\mbox{.}(2018)]%
        {SmoteReview}
\bibfield{author}{\bibinfo{person}{Alberto Fernández},
  \bibinfo{person}{Salvador Garcia}, \bibinfo{person}{Francisco Herrera}, {and}
  \bibinfo{person}{Nitesh~V Chawla}.} \bibinfo{year}{2018}\natexlab{}.
\newblock \showarticletitle{SMOTE for learning from imbalanced data: progress
  and challenges, marking the 15-year anniversary}.
\newblock \bibinfo{journal}{\emph{Journal of artificial intelligence research}}
   \bibinfo{volume}{61} (\bibinfo{year}{2018}), \bibinfo{pages}{863--905}.
\newblock
\showISSN{1076-9757}


\bibitem[for Health Statistics~(US) and on~Clinical~Classifications(1980)]%
        {ICD9}
\bibfield{author}{\bibinfo{person}{National~Center for Health Statistics~(US)}
  {and} \bibinfo{person}{Council on Clinical~Classifications}.}
  \bibinfo{year}{1980}\natexlab{}.
\newblock \bibinfo{booktitle}{\emph{The International classification of
  diseases, 9th revision, clinical modification: ICD-9-CM}}.
  Vol.~\bibinfo{volume}{2}.
\newblock \bibinfo{publisher}{US Department of Health and Human Services,
  Public Health Service, Health Care Financing Administration},
  \bibinfo{address}{USA}.
\newblock


\bibitem[for Medicare \& Medicaid~Services(2021)]%
        {EHRofficial2}
\bibfield{author}{\bibinfo{person}{Centers for Medicare \& Medicaid~Services}.}
  \bibinfo{year}{2021}\natexlab{}.
\newblock \bibinfo{title}{Electronic Health Records}.
\newblock
\newblock
\urldef\tempurl%
\url{https://www.cms.gov/Medicare/E-Health/EHealthRecords}
\showURL{%
\tempurl}


\bibitem[Goncalves et~al\mbox{.}(2020a)]%
        {ehrReview2}
\bibfield{author}{\bibinfo{person}{Andre Goncalves}, \bibinfo{person}{Priyadip
  Ray}, \bibinfo{person}{Braden Soper}, \bibinfo{person}{Jennifer Stevens},
  \bibinfo{person}{Linda Coyle}, {and} \bibinfo{person}{Ana~Paula Sales}.}
  \bibinfo{year}{2020}\natexlab{a}.
\newblock \showarticletitle{Generation and evaluation of synthetic patient
  data}.
\newblock \bibinfo{journal}{\emph{BMC medical research methodology}}
  \bibinfo{volume}{20}, \bibinfo{number}{1} (\bibinfo{year}{2020}),
  \bibinfo{pages}{1--40}.
\newblock


\bibitem[Goncalves et~al\mbox{.}(2020b)]%
        {goncalves2020generation}
\bibfield{author}{\bibinfo{person}{Andre Goncalves}, \bibinfo{person}{Priyadip
  Ray}, \bibinfo{person}{Braden Soper}, \bibinfo{person}{Jennifer Stevens},
  \bibinfo{person}{Linda Coyle}, {and} \bibinfo{person}{Ana~Paula Sales}.}
  \bibinfo{year}{2020}\natexlab{b}.
\newblock \showarticletitle{Generation and evaluation of synthetic patient
  data}.
\newblock \bibinfo{journal}{\emph{BMC medical research methodology}}
  \bibinfo{volume}{20}, \bibinfo{number}{1} (\bibinfo{year}{2020}),
  \bibinfo{pages}{1--40}.
\newblock


\bibitem[Gooch and Roudsari(2011)]%
        {gooch2011careflow}
\bibfield{author}{\bibinfo{person}{Phil Gooch} {and} \bibinfo{person}{Abdul
  Roudsari}.} \bibinfo{year}{2011}\natexlab{}.
\newblock \showarticletitle{Computerization of workflows, guidelines, and care
  pathways: a review of implementation challenges for process-oriented health
  information systems}.
\newblock \bibinfo{journal}{\emph{Journal of the American Medical Informatics
  Association}} \bibinfo{volume}{18}, \bibinfo{number}{6}
  (\bibinfo{year}{2011}), \bibinfo{pages}{738--748}.
\newblock


\bibitem[Goodfellow et~al\mbox{.}(2014)]%
        {gan2014}
\bibfield{author}{\bibinfo{person}{Ian Goodfellow}, \bibinfo{person}{Jean
  Pouget-Abadie}, \bibinfo{person}{Mehdi Mirza}, \bibinfo{person}{Bing Xu},
  \bibinfo{person}{David Warde-Farley}, \bibinfo{person}{Sherjil Ozair},
  \bibinfo{person}{Aaron Courville}, {and} \bibinfo{person}{Yoshua Bengio}.}
  \bibinfo{year}{2014}\natexlab{}.
\newblock \showarticletitle{Generative Adversarial Nets}. In
  \bibinfo{booktitle}{\emph{Advances in Neural Information Processing
  Systems}}, \bibfield{editor}{\bibinfo{person}{Z.~Ghahramani},
  \bibinfo{person}{M.~Welling}, \bibinfo{person}{C.~Cortes},
  \bibinfo{person}{N.~Lawrence}, {and} \bibinfo{person}{K.Q. Weinberger}}
  (Eds.), Vol.~\bibinfo{volume}{27}. \bibinfo{publisher}{Curran Associates,
  Inc.}, \bibinfo{address}{Montreal, Canada}.
\newblock
\urldef\tempurl%
\url{https://proceedings.neurips.cc/paper/2014/file/5ca3e9b122f61f8f06494c97b1afccf3-Paper.pdf}
\showURL{%
\tempurl}


\bibitem[Gun{\v{c}}ar et~al\mbox{.}(2018)]%
        {gunvcar2018application}
\bibfield{author}{\bibinfo{person}{Gregor Gun{\v{c}}ar},
  \bibinfo{person}{Matja{\v{z}} Kukar}, \bibinfo{person}{Mateja Notar},
  \bibinfo{person}{Miran Brvar}, \bibinfo{person}{Peter {\v{C}}ernel{\v{c}}},
  \bibinfo{person}{Manca Notar}, {and} \bibinfo{person}{Marko Notar}.}
  \bibinfo{year}{2018}\natexlab{}.
\newblock \showarticletitle{An application of machine learning to
  haematological diagnosis}.
\newblock \bibinfo{journal}{\emph{Scientific reports}} \bibinfo{volume}{8},
  \bibinfo{number}{1} (\bibinfo{year}{2018}), \bibinfo{pages}{1--12}.
\newblock


\bibitem[Han et~al\mbox{.}(2005)]%
        {Borderline-SMOTE}
\bibfield{author}{\bibinfo{person}{Hui Han}, \bibinfo{person}{Wen-Yuan Wang},
  {and} \bibinfo{person}{Bing-Huan Mao}.} \bibinfo{year}{2005}\natexlab{}.
\newblock \showarticletitle{Borderline-SMOTE: A New Over-Sampling Method in
  Imbalanced Data Sets Learning}. In \bibinfo{booktitle}{\emph{Advances in
  Intelligent Computing}}, \bibfield{editor}{\bibinfo{person}{De-Shuang Huang},
  \bibinfo{person}{Xiao-Ping Zhang}, {and} \bibinfo{person}{Guang-Bin Huang}}
  (Eds.). \bibinfo{publisher}{Springer Berlin Heidelberg},
  \bibinfo{address}{Berlin, Heidelberg}, \bibinfo{pages}{878--887}.
\newblock
\showISBNx{978-3-540-31902-3}


\bibitem[He et~al\mbox{.}(2008)]%
        {ADASYN}
\bibfield{author}{\bibinfo{person}{Haibo He}, \bibinfo{person}{Yang Bai},
  \bibinfo{person}{Edwardo~A Garcia}, {and} \bibinfo{person}{Shutao Li}.}
  \bibinfo{year}{2008}\natexlab{}.
\newblock \showarticletitle{ADASYN: Adaptive synthetic sampling approach for
  imbalanced learning}. In \bibinfo{booktitle}{\emph{2008 IEEE international
  joint conference on neural networks (IEEE world congress on computational
  intelligence)}}. \bibinfo{publisher}{IEEE}, \bibinfo{address}{Hong Kong},
  \bibinfo{pages}{1322--1328}.
\newblock
\showISBNx{1424418208}


\bibitem[He et~al\mbox{.}(2022)]%
        {AE-ELM}
\bibfield{author}{\bibinfo{person}{Yu-Lin He}, \bibinfo{person}{Sheng-Sheng
  Xu}, {and} \bibinfo{person}{Joshua~Zhexue Huang}.}
  \bibinfo{year}{2022}\natexlab{}.
\newblock \showarticletitle{Creating synthetic minority class samples based on
  autoencoder extreme learning machine}.
\newblock \bibinfo{journal}{\emph{Pattern Recognition}}  \bibinfo{volume}{121}
  (\bibinfo{year}{2022}), \bibinfo{pages}{108191}.
\newblock
\showISSN{0031-3203}


\bibitem[Hernandez et~al\mbox{.}(2022)]%
        {ehrReview3}
\bibfield{author}{\bibinfo{person}{Mikel Hernandez}, \bibinfo{person}{Gorka
  Epelde}, \bibinfo{person}{Ane Alberdi}, \bibinfo{person}{Rodrigo Cilla},
  {and} \bibinfo{person}{Debbie Rankin}.} \bibinfo{year}{2022}\natexlab{}.
\newblock \showarticletitle{Synthetic data generation for tabular health
  records: A systematic review}.
\newblock \bibinfo{journal}{\emph{Neurocomputing}}  \bibinfo{volume}{493}
  (\bibinfo{year}{2022}), \bibinfo{pages}{28--45}.
\newblock
\showISSN{0925-2312}
\urldef\tempurl%
\url{https://doi.org/10.1016/j.neucom.2022.04.053}
\showDOI{\tempurl}


\bibitem[Hinton and Roweis(2002)]%
        {tsne}
\bibfield{author}{\bibinfo{person}{Geoffrey~E Hinton} {and}
  \bibinfo{person}{Sam Roweis}.} \bibinfo{year}{2002}\natexlab{}.
\newblock \showarticletitle{Stochastic Neighbor Embedding}. In
  \bibinfo{booktitle}{\emph{Advances in Neural Information Processing
  Systems}}, \bibfield{editor}{\bibinfo{person}{S.~Becker},
  \bibinfo{person}{S.~Thrun}, {and} \bibinfo{person}{K.~Obermayer}} (Eds.),
  Vol.~\bibinfo{volume}{15}. \bibinfo{publisher}{MIT Press},
  \bibinfo{address}{Vancouver, British Columbia, Canada}.
\newblock
\urldef\tempurl%
\url{https://proceedings.neurips.cc/paper/2002/file/6150ccc6069bea6b5716254057a194ef-Paper.pdf}
\showURL{%
\tempurl}


\bibitem[Hochreiter and Schmidhuber(1997)]%
        {hochreiter1997lstm}
\bibfield{author}{\bibinfo{person}{Sepp Hochreiter} {and}
  \bibinfo{person}{J{\"u}rgen Schmidhuber}.} \bibinfo{year}{1997}\natexlab{}.
\newblock \showarticletitle{Long short-term memory}.
\newblock \bibinfo{journal}{\emph{Neural computation}} \bibinfo{volume}{9},
  \bibinfo{number}{8} (\bibinfo{year}{1997}), \bibinfo{pages}{1735--1780}.
\newblock


\bibitem[Hu et~al\mbox{.}(2022)]%
        {hu2021membership}
\bibfield{author}{\bibinfo{person}{Hongsheng Hu}, \bibinfo{person}{Zoran
  Salcic}, \bibinfo{person}{Lichao Sun}, \bibinfo{person}{Gillian Dobbie},
  \bibinfo{person}{Philip~S. Yu}, {and} \bibinfo{person}{Xuyun Zhang}.}
  \bibinfo{year}{2022}\natexlab{}.
\newblock \showarticletitle{Membership Inference Attacks on Machine Learning: A
  Survey}.
\newblock \bibinfo{journal}{\emph{ACM Comput. Surv.}}  \bibinfo{volume}{54}
  (\bibinfo{date}{jan} \bibinfo{year}{2022}), \bibinfo{pages}{1--37}.
\newblock
\showISSN{0360-0300}
\urldef\tempurl%
\url{https://doi.org/10.1145/3523273}
\showDOI{\tempurl}
\newblock
\shownote{Just Accepted}.


\bibitem[Huang et~al\mbox{.}(2013)]%
        {knowledge2013}
\bibfield{author}{\bibinfo{person}{Zhisheng Huang}, \bibinfo{person}{Frank van
  Harmelen}, \bibinfo{person}{Annette ten Teije}, {and}
  \bibinfo{person}{Kathrin Dentler}.} \bibinfo{year}{2013}\natexlab{}.
\newblock \showarticletitle{Knowledge-Based Patient Data Generation}. In
  \bibinfo{booktitle}{\emph{Process Support and Knowledge Representation in
  Health Care}}, \bibfield{editor}{\bibinfo{person}{David Ria{\~{n}}o},
  \bibinfo{person}{Richard Lenz}, \bibinfo{person}{Silvia Miksch},
  \bibinfo{person}{Mor Peleg}, \bibinfo{person}{Manfred Reichert}, {and}
  \bibinfo{person}{Annette ten Teije}} (Eds.). \bibinfo{publisher}{Springer
  International Publishing}, \bibinfo{address}{Cham}, \bibinfo{pages}{83--96}.
\newblock
\showISBNx{978-3-319-03916-9}


\bibitem[Jeong et~al\mbox{.}(2016)]%
        {jeong2016copula}
\bibfield{author}{\bibinfo{person}{Byungduk Jeong}, \bibinfo{person}{Wonjoon
  Lee}, \bibinfo{person}{Deok-Soo Kim}, {and} \bibinfo{person}{Hayong Shin}.}
  \bibinfo{year}{2016}\natexlab{}.
\newblock \showarticletitle{Copula-based approach to synthetic population
  generation}.
\newblock \bibinfo{journal}{\emph{PloS one}} \bibinfo{volume}{11},
  \bibinfo{number}{8} (\bibinfo{year}{2016}), \bibinfo{pages}{e0159496}.
\newblock


\bibitem[Jiang et~al\mbox{.}(2017)]%
        {healthcareAI}
\bibfield{author}{\bibinfo{person}{Fei Jiang}, \bibinfo{person}{Yong Jiang},
  \bibinfo{person}{Hui Zhi}, \bibinfo{person}{Yi Dong}, \bibinfo{person}{Hao
  Li}, \bibinfo{person}{Sufeng Ma}, \bibinfo{person}{Yilong Wang},
  \bibinfo{person}{Qiang Dong}, \bibinfo{person}{Haipeng Shen}, {and}
  \bibinfo{person}{Yongjun Wang}.} \bibinfo{year}{2017}\natexlab{}.
\newblock \showarticletitle{Artificial intelligence in healthcare: past,
  present and future}.
\newblock \bibinfo{journal}{\emph{Stroke and Vascular Neurology}}
  \bibinfo{volume}{2}, \bibinfo{number}{4} (\bibinfo{year}{2017}),
  \bibinfo{pages}{230--243}.
\newblock
\showISSN{2059-8688}
\urldef\tempurl%
\url{https://doi.org/10.1136/svn-2017-000101}
\showDOI{\tempurl}
\showeprint{https://svn.bmj.com/content/2/4/230.full.pdf}


\bibitem[Jing et~al\mbox{.}(2018)]%
        {medicalReport}
\bibfield{author}{\bibinfo{person}{Baoyu Jing}, \bibinfo{person}{Pengtao Xie},
  {and} \bibinfo{person}{Eric Xing}.} \bibinfo{year}{2018}\natexlab{}.
\newblock \bibinfo{title}{On the Automatic Generation of Medical Imaging
  Reports}.
\newblock
\newblock
\urldef\tempurl%
\url{https://doi.org/10.48550/arXiv.1711.08195}
\showDOI{\tempurl}


\bibitem[Jo and Kim(2022)]%
        {obgan}
\bibfield{author}{\bibinfo{person}{Wonkeun Jo} {and} \bibinfo{person}{Dongil
  Kim}.} \bibinfo{year}{2022}\natexlab{}.
\newblock \showarticletitle{OBGAN: Minority oversampling near borderline with
  generative adversarial networks}.
\newblock \bibinfo{journal}{\emph{Expert Systems with Applications}}
  \bibinfo{volume}{197} (\bibinfo{year}{2022}), \bibinfo{pages}{116694}.
\newblock
\showISSN{0957-4174}


\bibitem[Johnson et~al\mbox{.}(2022)]%
        {johnson2020mimic4}
\bibfield{author}{\bibinfo{person}{Alistair Johnson}, \bibinfo{person}{Lucas
  Bulgarelli}, \bibinfo{person}{Tom Pollard}, \bibinfo{person}{Steven Horng},
  \bibinfo{person}{Leo~Anthony Celi}, {and} \bibinfo{person}{Roger Mark}.}
  \bibinfo{year}{2022}\natexlab{}.
\newblock \bibinfo{title}{Mimic-IV}.
\newblock
\newblock
\urldef\tempurl%
\url{https://physionet.org/content/mimiciv/2.0/}
\showURL{%
\tempurl}


\bibitem[Johnson et~al\mbox{.}(2016)]%
        {johnson2016mimic3}
\bibfield{author}{\bibinfo{person}{Alistair~EW Johnson}, \bibinfo{person}{Tom~J
  Pollard}, \bibinfo{person}{Lu Shen}, \bibinfo{person}{Li-wei~H Lehman},
  \bibinfo{person}{Mengling Feng}, \bibinfo{person}{Mohammad Ghassemi},
  \bibinfo{person}{Benjamin Moody}, \bibinfo{person}{Peter Szolovits},
  \bibinfo{person}{Leo Anthony~Celi}, {and} \bibinfo{person}{Roger~G Mark}.}
  \bibinfo{year}{2016}\natexlab{}.
\newblock \showarticletitle{MIMIC-III, a freely accessible critical care
  database}.
\newblock \bibinfo{journal}{\emph{Scientific data}} \bibinfo{volume}{3},
  \bibinfo{number}{1} (\bibinfo{year}{2016}), \bibinfo{pages}{1--9}.
\newblock


\bibitem[Jordon et~al\mbox{.}(2022)]%
        {jordan2022synthetic}
\bibfield{author}{\bibinfo{person}{James Jordon}, \bibinfo{person}{Lukasz
  Szpruch}, \bibinfo{person}{Florimond Houssiau}, \bibinfo{person}{Mirko
  Bottarelli}, \bibinfo{person}{Giovanni Cherubin}, \bibinfo{person}{Carsten
  Maple}, \bibinfo{person}{Samuel~N. Cohen}, {and} \bibinfo{person}{Adrian
  Weller}.} \bibinfo{year}{2022}\natexlab{}.
\newblock \bibinfo{title}{Synthetic Data -- what, why and how?}
\newblock
\newblock
\urldef\tempurl%
\url{https://doi.org/10.48550/ARXIV.2205.03257}
\showDOI{\tempurl}


\bibitem[Jordon et~al\mbox{.}(2018)]%
        {pategan}
\bibfield{author}{\bibinfo{person}{James Jordon}, \bibinfo{person}{Jinsung
  Yoon}, {and} \bibinfo{person}{Mihaela Van Der~Schaar}.}
  \bibinfo{year}{2018}\natexlab{}.
\newblock \showarticletitle{PATE-GAN: Generating synthetic data with
  differential privacy guarantees}. In \bibinfo{booktitle}{\emph{International
  conference on learning representations}}. \bibinfo{publisher}{ICLR},
  \bibinfo{address}{New Orleans, Louisiana, USA}.
\newblock


\bibitem[Karpatne et~al\mbox{.}(2017)]%
        {karpatne2017theory}
\bibfield{author}{\bibinfo{person}{Anuj Karpatne}, \bibinfo{person}{Gowtham
  Atluri}, \bibinfo{person}{James~H Faghmous}, \bibinfo{person}{Michael
  Steinbach}, \bibinfo{person}{Arindam Banerjee}, \bibinfo{person}{Auroop
  Ganguly}, \bibinfo{person}{Shashi Shekhar}, \bibinfo{person}{Nagiza
  Samatova}, {and} \bibinfo{person}{Vipin Kumar}.}
  \bibinfo{year}{2017}\natexlab{}.
\newblock \showarticletitle{Theory-guided data science: A new paradigm for
  scientific discovery from data}.
\newblock \bibinfo{journal}{\emph{IEEE Transactions on knowledge and data
  engineering}} \bibinfo{volume}{29}, \bibinfo{number}{10}
  (\bibinfo{year}{2017}), \bibinfo{pages}{2318--2331}.
\newblock


\bibitem[Kartoun(2016)]%
        {EMRBot}
\bibfield{author}{\bibinfo{person}{Uri Kartoun}.}
  \bibinfo{year}{2016}\natexlab{}.
\newblock \bibinfo{title}{A Methodology to Generate Virtual Patient
  Repositories}.
\newblock
\newblock
\urldef\tempurl%
\url{https://doi.org/10.48550/ARXIV.1608.00570}
\showDOI{\tempurl}


\bibitem[Kaur et~al\mbox{.}(2021)]%
        {kaur2021application}
\bibfield{author}{\bibinfo{person}{Dhamanpreet Kaur}, \bibinfo{person}{Matthew
  Sobiesk}, \bibinfo{person}{Shubham Patil}, \bibinfo{person}{Jin Liu},
  \bibinfo{person}{Puran Bhagat}, \bibinfo{person}{Amar Gupta}, {and}
  \bibinfo{person}{Natasha Markuzon}.} \bibinfo{year}{2021}\natexlab{}.
\newblock \showarticletitle{Application of Bayesian networks to generate
  synthetic health data}.
\newblock \bibinfo{journal}{\emph{Journal of the American Medical Informatics
  Association}} \bibinfo{volume}{28}, \bibinfo{number}{4}
  (\bibinfo{year}{2021}), \bibinfo{pages}{801--811}.
\newblock


\bibitem[Kaur et~al\mbox{.}(2022)]%
        {trustworthyAI2}
\bibfield{author}{\bibinfo{person}{Davinder Kaur}, \bibinfo{person}{Suleyman
  Uslu}, \bibinfo{person}{Kaley~J Rittichier}, {and} \bibinfo{person}{Arjan
  Durresi}.} \bibinfo{year}{2022}\natexlab{}.
\newblock \showarticletitle{Trustworthy artificial intelligence: a review}.
\newblock \bibinfo{journal}{\emph{ACM Computing Surveys (CSUR)}}
  \bibinfo{volume}{55}, \bibinfo{number}{2} (\bibinfo{year}{2022}),
  \bibinfo{pages}{1--38}.
\newblock


\bibitem[Kingma and Welling(2013)]%
        {AE2013}
\bibfield{author}{\bibinfo{person}{Diederik~P Kingma} {and}
  \bibinfo{person}{Max Welling}.} \bibinfo{year}{2013}\natexlab{}.
\newblock \bibinfo{title}{Auto-Encoding Variational Bayes}.
\newblock
\newblock
\urldef\tempurl%
\url{https://doi.org/10.48550/ARXIV.1312.6114}
\showDOI{\tempurl}


\bibitem[Kullback and Leibler(1951)]%
        {kullback1951kldivergence}
\bibfield{author}{\bibinfo{person}{Solomon Kullback} {and}
  \bibinfo{person}{Richard~A Leibler}.} \bibinfo{year}{1951}\natexlab{}.
\newblock \showarticletitle{On information and sufficiency}.
\newblock \bibinfo{journal}{\emph{The annals of mathematical statistics}}
  \bibinfo{volume}{22}, \bibinfo{number}{1} (\bibinfo{year}{1951}),
  \bibinfo{pages}{79--86}.
\newblock


\bibitem[Kumar et~al\mbox{.}(2019)]%
        {energymodel}
\bibfield{author}{\bibinfo{person}{Rithesh Kumar}, \bibinfo{person}{Sherjil
  Ozair}, \bibinfo{person}{Anirudh Goyal}, \bibinfo{person}{Aaron Courville},
  {and} \bibinfo{person}{Yoshua Bengio}.} \bibinfo{year}{2019}\natexlab{}.
\newblock \bibinfo{title}{Maximum Entropy Generators for Energy-Based Models}.
\newblock
\newblock
\urldef\tempurl%
\url{https://doi.org/10.48550/ARXIV.1901.08508}
\showDOI{\tempurl}


\bibitem[Lam and Bacchus(1994)]%
        {MDL}
\bibfield{author}{\bibinfo{person}{Wai Lam} {and} \bibinfo{person}{Fahiem
  Bacchus}.} \bibinfo{year}{1994}\natexlab{}.
\newblock \showarticletitle{Learning Bayesian belief networks: An approach
  based on the MDL principle}.
\newblock \bibinfo{journal}{\emph{Computational intelligence}}
  \bibinfo{volume}{10}, \bibinfo{number}{3} (\bibinfo{year}{1994}),
  \bibinfo{pages}{269--293}.
\newblock
\showISSN{0824-7935}


\bibitem[Lan et~al\mbox{.}(2022)]%
        {SMOGAN}
\bibfield{author}{\bibinfo{person}{Zi-Ching Lan}, \bibinfo{person}{Guan-Yu
  Huang}, \bibinfo{person}{Yun-Pei Li}, \bibinfo{person}{Seungmin Rho},
  \bibinfo{person}{S. Vimal}, {and} \bibinfo{person}{Bo-Wei Chen}.}
  \bibinfo{year}{2022}\natexlab{}.
\newblock \showarticletitle{Conquering insufficient/imbalanced data learning
  for the Internet of Medical Things}.
\newblock \bibinfo{journal}{\emph{Neural Computing and Applications}}
  (\bibinfo{year}{2022}).
\newblock
\showISSN{1433-3058}
\urldef\tempurl%
\url{https://doi.org/10.1007/s00521-022-06897-z}
\showDOI{\tempurl}


\bibitem[Lauritzen and Spiegelhalter(1988)]%
        {lauritzen1988local}
\bibfield{author}{\bibinfo{person}{Steffen~L Lauritzen} {and}
  \bibinfo{person}{David~J Spiegelhalter}.} \bibinfo{year}{1988}\natexlab{}.
\newblock \showarticletitle{Local computations with probabilities on graphical
  structures and their application to expert systems}.
\newblock \bibinfo{journal}{\emph{Journal of the Royal Statistical Society:
  Series B (Methodological)}} \bibinfo{volume}{50}, \bibinfo{number}{2}
  (\bibinfo{year}{1988}), \bibinfo{pages}{157--194}.
\newblock


\bibitem[Lee et~al\mbox{.}(2020)]%
        {DAAE}
\bibfield{author}{\bibinfo{person}{Dongha Lee}, \bibinfo{person}{Hwanjo Yu},
  \bibinfo{person}{Xiaoqian Jiang}, \bibinfo{person}{Deevakar Rogith},
  \bibinfo{person}{Meghana Gudala}, \bibinfo{person}{Mubeen Tejani},
  \bibinfo{person}{Qiuchen Zhang}, {and} \bibinfo{person}{Li Xiong}.}
  \bibinfo{year}{2020}\natexlab{}.
\newblock \showarticletitle{Generating sequential electronic health records
  using dual adversarial autoencoder}.
\newblock \bibinfo{journal}{\emph{Journal of the American Medical Informatics
  Association}} \bibinfo{volume}{27}, \bibinfo{number}{9}
  (\bibinfo{year}{2020}), \bibinfo{pages}{1411--1419}.
\newblock
\showISSN{1067-5027}


\bibitem[Lee and Rich(2021)]%
        {humanandAI}
\bibfield{author}{\bibinfo{person}{Min~Kyung Lee} {and}
  \bibinfo{person}{Katherine Rich}.} \bibinfo{year}{2021}\natexlab{}.
\newblock \showarticletitle{Who Is Included in Human Perceptions of AI?: Trust
  and Perceived Fairness around Healthcare AI and Cultural Mistrust}. In
  \bibinfo{booktitle}{\emph{Proceedings of the 2021 CHI Conference on Human
  Factors in Computing Systems}} (Yokohama, Japan) \emph{(\bibinfo{series}{CHI
  '21})}. \bibinfo{publisher}{Association for Computing Machinery},
  \bibinfo{address}{New York, NY, USA}, Article \bibinfo{articleno}{138},
  \bibinfo{numpages}{14}~pages.
\newblock
\showISBNx{9781450380966}
\urldef\tempurl%
\url{https://doi.org/10.1145/3411764.3445570}
\showDOI{\tempurl}


\bibitem[Lema{{\^i}}tre et~al\mbox{.}(2017)]%
        {imbalancedlearn}
\bibfield{author}{\bibinfo{person}{Guillaume Lema{{\^i}}tre},
  \bibinfo{person}{Fernando Nogueira}, {and} \bibinfo{person}{Christos~K.
  Aridas}.} \bibinfo{year}{2017}\natexlab{}.
\newblock \showarticletitle{Imbalanced-learn: A Python Toolbox to Tackle the
  Curse of Imbalanced Datasets in Machine Learning}.
\newblock \bibinfo{journal}{\emph{Journal of Machine Learning Research}}
  \bibinfo{volume}{18}, \bibinfo{number}{17} (\bibinfo{year}{2017}),
  \bibinfo{pages}{1--5}.
\newblock
\urldef\tempurl%
\url{http://jmlr.org/papers/v18/16-365.html}
\showURL{%
\tempurl}


\bibitem[Li et~al\mbox{.}(2014a)]%
        {dpcopula}
\bibfield{author}{\bibinfo{person}{Haoran Li}, \bibinfo{person}{Li Xiong},
  {and} \bibinfo{person}{Xiaoqian Jiang}.} \bibinfo{year}{2014}\natexlab{a}.
\newblock \showarticletitle{Differentially private synthesization of
  multi-dimensional data using copula functions}. In
  \bibinfo{booktitle}{\emph{Advances in database technology: proceedings.
  International conference on extending database technology}},
  Vol.~\bibinfo{volume}{2014}. \bibinfo{publisher}{NIH Public Access},
  \bibinfo{address}{Bethesda, Maryland, USA}, \bibinfo{pages}{475}.
\newblock


\bibitem[Li et~al\mbox{.}(2014b)]%
        {li2014dpsynthesizer}
\bibfield{author}{\bibinfo{person}{Haoran Li}, \bibinfo{person}{Li Xiong},
  \bibinfo{person}{Lifan Zhang}, {and} \bibinfo{person}{Xiaoqian Jiang}.}
  \bibinfo{year}{2014}\natexlab{b}.
\newblock \showarticletitle{DPSynthesizer: Differentially private data
  synthesizer for privacy preserving data sharing}. In
  \bibinfo{booktitle}{\emph{Proceedings of the VLDB Endowment International
  Conference on Very Large Data Bases}}, Vol.~\bibinfo{volume}{7}. NIH Public
  Access, \bibinfo{publisher}{NIH Public Access}, \bibinfo{address}{Hangzhou,
  China}, \bibinfo{pages}{1677}.
\newblock


\bibitem[Liang et~al\mbox{.}(2022)]%
        {liang2022advances}
\bibfield{author}{\bibinfo{person}{Weixin Liang}, \bibinfo{person}{Girmaw~Abebe
  Tadesse}, \bibinfo{person}{Daniel Ho}, \bibinfo{person}{Fei-Fei Li},
  \bibinfo{person}{Matei Zaharia}, \bibinfo{person}{Ce Zhang}, {and}
  \bibinfo{person}{James Zou}.} \bibinfo{year}{2022}\natexlab{}.
\newblock \showarticletitle{Advances, challenges and opportunities in creating
  data for trustworthy AI}.
\newblock \bibinfo{journal}{\emph{Nature Machine Intelligence}}
  \bibinfo{volume}{4} (\bibinfo{year}{2022}), \bibinfo{pages}{669–677}.
\newblock


\bibitem[Lindquist et~al\mbox{.}(2007)]%
        {lindquist2007modeling}
\bibfield{author}{\bibinfo{person}{Martin~A Lindquist},
  \bibinfo{person}{Christian Waugh}, {and} \bibinfo{person}{Tor~D Wager}.}
  \bibinfo{year}{2007}\natexlab{}.
\newblock \showarticletitle{Modeling state-related fMRI activity using
  change-point theory}.
\newblock \bibinfo{journal}{\emph{NeuroImage}} \bibinfo{volume}{35},
  \bibinfo{number}{3} (\bibinfo{year}{2007}), \bibinfo{pages}{1125--1141}.
\newblock


\bibitem[Litvak et~al\mbox{.}(2011)]%
        {litvak2011eeg}
\bibfield{author}{\bibinfo{person}{Vladimir Litvak},
  \bibinfo{person}{J{\'e}r{\'e}mie Mattout}, \bibinfo{person}{Stefan Kiebel},
  \bibinfo{person}{Christophe Phillips}, \bibinfo{person}{Richard Henson},
  \bibinfo{person}{James Kilner}, \bibinfo{person}{Gareth Barnes},
  \bibinfo{person}{Robert Oostenveld}, \bibinfo{person}{Jean Daunizeau},
  \bibinfo{person}{Guillaume Flandin}, {et~al\mbox{.}}}
  \bibinfo{year}{2011}\natexlab{}.
\newblock \showarticletitle{EEG and MEG data analysis in SPM8}.
\newblock \bibinfo{journal}{\emph{Computational intelligence and neuroscience}}
   \bibinfo{volume}{2011} (\bibinfo{year}{2011}), \bibinfo{pages}{852961}.
\newblock


\bibitem[Luo et~al\mbox{.}(2021)]%
        {luo2021oversampling}
\bibfield{author}{\bibinfo{person}{Hao Luo}, \bibinfo{person}{Jun Liao},
  \bibinfo{person}{Xuewen Yan}, {and} \bibinfo{person}{Li LiU}.}
  \bibinfo{year}{2021}\natexlab{}.
\newblock \showarticletitle{Oversampling by a Constraint-Based Causal Network
  in Medical Imbalanced Data Classification}. In \bibinfo{booktitle}{\emph{2021
  IEEE International Conference on Multimedia and Expo (ICME)}}. IEEE,
  \bibinfo{publisher}{IEEE}, \bibinfo{address}{Taipei, Taiwan},
  \bibinfo{pages}{1--6}.
\newblock


\bibitem[Machanavajjhala et~al\mbox{.}(2008)]%
        {machanavajjhala2008privacy}
\bibfield{author}{\bibinfo{person}{Ashwin Machanavajjhala},
  \bibinfo{person}{Daniel Kifer}, \bibinfo{person}{John Abowd},
  \bibinfo{person}{Johannes Gehrke}, {and} \bibinfo{person}{Lars Vilhuber}.}
  \bibinfo{year}{2008}\natexlab{}.
\newblock \showarticletitle{Privacy: Theory meets practice on the map}. In
  \bibinfo{booktitle}{\emph{2008 IEEE 24th international conference on data
  engineering}}. IEEE, \bibinfo{publisher}{IEEE}, \bibinfo{address}{Dallas, TX,
  USA}, \bibinfo{pages}{277--286}.
\newblock


\bibitem[Massey~Jr(1951)]%
        {massey1951kolmogorov}
\bibfield{author}{\bibinfo{person}{Frank~J Massey~Jr}.}
  \bibinfo{year}{1951}\natexlab{}.
\newblock \showarticletitle{The Kolmogorov-Smirnov test for goodness of fit}.
\newblock \bibinfo{journal}{\emph{Journal of the American statistical
  Association}} \bibinfo{volume}{46}, \bibinfo{number}{253}
  (\bibinfo{year}{1951}), \bibinfo{pages}{68--78}.
\newblock


\bibitem[Matwin et~al\mbox{.}(2015)]%
        {attributeReview}
\bibfield{author}{\bibinfo{person}{Stan Matwin}, \bibinfo{person}{Jordi Nin},
  \bibinfo{person}{Morvarid Sehatkar}, {and} \bibinfo{person}{Tomasz Szapiro}.}
  \bibinfo{year}{2015}\natexlab{}.
\newblock \bibinfo{booktitle}{\emph{A Review of Attribute Disclosure Control}}.
\newblock \bibinfo{publisher}{Springer International Publishing},
  \bibinfo{address}{Cham}, \bibinfo{pages}{41--61}.
\newblock
\showISBNx{978-3-319-09885-2}
\urldef\tempurl%
\url{https://doi.org/10.1007/978-3-319-09885-2_4}
\showDOI{\tempurl}


\bibitem[McLachlan et~al\mbox{.}(2016)]%
        {CorMESR}
\bibfield{author}{\bibinfo{person}{Scott McLachlan},
  \bibinfo{person}{Kudakwashe Dube}, {and} \bibinfo{person}{Thomas Gallagher}.}
  \bibinfo{year}{2016}\natexlab{}.
\newblock \showarticletitle{Using the caremap with health incidents statistics
  for generating the realistic synthetic electronic healthcare record}. In
  \bibinfo{booktitle}{\emph{2016 IEEE International Conference on Healthcare
  Informatics (ICHI)}}. IEEE, \bibinfo{publisher}{IEEE},
  \bibinfo{address}{Chicago, Illinois, USA}, \bibinfo{pages}{439--448}.
\newblock


\bibitem[McSherry and Talwar(2007)]%
        {expdp}
\bibfield{author}{\bibinfo{person}{Frank McSherry} {and} \bibinfo{person}{Kunal
  Talwar}.} \bibinfo{year}{2007}\natexlab{}.
\newblock \showarticletitle{Mechanism design via differential privacy}. In
  \bibinfo{booktitle}{\emph{48th Annual IEEE Symposium on Foundations of
  Computer Science (FOCS'07)}}. IEEE, \bibinfo{publisher}{IEEE},
  \bibinfo{address}{Providence, Rhode Island, USA}, \bibinfo{pages}{94--103}.
\newblock


\bibitem[Moniz et~al\mbox{.}(2009)]%
        {emerge2009}
\bibfield{author}{\bibinfo{person}{Linda Moniz}, \bibinfo{person}{Anna~L
  Buczak}, \bibinfo{person}{Lang Hung}, \bibinfo{person}{Steven Babin},
  \bibinfo{person}{Michael Dorko}, {and} \bibinfo{person}{Joseph Lombardo}.}
  \bibinfo{year}{2009}\natexlab{}.
\newblock \showarticletitle{Construction and validation of synthetic electronic
  medical records}.
\newblock \bibinfo{journal}{\emph{Online journal of public health informatics}}
  \bibinfo{volume}{1}, \bibinfo{number}{1} (\bibinfo{year}{2009}),
  \bibinfo{pages}{1--36}.
\newblock


\bibitem[Moody and Mark(1996)]%
        {moody1996mimic1}
\bibfield{author}{\bibinfo{person}{George~B Moody} {and}
  \bibinfo{person}{Roger~G Mark}.} \bibinfo{year}{1996}\natexlab{}.
\newblock \showarticletitle{A database to support development and evaluation of
  intelligent intensive care monitoring}. In
  \bibinfo{booktitle}{\emph{Computers in Cardiology 1996}}. IEEE,
  \bibinfo{publisher}{IEEE}, \bibinfo{address}{Indianapolis,USA},
  \bibinfo{pages}{657--660}.
\newblock


\bibitem[Muralidhar et~al\mbox{.}(1999)]%
        {GADP}
\bibfield{author}{\bibinfo{person}{Krishnamurty Muralidhar},
  \bibinfo{person}{Rahul Parsa}, {and} \bibinfo{person}{Rathindra Sarathy}.}
  \bibinfo{year}{1999}\natexlab{}.
\newblock \showarticletitle{A general additive data perturbation method for
  database security}.
\newblock \bibinfo{journal}{\emph{management science}} \bibinfo{volume}{45},
  \bibinfo{number}{10} (\bibinfo{year}{1999}), \bibinfo{pages}{1399--1415}.
\newblock
\showISSN{0025-1909}


\bibitem[Müller et~al\mbox{.}(2001)]%
        {fmrianalysisSpectral}
\bibfield{author}{\bibinfo{person}{Karsten Müller}, \bibinfo{person}{Gabriele
  Lohmann}, \bibinfo{person}{Volker Bosch}, {and} \bibinfo{person}{D.Yves {von
  Cramon}}.} \bibinfo{year}{2001}\natexlab{}.
\newblock \showarticletitle{On Multivariate Spectral Analysis of fMRI Time
  Series}.
\newblock \bibinfo{journal}{\emph{NeuroImage}} \bibinfo{volume}{14},
  \bibinfo{number}{2} (\bibinfo{year}{2001}), \bibinfo{pages}{347--356}.
\newblock
\showISSN{1053-8119}
\urldef\tempurl%
\url{https://doi.org/10.1006/nimg.2001.0804}
\showDOI{\tempurl}


\bibitem[Neubig(2017)]%
        {seq2seq}
\bibfield{author}{\bibinfo{person}{Graham Neubig}.}
  \bibinfo{year}{2017}\natexlab{}.
\newblock \bibinfo{title}{Neural Machine Translation and Sequence-to-sequence
  Models: A Tutorial}.
\newblock
\newblock
\urldef\tempurl%
\url{https://doi.org/10.48550/ARXIV.1703.01619}
\showDOI{\tempurl}


\bibitem[Neves et~al\mbox{.}(2022)]%
        {gainwithtricks}
\bibfield{author}{\bibinfo{person}{Diogo~Telmo Neves}, \bibinfo{person}{João
  Alves}, \bibinfo{person}{Marcel~Ganesh Naik}, \bibinfo{person}{Alberto~José
  Proença}, {and} \bibinfo{person}{Fabian Prasser}.}
  \bibinfo{year}{2022}\natexlab{}.
\newblock \showarticletitle{From Missing Data Imputation to Data Generation}.
\newblock \bibinfo{journal}{\emph{Journal of Computational Science}}
  \bibinfo{volume}{61} (\bibinfo{year}{2022}), \bibinfo{pages}{101640}.
\newblock
\showISSN{1877-7503}


\bibitem[Nightingale and Farid(2022)]%
        {nightingale2022ai}
\bibfield{author}{\bibinfo{person}{Sophie~J Nightingale} {and}
  \bibinfo{person}{Hany Farid}.} \bibinfo{year}{2022}\natexlab{}.
\newblock \showarticletitle{AI-synthesized faces are indistinguishable from
  real faces and more trustworthy}.
\newblock \bibinfo{journal}{\emph{Proceedings of the National Academy of
  Sciences}} \bibinfo{volume}{119}, \bibinfo{number}{8} (\bibinfo{year}{2022}),
  \bibinfo{pages}{e2120481119}.
\newblock


\bibitem[Nuwer(1988)]%
        {nuwer1988quantitative}
\bibfield{author}{\bibinfo{person}{Marc~R Nuwer}.}
  \bibinfo{year}{1988}\natexlab{}.
\newblock \showarticletitle{Quantitative EEG: I. Techniques and problems of
  frequency analysis and topographic mapping.}
\newblock \bibinfo{journal}{\emph{Journal of clinical neurophysiology: official
  publication of the American Electroencephalographic Society}}
  \bibinfo{volume}{5}, \bibinfo{number}{1} (\bibinfo{year}{1988}),
  \bibinfo{pages}{1--43}.
\newblock


\bibitem[Odena et~al\mbox{.}(2017)]%
        {odena2017acgan}
\bibfield{author}{\bibinfo{person}{Augustus Odena},
  \bibinfo{person}{Christopher Olah}, {and} \bibinfo{person}{Jonathon Shlens}.}
  \bibinfo{year}{2017}\natexlab{}.
\newblock \showarticletitle{Conditional Image Synthesis with Auxiliary
  Classifier {GAN}s}. In \bibinfo{booktitle}{\emph{Proceedings of the 34th
  International Conference on Machine Learning}}
  \emph{(\bibinfo{series}{Proceedings of Machine Learning Research},
  Vol.~\bibinfo{volume}{70})}, \bibfield{editor}{\bibinfo{person}{Doina Precup}
  {and} \bibinfo{person}{Yee~Whye Teh}} (Eds.). \bibinfo{publisher}{PMLR},
  \bibinfo{address}{Sydney, Australia}, \bibinfo{pages}{2642--2651}.
\newblock
\urldef\tempurl%
\url{https://proceedings.mlr.press/v70/odena17a.html}
\showURL{%
\tempurl}


\bibitem[of~Health \& Human~Services(1996)]%
        {hipaa}
\bibfield{author}{\bibinfo{person}{U.S.~Department of Health \&
  Human~Services}.} \bibinfo{year}{1996}\natexlab{}.
\newblock \bibinfo{title}{Health Insurance Portability and Accountability Act
  (HIPAA)}.
\newblock
\newblock
\urldef\tempurl%
\url{https://www.hhs.gov/hipaa/index.html}
\showURL{%
\tempurl}


\bibitem[of~the National Coordinator~for Health Information
  Technology~(ONC)(2019)]%
        {EHRofficial1}
\bibfield{author}{\bibinfo{person}{The~Office of~the National Coordinator~for
  Health Information Technology~(ONC)}.} \bibinfo{year}{2019}\natexlab{}.
\newblock \bibinfo{title}{What is an electronic healthcare record?}
\newblock
\newblock
\urldef\tempurl%
\url{https://www.healthit.gov/faq/what-electronic-health-record-ehr}
\showURL{%
\tempurl}


\bibitem[Organization(2020)]%
        {DE-SynPUF}
\bibfield{author}{\bibinfo{person}{Redivis~Demo Organization}.}
  \bibinfo{year}{2020}\natexlab{}.
\newblock \bibinfo{title}{CMS Synthetic Patient Data OMOP}.
\newblock
\newblock
\urldef\tempurl%
\url{https://redivis.com/datasets/ye2v-6skh7wdr7?v=2.0}
\showURL{%
\tempurl}


\bibitem[Organization(2004)]%
        {ICD10}
\bibfield{author}{\bibinfo{person}{World~Health Organization}.}
  \bibinfo{year}{2004}\natexlab{}.
\newblock \bibinfo{booktitle}{\emph{International Statistical Classification of
  Diseases and related health problems: Alphabetical index}}.
  Vol.~\bibinfo{volume}{3}.
\newblock \bibinfo{publisher}{World Health Organization},
  \bibinfo{address}{USA}.
\newblock


\bibitem[Pardey et~al\mbox{.}(1996)]%
        {pardey1996review}
\bibfield{author}{\bibinfo{person}{James Pardey}, \bibinfo{person}{Stephen
  Roberts}, {and} \bibinfo{person}{Lionel Tarassenko}.}
  \bibinfo{year}{1996}\natexlab{}.
\newblock \showarticletitle{A review of parametric modelling techniques for EEG
  analysis}.
\newblock \bibinfo{journal}{\emph{Medical engineering \& physics}}
  \bibinfo{volume}{18}, \bibinfo{number}{1} (\bibinfo{year}{1996}),
  \bibinfo{pages}{2--11}.
\newblock


\bibitem[Park et~al\mbox{.}(2018)]%
        {table-gan}
\bibfield{author}{\bibinfo{person}{Noseong Park}, \bibinfo{person}{Mahmoud
  Mohammadi}, \bibinfo{person}{Kshitij Gorde}, \bibinfo{person}{Sushil
  Jajodia}, \bibinfo{person}{Hongkyu Park}, {and} \bibinfo{person}{Youngmin
  Kim}.} \bibinfo{year}{2018}\natexlab{}.
\newblock \showarticletitle{Data synthesis based on generative adversarial
  networks}.
\newblock \bibinfo{journal}{\emph{Proceedings of the {VLDB} Endowment}}
  \bibinfo{volume}{11}, \bibinfo{number}{10} (\bibinfo{date}{jun}
  \bibinfo{year}{2018}), \bibinfo{pages}{1071--1083}.
\newblock
\urldef\tempurl%
\url{https://doi.org/10.14778/3231751.3231757}
\showDOI{\tempurl}


\bibitem[Park and Ghosh(2013)]%
        {PGS2013}
\bibfield{author}{\bibinfo{person}{Yubin Park} {and} \bibinfo{person}{Joydeep
  Ghosh}.} \bibinfo{year}{2013}\natexlab{}.
\newblock \bibinfo{title}{Perturbed Gibbs Samplers for Synthetic Data Release}.
\newblock
\newblock
\urldef\tempurl%
\url{https://doi.org/10.48550/ARXIV.1312.5370}
\showDOI{\tempurl}


\bibitem[Park et~al\mbox{.}(2013)]%
        {PGSapplication}
\bibfield{author}{\bibinfo{person}{Yubin Park}, \bibinfo{person}{Joydeep
  Ghosh}, {and} \bibinfo{person}{Mallikarjun Shankar}.}
  \bibinfo{year}{2013}\natexlab{}.
\newblock \showarticletitle{Perturbed gibbs samplers for generating large-scale
  privacy-safe synthetic health data}. In \bibinfo{booktitle}{\emph{2013 IEEE
  International Conference on Healthcare Informatics}}.
  \bibinfo{publisher}{IEEE}, \bibinfo{address}{Philadelphia, Pennsylvania,
  USA}, \bibinfo{pages}{493--498}.
\newblock
\showISBNx{0769550894}


\bibitem[Patki et~al\mbox{.}(2016)]%
        {patki2016syntheticdatavault}
\bibfield{author}{\bibinfo{person}{Neha Patki}, \bibinfo{person}{Roy Wedge},
  {and} \bibinfo{person}{Kalyan Veeramachaneni}.}
  \bibinfo{year}{2016}\natexlab{}.
\newblock \showarticletitle{The synthetic data vault}. In
  \bibinfo{booktitle}{\emph{2016 IEEE International Conference on Data Science
  and Advanced Analytics (DSAA)}}. IEEE, \bibinfo{publisher}{IEEE},
  \bibinfo{address}{Montreal, Quebec, Canada}, \bibinfo{pages}{399--410}.
\newblock


\bibitem[Pessoa et~al\mbox{.}(2002)]%
        {pessoa2002neural}
\bibfield{author}{\bibinfo{person}{Luiz Pessoa}, \bibinfo{person}{Eva
  Gutierrez}, \bibinfo{person}{Peter~A Bandettini}, {and}
  \bibinfo{person}{Leslie~G Ungerleider}.} \bibinfo{year}{2002}\natexlab{}.
\newblock \showarticletitle{Neural correlates of visual working memory: fMRI
  amplitude predicts task performance}.
\newblock \bibinfo{journal}{\emph{Neuron}} \bibinfo{volume}{35},
  \bibinfo{number}{5} (\bibinfo{year}{2002}), \bibinfo{pages}{975--987}.
\newblock


\bibitem[Ping et~al\mbox{.}(2017)]%
        {ping2017datasynthesizer}
\bibfield{author}{\bibinfo{person}{Haoyue Ping}, \bibinfo{person}{Julia
  Stoyanovich}, {and} \bibinfo{person}{Bill Howe}.}
  \bibinfo{year}{2017}\natexlab{}.
\newblock \showarticletitle{DataSynthesizer: Privacy-Preserving Synthetic
  Datasets}. In \bibinfo{booktitle}{\emph{Proceedings of the 29th International
  Conference on Scientific and Statistical Database Management}} (Chicago, IL,
  USA) \emph{(\bibinfo{series}{SSDBM '17})}. \bibinfo{publisher}{Association
  for Computing Machinery}, \bibinfo{address}{New York, NY, USA}, Article
  \bibinfo{articleno}{42}, \bibinfo{numpages}{5}~pages.
\newblock
\showISBNx{9781450352826}
\urldef\tempurl%
\url{https://doi.org/10.1145/3085504.3091117}
\showDOI{\tempurl}


\bibitem[Podnar et~al\mbox{.}(2019)]%
        {podnar2019diagnosing}
\bibfield{author}{\bibinfo{person}{Simon Podnar}, \bibinfo{person}{Matja{\v{z}}
  Kukar}, \bibinfo{person}{Gregor Gun{\v{c}}ar}, \bibinfo{person}{Mateja
  Notar}, \bibinfo{person}{Nina Go{\v{s}}njak}, {and} \bibinfo{person}{Marko
  Notar}.} \bibinfo{year}{2019}\natexlab{}.
\newblock \showarticletitle{Diagnosing brain tumours by routine blood tests
  using machine learning}.
\newblock \bibinfo{journal}{\emph{Scientific reports}} \bibinfo{volume}{9},
  \bibinfo{number}{1} (\bibinfo{year}{2019}), \bibinfo{pages}{1--7}.
\newblock


\bibitem[Polat(2019)]%
        {smotepd}
\bibfield{author}{\bibinfo{person}{Kemal Polat}.}
  \bibinfo{year}{2019}\natexlab{}.
\newblock \showarticletitle{A hybrid approach to Parkinson disease
  classification using speech signal: the combination of smote and random
  forests}. In \bibinfo{booktitle}{\emph{2019 Scientific Meeting on
  Electrical-Electronics \& Biomedical Engineering and Computer Science
  (EBBT)}}. Ieee, \bibinfo{publisher}{IEEE}, \bibinfo{address}{Istanbul,
  Turkey}, \bibinfo{pages}{1--3}.
\newblock


\bibitem[Pollard(2005)]%
        {pollard2005total}
\bibfield{author}{\bibinfo{person}{David Pollard}.}
  \bibinfo{year}{2005}\natexlab{}.
\newblock \showarticletitle{Total variation distance between measures}.
\newblock \bibinfo{journal}{\emph{Asymptopia}} \bibinfo{volume}{1},
  \bibinfo{number}{chap. 3} (\bibinfo{year}{2005}), \bibinfo{pages}{1}.
\newblock


\bibitem[Pollard et~al\mbox{.}(2018)]%
        {pollard2019icu}
\bibfield{author}{\bibinfo{person}{Tom~J Pollard}, \bibinfo{person}{Alistair~EW
  Johnson}, \bibinfo{person}{Jesse~D Raffa}, \bibinfo{person}{Leo~A Celi},
  \bibinfo{person}{Roger~G Mark}, {and} \bibinfo{person}{Omar Badawi}.}
  \bibinfo{year}{2018}\natexlab{}.
\newblock \showarticletitle{The eICU Collaborative Research Database, a freely
  available multi-center database for critical care research}.
\newblock \bibinfo{journal}{\emph{Scientific data}} \bibinfo{volume}{5},
  \bibinfo{number}{1} (\bibinfo{year}{2018}), \bibinfo{pages}{1--13}.
\newblock


\bibitem[Pytorch(2020)]%
        {RNN}
\bibfield{author}{\bibinfo{person}{Pytorch}.} \bibinfo{year}{2020}\natexlab{}.
\newblock \bibinfo{title}{RNN Pytorch 1.12 document}.
\newblock
\newblock
\urldef\tempurl%
\url{https://pytorch.org/docs/stable/generated/torch.nn.RNN.html}
\showURL{%
Retrieved Aug 19, 2022 from \tempurl}


\bibitem[Raissi et~al\mbox{.}(2019)]%
        {raissi2019physics}
\bibfield{author}{\bibinfo{person}{Maziar Raissi}, \bibinfo{person}{Paris
  Perdikaris}, {and} \bibinfo{person}{George~E Karniadakis}.}
  \bibinfo{year}{2019}\natexlab{}.
\newblock \showarticletitle{Physics-informed neural networks: A deep learning
  framework for solving forward and inverse problems involving nonlinear
  partial differential equations}.
\newblock \bibinfo{journal}{\emph{Journal of Computational physics}}
  \bibinfo{volume}{378} (\bibinfo{year}{2019}), \bibinfo{pages}{686--707}.
\newblock


\bibitem[Rashidian et~al\mbox{.}(2020)]%
        {rashidian2020smooth}
\bibfield{author}{\bibinfo{person}{Sina Rashidian}, \bibinfo{person}{Fusheng
  Wang}, \bibinfo{person}{Richard Moffitt}, \bibinfo{person}{Victor Garcia},
  \bibinfo{person}{Anurag Dutt}, \bibinfo{person}{Wei Chang},
  \bibinfo{person}{Vishwam Pandya}, \bibinfo{person}{Janos Hajagos},
  \bibinfo{person}{Mary Saltz}, {and} \bibinfo{person}{Joel Saltz}.}
  \bibinfo{year}{2020}\natexlab{}.
\newblock \showarticletitle{SMOOTH-GAN: Towards Sharp and Smooth Synthetic EHR
  Data Generation}. In \bibinfo{booktitle}{\emph{Artificial Intelligence in
  Medicine}}, \bibfield{editor}{\bibinfo{person}{Martin Michalowski} {and}
  \bibinfo{person}{Robert Moskovitch}} (Eds.). \bibinfo{publisher}{Springer
  International Publishing}, \bibinfo{address}{Cham}, \bibinfo{pages}{37--48}.
\newblock
\showISBNx{978-3-030-59137-3}


\bibitem[Ria{\~{n}}o and Fern{\'a}ndez-P{\'e}rez(2017)]%
        {knowledge2016}
\bibfield{author}{\bibinfo{person}{David Ria{\~{n}}o} {and}
  \bibinfo{person}{Alberto Fern{\'a}ndez-P{\'e}rez}.}
  \bibinfo{year}{2017}\natexlab{}.
\newblock \showarticletitle{Simulation-Based Episodes of Care Data
  Synthetization for Chronic Disease Patients}. In
  \bibinfo{booktitle}{\emph{Knowledge Representation for Health Care}},
  \bibfield{editor}{\bibinfo{person}{David Ria{\~{n}}o},
  \bibinfo{person}{Richard Lenz}, {and} \bibinfo{person}{Manfred Reichert}}
  (Eds.). \bibinfo{publisher}{Springer International Publishing},
  \bibinfo{address}{Cham}, \bibinfo{pages}{36--50}.
\newblock
\showISBNx{978-3-319-55014-5}


\bibitem[Rubin(1993)]%
        {rubin93}
\bibfield{author}{\bibinfo{person}{Donald~B Rubin}.}
  \bibinfo{year}{1993}\natexlab{}.
\newblock \showarticletitle{Statistical disclosure limitation}.
\newblock \bibinfo{journal}{\emph{Journal of official Statistics}}
  \bibinfo{volume}{9}, \bibinfo{number}{2} (\bibinfo{year}{1993}),
  \bibinfo{pages}{461--468}.
\newblock


\bibitem[Rumelhart et~al\mbox{.}(1986)]%
        {rumelhart1986rnn}
\bibfield{author}{\bibinfo{person}{David~E Rumelhart},
  \bibinfo{person}{Geoffrey~E Hinton}, {and} \bibinfo{person}{Ronald~J
  Williams}.} \bibinfo{year}{1986}\natexlab{}.
\newblock \showarticletitle{Learning representations by back-propagating
  errors}.
\newblock \bibinfo{journal}{\emph{nature}} \bibinfo{volume}{323},
  \bibinfo{number}{6088} (\bibinfo{year}{1986}), \bibinfo{pages}{533--536}.
\newblock


\bibitem[Saeed et~al\mbox{.}(2002)]%
        {saeed2002mimic2}
\bibfield{author}{\bibinfo{person}{Mohammed Saeed}, \bibinfo{person}{Christine
  Lieu}, \bibinfo{person}{Greg Raber}, {and} \bibinfo{person}{Roger~G Mark}.}
  \bibinfo{year}{2002}\natexlab{}.
\newblock \showarticletitle{MIMIC II: a massive temporal ICU patient database
  to support research in intelligent patient monitoring}. In
  \bibinfo{booktitle}{\emph{Computers in cardiology}}. IEEE,
  \bibinfo{publisher}{IEEE}, \bibinfo{address}{Memphis, Tennessee, USA},
  \bibinfo{pages}{641--644}.
\newblock


\bibitem[Schreiber(2017)]%
        {schreiber2017pomegranate}
\bibfield{author}{\bibinfo{person}{Jacob Schreiber}.}
  \bibinfo{year}{2017}\natexlab{}.
\newblock \showarticletitle{Pomegranate: fast and flexible probabilistic
  modeling in python}.
\newblock \bibinfo{journal}{\emph{The Journal of Machine Learning Research}}
  \bibinfo{volume}{18}, \bibinfo{number}{1} (\bibinfo{year}{2017}),
  \bibinfo{pages}{5992--5997}.
\newblock


\bibitem[Schwarz(1978)]%
        {BIC}
\bibfield{author}{\bibinfo{person}{Gideon Schwarz}.}
  \bibinfo{year}{1978}\natexlab{}.
\newblock \showarticletitle{Estimating the dimension of a model}.
\newblock \bibinfo{journal}{\emph{The annals of statistics}}
  \bibinfo{volume}{6}, \bibinfo{number}{2} (\bibinfo{year}{1978}),
  \bibinfo{pages}{461--464}.
\newblock
\showISSN{0090-5364}


\bibitem[Service(2018)]%
        {bloodtestsNHS}
\bibfield{author}{\bibinfo{person}{National~Health Service}.}
  \bibinfo{year}{2018}\natexlab{}.
\newblock \bibinfo{title}{Blood tests-Example}.
\newblock
\newblock
\urldef\tempurl%
\url{https://www.nhs.uk/conditions/blood-tests/types/}
\showURL{%
\tempurl}


\bibitem[Shokri et~al\mbox{.}(2017)]%
        {shokri2017membership}
\bibfield{author}{\bibinfo{person}{Reza Shokri}, \bibinfo{person}{Marco
  Stronati}, \bibinfo{person}{Congzheng Song}, {and} \bibinfo{person}{Vitaly
  Shmatikov}.} \bibinfo{year}{2017}\natexlab{}.
\newblock \showarticletitle{Membership inference attacks against machine
  learning models}. In \bibinfo{booktitle}{\emph{2017 IEEE symposium on
  security and privacy (SP)}}. IEEE, \bibinfo{publisher}{IEEE},
  \bibinfo{address}{San Jose, California, USA}, \bibinfo{pages}{3--18}.
\newblock


\bibitem[Sikstrom et~al\mbox{.}(2022)]%
        {medicalfairness}
\bibfield{author}{\bibinfo{person}{Laura Sikstrom}, \bibinfo{person}{Marta~M
  Maslej}, \bibinfo{person}{Katrina Hui}, \bibinfo{person}{Zoe Findlay},
  \bibinfo{person}{Daniel~Z Buchman}, {and} \bibinfo{person}{Sean~L Hill}.}
  \bibinfo{year}{2022}\natexlab{}.
\newblock \showarticletitle{Conceptualising fairness: three pillars for medical
  algorithms and health equity}.
\newblock \bibinfo{journal}{\emph{BMJ Health \& Care Informatics}}
  \bibinfo{volume}{29}, \bibinfo{number}{1} (\bibinfo{year}{2022}),
  \bibinfo{pages}{e100459}.
\newblock
\urldef\tempurl%
\url{https://doi.org/10.1136/bmjhci-2021-100459}
\showDOI{\tempurl}
\showeprint{https://informatics.bmj.com/content/29/1/e100459.full.pdf}


\bibitem[Sklar(1959)]%
        {sklar1959fonctions}
\bibfield{author}{\bibinfo{person}{M Sklar}.} \bibinfo{year}{1959}\natexlab{}.
\newblock \showarticletitle{Fonctions de repartition an dimensions et leurs
  marges}.
\newblock \bibinfo{journal}{\emph{Publ. inst. statist. univ. Paris}}
  \bibinfo{volume}{8} (\bibinfo{year}{1959}), \bibinfo{pages}{229--231}.
\newblock


\bibitem[Smith et~al\mbox{.}(2013)]%
        {smith2013identifying}
\bibfield{author}{\bibinfo{person}{Jason~F Smith}, \bibinfo{person}{Kewei
  Chen}, \bibinfo{person}{Ajay~S Pillai}, {and} \bibinfo{person}{Barry
  Horwitz}.} \bibinfo{year}{2013}\natexlab{}.
\newblock \showarticletitle{Identifying effective connectivity parameters in
  simulated fMRI: a direct comparison of switching linear dynamic system,
  stochastic dynamic causal, and multivariate autoregressive models}.
\newblock \bibinfo{journal}{\emph{Frontiers in neuroscience}}
  \bibinfo{volume}{7} (\bibinfo{year}{2013}), \bibinfo{pages}{70}.
\newblock


\bibitem[Smith(2004)]%
        {fmrianalysisTemporal}
\bibfield{author}{\bibinfo{person}{S~M Smith}.}
  \bibinfo{year}{2004}\natexlab{}.
\newblock \showarticletitle{Overview of fMRI analysis}.
\newblock \bibinfo{journal}{\emph{The British Journal of Radiology}}
  \bibinfo{volume}{77}, \bibinfo{number}{suppl\_2} (\bibinfo{year}{2004}),
  \bibinfo{pages}{S167--S175}.
\newblock
\urldef\tempurl%
\url{https://doi.org/10.1259/bjr/33553595}
\showDOI{\tempurl}
\showeprint{https://doi.org/10.1259/bjr/33553595}
\newblock
\shownote{PMID: 15677358}.


\bibitem[Son et~al\mbox{.}(2020)]%
        {bcgan}
\bibfield{author}{\bibinfo{person}{Minjae Son}, \bibinfo{person}{Seungwon
  Jung}, \bibinfo{person}{Jihoon Moon}, {and} \bibinfo{person}{Eenjun Hwang}.}
  \bibinfo{year}{2020}\natexlab{}.
\newblock \showarticletitle{BCGAN-based over-sampling scheme for imbalanced
  data}. In \bibinfo{booktitle}{\emph{2020 IEEE International Conference on Big
  Data and Smart Computing (BigComp)}}. \bibinfo{publisher}{IEEE},
  \bibinfo{address}{Busan, Korea}, \bibinfo{pages}{155--160}.
\newblock
\showISBNx{1728160340}


\bibitem[Song et~al\mbox{.}(2019)]%
        {adasynad}
\bibfield{author}{\bibinfo{person}{Tzu-An Song}, \bibinfo{person}{Samadrita~Roy
  Chowdhury}, \bibinfo{person}{Fan Yang}, \bibinfo{person}{Heidi Jacobs},
  \bibinfo{person}{Georges El~Fakhri}, \bibinfo{person}{Quanzheng Li},
  \bibinfo{person}{Keith Johnson}, {and} \bibinfo{person}{Joyita Dutta}.}
  \bibinfo{year}{2019}\natexlab{}.
\newblock \showarticletitle{Graph convolutional neural networks for
  Alzheimer’s disease classification}. In \bibinfo{booktitle}{\emph{2019 IEEE
  16th International Symposium on Biomedical Imaging (ISBI 2019)}}.
  \bibinfo{publisher}{IEEE}, \bibinfo{address}{Venice, Italy},
  \bibinfo{pages}{414--417}.
\newblock


\bibitem[Srivastava et~al\mbox{.}(2017)]%
        {srivastava2017veegan}
\bibfield{author}{\bibinfo{person}{Akash Srivastava}, \bibinfo{person}{Lazar
  Valkov}, \bibinfo{person}{Chris Russell}, \bibinfo{person}{Michael~U.
  Gutmann}, {and} \bibinfo{person}{Charles Sutton}.}
  \bibinfo{year}{2017}\natexlab{}.
\newblock \showarticletitle{VEEGAN: Reducing Mode Collapse in GANs using
  Implicit Variational Learning}. In \bibinfo{booktitle}{\emph{Advances in
  Neural Information Processing Systems}},
  \bibfield{editor}{\bibinfo{person}{I.~Guyon}, \bibinfo{person}{U.~Von
  Luxburg}, \bibinfo{person}{S.~Bengio}, \bibinfo{person}{H.~Wallach},
  \bibinfo{person}{R.~Fergus}, \bibinfo{person}{S.~Vishwanathan}, {and}
  \bibinfo{person}{R.~Garnett}} (Eds.), Vol.~\bibinfo{volume}{30}.
  \bibinfo{publisher}{Curran Associates, Inc.}, \bibinfo{address}{Long Beach,
  California, USA}.
\newblock
\urldef\tempurl%
\url{https://proceedings.neurips.cc/paper/2017/file/44a2e0804995faf8d2e3b084a1e2db1d-Paper.pdf}
\showURL{%
\tempurl}


\bibitem[Stadler et~al\mbox{.}(2022)]%
        {stadler2021synthetic}
\bibfield{author}{\bibinfo{person}{Theresa Stadler}, \bibinfo{person}{Bristena
  Oprisanu}, {and} \bibinfo{person}{Carmela Troncoso}.}
  \bibinfo{year}{2022}\natexlab{}.
\newblock \showarticletitle{Synthetic Data {\textendash} Anonymisation
  Groundhog Day}. In \bibinfo{booktitle}{\emph{31st USENIX Security Symposium
  (USENIX Security 22)}}. \bibinfo{publisher}{USENIX Association},
  \bibinfo{address}{Boston, MA}, \bibinfo{pages}{1451--1468}.
\newblock
\showISBNx{978-1-939133-31-1}
\urldef\tempurl%
\url{https://www.usenix.org/conference/usenixsecurity22/presentation/stadler}
\showURL{%
\tempurl}


\bibitem[Sun and Erath(2015)]%
        {sun2015bayesian}
\bibfield{author}{\bibinfo{person}{Lijun Sun} {and} \bibinfo{person}{Alexander
  Erath}.} \bibinfo{year}{2015}\natexlab{}.
\newblock \showarticletitle{A Bayesian network approach for population
  synthesis}.
\newblock \bibinfo{journal}{\emph{Transportation Research Part C: Emerging
  Technologies}}  \bibinfo{volume}{61} (\bibinfo{year}{2015}),
  \bibinfo{pages}{49--62}.
\newblock


\bibitem[Sun et~al\mbox{.}(2021)]%
        {longgan}
\bibfield{author}{\bibinfo{person}{Siao Sun}, \bibinfo{person}{Fusheng Wang},
  \bibinfo{person}{Sina Rashidian}, \bibinfo{person}{Tahsin Kurc},
  \bibinfo{person}{Kayley Abell-Hart}, \bibinfo{person}{Janos Hajagos},
  \bibinfo{person}{Wei Zhu}, \bibinfo{person}{Mary Saltz}, {and}
  \bibinfo{person}{Joel Saltz}.} \bibinfo{year}{2021}\natexlab{}.
\newblock \showarticletitle{Generating Longitudinal Synthetic EHR Data with
  Recurrent Autoencoders and Generative Adversarial Networks}. In
  \bibinfo{booktitle}{\emph{Heterogeneous Data Management, Polystores, and
  Analytics for Healthcare}}. \bibinfo{publisher}{Springer International
  Publishing}, \bibinfo{address}{Cham}, \bibinfo{pages}{153--165}.
\newblock
\showISBNx{978-3-030-93663-1}


\bibitem[Swinscow et~al\mbox{.}(2002)]%
        {swinscow2002statistics}
\bibfield{author}{\bibinfo{person}{Thomas Douglas~Victor Swinscow},
  \bibinfo{person}{Michael~J Campbell}, {et~al\mbox{.}}}
  \bibinfo{year}{2002}\natexlab{}.
\newblock \bibinfo{booktitle}{\emph{Statistics at square one}}.
\newblock \bibinfo{publisher}{Bmj London}, \bibinfo{address}{London, UK}.
\newblock


\bibitem[Taskesen(2020)]%
        {bnlearn}
\bibfield{author}{\bibinfo{person}{Erdogan Taskesen}.}
  \bibinfo{year}{2020}\natexlab{}.
\newblock \bibinfo{title}{bnlearn - Library for Bayesian network learning and
  inference}.
\newblock
\newblock
\urldef\tempurl%
\url{https://erdogant.github.io/bnlearn}
\showURL{%
\tempurl}


\bibitem[Thoral et~al\mbox{.}(2021)]%
        {thoral2021umcdb}
\bibfield{author}{\bibinfo{person}{Patrick~J Thoral}, \bibinfo{person}{Jan~M
  Peppink}, \bibinfo{person}{Ronald~H Driessen}, \bibinfo{person}{Eric~JG
  Sijbrands}, \bibinfo{person}{Erwin~JO Kompanje}, \bibinfo{person}{Lewis
  Kaplan}, \bibinfo{person}{Heatherlee Bailey}, \bibinfo{person}{Jozef
  Kesecioglu}, \bibinfo{person}{Maurizio Cecconi}, \bibinfo{person}{Matthew
  Churpek}, {et~al\mbox{.}}} \bibinfo{year}{2021}\natexlab{}.
\newblock \showarticletitle{Sharing ICU patient data Responsibly under the
  Society of critical care Medicine/European Society of intensive care medicine
  joint data science collaboration: the Amsterdam University medical centers
  database (AmsterdamUMCdb) example}.
\newblock \bibinfo{journal}{\emph{Critical care medicine}}
  \bibinfo{volume}{49}, \bibinfo{number}{6} (\bibinfo{year}{2021}),
  \bibinfo{pages}{e563}.
\newblock


\bibitem[Torfi and Fox(2020)]%
        {CorGAN}
\bibfield{author}{\bibinfo{person}{Amirsina Torfi} {and}
  \bibinfo{person}{Edward~A Fox}.} \bibinfo{year}{2020}\natexlab{}.
\newblock \showarticletitle{CorGAN: Correlation-capturing convolutional
  generative adversarial networks for generating synthetic healthcare records}.
  In \bibinfo{booktitle}{\emph{The Thirty-Third International Flairs
  Conference}}. \bibinfo{publisher}{AAAI Press}, \bibinfo{address}{Florida,
  USA}, \bibinfo{pages}{1--6}.
\newblock


\bibitem[Tucker et~al\mbox{.}(2020)]%
        {tucker2020generating}
\bibfield{author}{\bibinfo{person}{Allan Tucker}, \bibinfo{person}{Zhenchen
  Wang}, \bibinfo{person}{Ylenia Rotalinti}, {and} \bibinfo{person}{Puja
  Myles}.} \bibinfo{year}{2020}\natexlab{}.
\newblock \showarticletitle{Generating high-fidelity synthetic patient data for
  assessing machine learning healthcare software}.
\newblock \bibinfo{journal}{\emph{NPJ digital medicine}} \bibinfo{volume}{3},
  \bibinfo{number}{1} (\bibinfo{year}{2020}), \bibinfo{pages}{1--13}.
\newblock


\bibitem[uth.edu(2022)]%
        {UTPdata}
\bibfield{author}{\bibinfo{person}{uth.edu}.} \bibinfo{year}{2022}\natexlab{}.
\newblock \bibinfo{title}{BIG-Arc - Clinical Data Warehouse - Data Dashboard}.
\newblock
\newblock
\urldef\tempurl%
\url{https://big.uth.edu/bigarc/}
\showURL{%
Retrieved Aug 19, 2022 from \tempurl}


\bibitem[Vardhan and Kok(2020)]%
        {OVAE}
\bibfield{author}{\bibinfo{person}{L~Vivek~Harsha Vardhan} {and}
  \bibinfo{person}{Stanley Kok}.} \bibinfo{year}{2020}\natexlab{}.
\newblock \showarticletitle{Generating privacy-preserving synthetic tabular
  data using oblivious variational autoencoders}. In
  \bibinfo{booktitle}{\emph{Proceedings of the Workshop on Economics of Privacy
  and Data Labor at the 37 th International Conference on Machine Learning}}.
  \bibinfo{publisher}{PMLR}, \bibinfo{address}{Vienna, Austria},
  \bibinfo{pages}{1--8}.
\newblock


\bibitem[Vaswani et~al\mbox{.}(2017)]%
        {vaswani2017transformer}
\bibfield{author}{\bibinfo{person}{Ashish Vaswani}, \bibinfo{person}{Noam
  Shazeer}, \bibinfo{person}{Niki Parmar}, \bibinfo{person}{Jakob Uszkoreit},
  \bibinfo{person}{Llion Jones}, \bibinfo{person}{Aidan~N Gomez},
  \bibinfo{person}{\L~ukasz Kaiser}, {and} \bibinfo{person}{Illia Polosukhin}.}
  \bibinfo{year}{2017}\natexlab{}.
\newblock \showarticletitle{Attention is All you Need}. In
  \bibinfo{booktitle}{\emph{Advances in Neural Information Processing
  Systems}}, \bibfield{editor}{\bibinfo{person}{I.~Guyon},
  \bibinfo{person}{U.~Von Luxburg}, \bibinfo{person}{S.~Bengio},
  \bibinfo{person}{H.~Wallach}, \bibinfo{person}{R.~Fergus},
  \bibinfo{person}{S.~Vishwanathan}, {and} \bibinfo{person}{R.~Garnett}}
  (Eds.), Vol.~\bibinfo{volume}{30}. \bibinfo{publisher}{Curran Associates,
  Inc.}, \bibinfo{address}{Long Beach, California, United States}.
\newblock
\urldef\tempurl%
\url{https://proceedings.neurips.cc/paper/2017/file/3f5ee243547dee91fbd053c1c4a845aa-Paper.pdf}
\showURL{%
\tempurl}


\bibitem[Vincent et~al\mbox{.}(2008)]%
        {imputationAE2008}
\bibfield{author}{\bibinfo{person}{Pascal Vincent}, \bibinfo{person}{Hugo
  Larochelle}, \bibinfo{person}{Yoshua Bengio}, {and}
  \bibinfo{person}{Pierre-Antoine Manzagol}.} \bibinfo{year}{2008}\natexlab{}.
\newblock \showarticletitle{Extracting and composing robust features with
  denoising autoencoders}. In \bibinfo{booktitle}{\emph{Proceedings of the 25th
  international conference on Machine learning}}. \bibinfo{publisher}{PMLR},
  \bibinfo{address}{Valencia, Spain}, \bibinfo{pages}{1096--1103}.
\newblock


\bibitem[Voigt and Von~dem Bussche(2017)]%
        {voigt2017gdpr}
\bibfield{author}{\bibinfo{person}{Paul Voigt} {and} \bibinfo{person}{Axel
  Von~dem Bussche}.} \bibinfo{year}{2017}\natexlab{}.
\newblock \showarticletitle{The eu general data protection regulation (gdpr)}.
\newblock \bibinfo{journal}{\emph{A Practical Guide, 1st Ed., Cham: Springer
  International Publishing}} \bibinfo{volume}{10}, \bibinfo{number}{3152676}
  (\bibinfo{year}{2017}), \bibinfo{pages}{10--5555}.
\newblock


\bibitem[Walck(2007)]%
        {distributions}
\bibfield{author}{\bibinfo{person}{Christian Walck}.}
  \bibinfo{year}{2007}\natexlab{}.
\newblock \showarticletitle{Hand-book on statistical distributions for
  experimentalists}.
\newblock \bibinfo{journal}{\emph{University of Stockholm}}
  \bibinfo{volume}{10} (\bibinfo{year}{2007}), \bibinfo{pages}{96--01}.
\newblock


\bibitem[Walonoski et~al\mbox{.}(2018)]%
        {walonoski2018synthea}
\bibfield{author}{\bibinfo{person}{Jason Walonoski}, \bibinfo{person}{Mark
  Kramer}, \bibinfo{person}{Joseph Nichols}, \bibinfo{person}{Andre Quina},
  \bibinfo{person}{Chris Moesel}, \bibinfo{person}{Dylan Hall},
  \bibinfo{person}{Carlton Duffett}, \bibinfo{person}{Kudakwashe Dube},
  \bibinfo{person}{Thomas Gallagher}, {and} \bibinfo{person}{Scott McLachlan}.}
  \bibinfo{year}{2018}\natexlab{}.
\newblock \showarticletitle{Synthea: An approach, method, and software
  mechanism for generating synthetic patients and the synthetic electronic
  health care record}.
\newblock \bibinfo{journal}{\emph{Journal of the American Medical Informatics
  Association}} \bibinfo{volume}{25}, \bibinfo{number}{3}
  (\bibinfo{year}{2018}), \bibinfo{pages}{230--238}.
\newblock


\bibitem[Wang et~al\mbox{.}(2019)]%
        {wang2019scGAN}
\bibfield{author}{\bibinfo{person}{Lu Wang}, \bibinfo{person}{Wei Zhang}, {and}
  \bibinfo{person}{Xiaofeng He}.} \bibinfo{year}{2019}\natexlab{}.
\newblock \showarticletitle{Continuous Patient-Centric Sequence Generation via
  Sequentially Coupled Adversarial Learning}. In
  \bibinfo{booktitle}{\emph{Database Systems for Advanced Applications}},
  \bibfield{editor}{\bibinfo{person}{Guoliang Li}, \bibinfo{person}{Jun Yang},
  \bibinfo{person}{Joao Gama}, \bibinfo{person}{Juggapong Natwichai}, {and}
  \bibinfo{person}{Yongxin Tong}} (Eds.). \bibinfo{publisher}{Springer
  International Publishing}, \bibinfo{address}{Cham}, \bibinfo{pages}{36--52}.
\newblock
\showISBNx{978-3-030-18579-4}


\bibitem[Wang et~al\mbox{.}(2021)]%
        {ehrReview1}
\bibfield{author}{\bibinfo{person}{Zhenchen Wang}, \bibinfo{person}{Puja
  Myles}, {and} \bibinfo{person}{Allan Tucker}.}
  \bibinfo{year}{2021}\natexlab{}.
\newblock \showarticletitle{Generating and evaluating cross-sectional synthetic
  electronic healthcare data: Preserving data utility and patient privacy}.
\newblock \bibinfo{journal}{\emph{Computational Intelligence}}
  \bibinfo{volume}{37}, \bibinfo{number}{2} (\bibinfo{year}{2021}),
  \bibinfo{pages}{819--851}.
\newblock


\bibitem[Welvaert and Rosseel(2014)]%
        {fmrisynthesisreview}
\bibfield{author}{\bibinfo{person}{Marijke Welvaert} {and}
  \bibinfo{person}{Yves Rosseel}.} \bibinfo{year}{2014}\natexlab{}.
\newblock \showarticletitle{A review of fMRI simulation studies}.
\newblock \bibinfo{journal}{\emph{PloS one}} \bibinfo{volume}{9},
  \bibinfo{number}{7} (\bibinfo{year}{2014}), \bibinfo{pages}{e101953}.
\newblock
\showISSN{1932-6203}


\bibitem[Wing(2021)]%
        {trustworthyAI}
\bibfield{author}{\bibinfo{person}{Jeannette~M. Wing}.}
  \bibinfo{year}{2021}\natexlab{}.
\newblock \showarticletitle{Trustworthy AI}.
\newblock \bibinfo{journal}{\emph{Commun. ACM}} \bibinfo{volume}{64},
  \bibinfo{number}{10} (\bibinfo{date}{sep} \bibinfo{year}{2021}),
  \bibinfo{pages}{64–71}.
\newblock
\showISSN{0001-0782}
\urldef\tempurl%
\url{https://doi.org/10.1145/3448248}
\showDOI{\tempurl}


\bibitem[Wulff and Jeppesen(2017)]%
        {wulff2017multiple}
\bibfield{author}{\bibinfo{person}{Jesper~N Wulff} {and}
  \bibinfo{person}{Linda~Ejlskov Jeppesen}.} \bibinfo{year}{2017}\natexlab{}.
\newblock \showarticletitle{Multiple imputation by chained equations in praxis:
  guidelines and review}.
\newblock \bibinfo{journal}{\emph{Electronic Journal of Business Research
  Methods}} \bibinfo{volume}{15}, \bibinfo{number}{1} (\bibinfo{year}{2017}),
  \bibinfo{pages}{41--56}.
\newblock


\bibitem[Xie et~al\mbox{.}(2018)]%
        {dpgan}
\bibfield{author}{\bibinfo{person}{Liyang Xie}, \bibinfo{person}{Kaixiang Lin},
  \bibinfo{person}{Shu Wang}, \bibinfo{person}{Fei Wang}, {and}
  \bibinfo{person}{Jiayu Zhou}.} \bibinfo{year}{2018}\natexlab{}.
\newblock \bibinfo{title}{Differentially Private Generative Adversarial
  Network}.
\newblock
\newblock
\urldef\tempurl%
\url{https://doi.org/10.48550/ARXIV.1802.06739}
\showDOI{\tempurl}


\bibitem[Xu et~al\mbox{.}(2018)]%
        {fairgan}
\bibfield{author}{\bibinfo{person}{Depeng Xu}, \bibinfo{person}{Shuhan Yuan},
  \bibinfo{person}{Lu Zhang}, {and} \bibinfo{person}{Xintao Wu}.}
  \bibinfo{year}{2018}\natexlab{}.
\newblock \showarticletitle{Fairgan: Fairness-aware generative adversarial
  networks}. In \bibinfo{booktitle}{\emph{2018 IEEE International Conference on
  Big Data (Big Data)}}. \bibinfo{publisher}{IEEE}, \bibinfo{address}{Seattle,
  Washington, USA}, \bibinfo{pages}{570--575}.
\newblock
\showISBNx{1538650355}


\bibitem[Xu et~al\mbox{.}(2019b)]%
        {fairganP}
\bibfield{author}{\bibinfo{person}{Depeng Xu}, \bibinfo{person}{Shuhan Yuan},
  \bibinfo{person}{Lu Zhang}, {and} \bibinfo{person}{Xintao Wu}.}
  \bibinfo{year}{2019}\natexlab{b}.
\newblock \showarticletitle{Fairgan+: Achieving fair data generation and
  classification through generative adversarial nets}. In
  \bibinfo{booktitle}{\emph{2019 IEEE International Conference on Big Data (Big
  Data)}}. \bibinfo{publisher}{IEEE}, \bibinfo{address}{Los Angeles,
  California, USA}, \bibinfo{pages}{1401--1406}.
\newblock
\showISBNx{1728108586}


\bibitem[Xu et~al\mbox{.}(2019a)]%
        {CTGAN}
\bibfield{author}{\bibinfo{person}{Lei Xu}, \bibinfo{person}{Maria
  Skoularidou}, \bibinfo{person}{Alfredo Cuesta-Infante}, {and}
  \bibinfo{person}{Kalyan Veeramachaneni}.} \bibinfo{year}{2019}\natexlab{a}.
\newblock \showarticletitle{Modeling Tabular data using Conditional GAN}. In
  \bibinfo{booktitle}{\emph{Advances in Neural Information Processing
  Systems}}, \bibfield{editor}{\bibinfo{person}{H.~Wallach},
  \bibinfo{person}{H.~Larochelle}, \bibinfo{person}{A.~Beygelzimer},
  \bibinfo{person}{F.~d\textquotesingle Alch\'{e}-Buc},
  \bibinfo{person}{E.~Fox}, {and} \bibinfo{person}{R.~Garnett}} (Eds.),
  Vol.~\bibinfo{volume}{32}. \bibinfo{publisher}{Curran Associates, Inc.},
  \bibinfo{address}{Vancouver, British Columbia, Canada}.
\newblock
\urldef\tempurl%
\url{https://proceedings.neurips.cc/paper/2019/file/254ed7d2de3b23ab10936522dd547b78-Paper.pdf}
\showURL{%
\tempurl}


\bibitem[Xu and Veeramachaneni(2018)]%
        {TGAN}
\bibfield{author}{\bibinfo{person}{Lei Xu} {and} \bibinfo{person}{Kalyan
  Veeramachaneni}.} \bibinfo{year}{2018}\natexlab{}.
\newblock \bibinfo{title}{Synthesizing Tabular Data using Generative
  Adversarial Networks}.
\newblock
\newblock
\urldef\tempurl%
\url{https://doi.org/10.48550/ARXIV.1811.11264}
\showDOI{\tempurl}


\bibitem[Yale et~al\mbox{.}(2020)]%
        {dash2019healthGAN}
\bibfield{author}{\bibinfo{person}{Andrew Yale}, \bibinfo{person}{Saloni Dash},
  \bibinfo{person}{Ritik Dutta}, \bibinfo{person}{Isabelle Guyon},
  \bibinfo{person}{Adrien Pavao}, {and} \bibinfo{person}{Kristin~P Bennett}.}
  \bibinfo{year}{2020}\natexlab{}.
\newblock \showarticletitle{Generation and evaluation of privacy preserving
  synthetic health data}.
\newblock \bibinfo{journal}{\emph{Neurocomputing}}  \bibinfo{volume}{416}
  (\bibinfo{year}{2020}), \bibinfo{pages}{244--255}.
\newblock


\bibitem[Yan et~al\mbox{.}(2020)]%
        {heterogeneousGAN}
\bibfield{author}{\bibinfo{person}{Chao Yan}, \bibinfo{person}{Ziqi Zhang},
  \bibinfo{person}{Steve Nyemba}, {and} \bibinfo{person}{Bradley~A Malin}.}
  \bibinfo{year}{2020}\natexlab{}.
\newblock \showarticletitle{Generating electronic health records with multiple
  data types and constraints}. In \bibinfo{booktitle}{\emph{AMIA annual
  symposium proceedings}}, Vol.~\bibinfo{volume}{2020}. American Medical
  Informatics Association, \bibinfo{publisher}{American Medical Informatics
  Association}, \bibinfo{address}{USA}, \bibinfo{pages}{1335}.
\newblock


\bibitem[Yang and Qian(2021)]%
        {yang2021gan}
\bibfield{author}{\bibinfo{person}{Huan Yang} {and} \bibinfo{person}{Pengjiang
  Qian}.} \bibinfo{year}{2021}\natexlab{}.
\newblock \showarticletitle{GAN-based medical images synthesis: a review}.
\newblock \bibinfo{journal}{\emph{International Journal of Health Systems and
  Translational Medicine (IJHSTM)}} \bibinfo{volume}{1}, \bibinfo{number}{2}
  (\bibinfo{year}{2021}), \bibinfo{pages}{1--9}.
\newblock


\bibitem[Yoon et~al\mbox{.}(2020)]%
        {yoon2020adsgan}
\bibfield{author}{\bibinfo{person}{Jinsung Yoon}, \bibinfo{person}{Lydia~N
  Drumright}, {and} \bibinfo{person}{Mihaela Van Der~Schaar}.}
  \bibinfo{year}{2020}\natexlab{}.
\newblock \showarticletitle{Anonymization through data synthesis using
  generative adversarial networks (ads-gan)}.
\newblock \bibinfo{journal}{\emph{IEEE journal of biomedical and health
  informatics}} \bibinfo{volume}{24}, \bibinfo{number}{8}
  (\bibinfo{year}{2020}), \bibinfo{pages}{2378--2388}.
\newblock


\bibitem[Yoon et~al\mbox{.}(2019)]%
        {timegan}
\bibfield{author}{\bibinfo{person}{Jinsung Yoon}, \bibinfo{person}{Daniel
  Jarrett}, {and} \bibinfo{person}{Mihaela van~der Schaar}.}
  \bibinfo{year}{2019}\natexlab{}.
\newblock \showarticletitle{Time-series Generative Adversarial Networks}. In
  \bibinfo{booktitle}{\emph{Advances in Neural Information Processing
  Systems}}, \bibfield{editor}{\bibinfo{person}{H.~Wallach},
  \bibinfo{person}{H.~Larochelle}, \bibinfo{person}{A.~Beygelzimer},
  \bibinfo{person}{F.~d\textquotesingle Alch\'{e}-Buc},
  \bibinfo{person}{E.~Fox}, {and} \bibinfo{person}{R.~Garnett}} (Eds.),
  Vol.~\bibinfo{volume}{32}. \bibinfo{publisher}{Curran Associates, Inc.},
  \bibinfo{address}{Vancouver, British Columbia, Canada}.
\newblock
\urldef\tempurl%
\url{https://proceedings.neurips.cc/paper/2019/file/c9efe5f26cd17ba6216bbe2a7d26d490-Paper.pdf}
\showURL{%
\tempurl}


\bibitem[Yoon et~al\mbox{.}(2018)]%
        {gain}
\bibfield{author}{\bibinfo{person}{Jinsung Yoon}, \bibinfo{person}{James
  Jordon}, {and} \bibinfo{person}{Mihaela van~der Schaar}.}
  \bibinfo{year}{2018}\natexlab{}.
\newblock \showarticletitle{{GAIN}: Missing Data Imputation using Generative
  Adversarial Nets}. In \bibinfo{booktitle}{\emph{Proceedings of the 35th
  International Conference on Machine Learning}}
  \emph{(\bibinfo{series}{Proceedings of Machine Learning Research},
  Vol.~\bibinfo{volume}{80})}, \bibfield{editor}{\bibinfo{person}{Jennifer Dy}
  {and} \bibinfo{person}{Andreas Krause}} (Eds.). \bibinfo{publisher}{PMLR},
  \bibinfo{address}{Stockholm, Sweden}, \bibinfo{pages}{5689--5698}.
\newblock
\urldef\tempurl%
\url{https://proceedings.mlr.press/v80/yoon18a.html}
\showURL{%
\tempurl}


\bibitem[Zhai et~al\mbox{.}(2016)]%
        {aeapproximation}
\bibfield{author}{\bibinfo{person}{Shuangfei Zhai}, \bibinfo{person}{Yu Cheng},
  \bibinfo{person}{Rogerio Feris}, {and} \bibinfo{person}{Zhongfei Zhang}.}
  \bibinfo{year}{2016}\natexlab{}.
\newblock \bibinfo{title}{Generative Adversarial Networks as Variational
  Training of Energy Based Models}.
\newblock
\newblock
\urldef\tempurl%
\url{https://doi.org/10.48550/ARXIV.1611.01799}
\showDOI{\tempurl}


\bibitem[Zhan et~al\mbox{.}(2021)]%
        {multimodal}
\bibfield{author}{\bibinfo{person}{Fangneng Zhan}, \bibinfo{person}{Yingchen
  Yu}, \bibinfo{person}{Rongliang Wu}, \bibinfo{person}{Jiahui Zhang},
  \bibinfo{person}{Shijian Lu}, \bibinfo{person}{Lingjie Liu},
  \bibinfo{person}{Adam Kortylewski}, \bibinfo{person}{Christian Theobalt},
  {and} \bibinfo{person}{Eric Xing}.} \bibinfo{year}{2021}\natexlab{}.
\newblock \bibinfo{title}{Multimodal Image Synthesis and Editing: A Survey}.
\newblock
\newblock
\urldef\tempurl%
\url{https://doi.org/10.48550/ARXIV.2112.13592}
\showDOI{\tempurl}


\bibitem[Zhang et~al\mbox{.}(2017)]%
        {Privbayes}
\bibfield{author}{\bibinfo{person}{Jun Zhang}, \bibinfo{person}{Graham
  Cormode}, \bibinfo{person}{Cecilia~M Procopiuc}, \bibinfo{person}{Divesh
  Srivastava}, {and} \bibinfo{person}{Xiaokui Xiao}.}
  \bibinfo{year}{2017}\natexlab{}.
\newblock \showarticletitle{Privbayes: Private data release via bayesian
  networks}.
\newblock \bibinfo{journal}{\emph{ACM Transactions on Database Systems (TODS)}}
  \bibinfo{volume}{42}, \bibinfo{number}{4} (\bibinfo{year}{2017}),
  \bibinfo{pages}{1--41}.
\newblock
\showISSN{0362-5915}


\bibitem[Zhang et~al\mbox{.}(2020)]%
        {zhang2021privsyn}
\bibfield{author}{\bibinfo{person}{Zhikun Zhang}, \bibinfo{person}{Tianhao
  Wang}, \bibinfo{person}{Ninghui Li}, \bibinfo{person}{Jean Honorio},
  \bibinfo{person}{Michael Backes}, \bibinfo{person}{Shibo He},
  \bibinfo{person}{Jiming Chen}, {and} \bibinfo{person}{Yang Zhang}.}
  \bibinfo{year}{2020}\natexlab{}.
\newblock \bibinfo{title}{PrivSyn: Differentially Private Data Synthesis}.
\newblock
\newblock
\urldef\tempurl%
\url{https://doi.org/10.48550/ARXIV.2012.15128}
\showDOI{\tempurl}


\bibitem[Zhang et~al\mbox{.}(2021)]%
        {SynTEG}
\bibfield{author}{\bibinfo{person}{Ziqi Zhang}, \bibinfo{person}{Chao Yan},
  \bibinfo{person}{Thomas~A Lasko}, \bibinfo{person}{Jimeng Sun}, {and}
  \bibinfo{person}{Bradley~A Malin}.} \bibinfo{year}{2021}\natexlab{}.
\newblock \showarticletitle{SynTEG: a framework for temporal structured
  electronic health data simulation}.
\newblock \bibinfo{journal}{\emph{Journal of the American Medical Informatics
  Association}} \bibinfo{volume}{28}, \bibinfo{number}{3}
  (\bibinfo{year}{2021}), \bibinfo{pages}{596--604}.
\newblock
\showISSN{1527-974X}


\bibitem[Zhang et~al\mbox{.}(2019)]%
        {EMRGAN}
\bibfield{author}{\bibinfo{person}{Ziqi Zhang}, \bibinfo{person}{Chao Yan},
  \bibinfo{person}{Diego~A Mesa}, \bibinfo{person}{Jimeng Sun}, {and}
  \bibinfo{person}{Bradley~A Malin}.} \bibinfo{year}{2019}\natexlab{}.
\newblock \showarticletitle{{Ensuring electronic medical record simulation
  through better training, modeling, and evaluation}}.
\newblock \bibinfo{journal}{\emph{Journal of the American Medical Informatics
  Association}} \bibinfo{volume}{27}, \bibinfo{number}{1} (\bibinfo{date}{10}
  \bibinfo{year}{2019}), \bibinfo{pages}{99--108}.
\newblock
\showISSN{1527-974X}
\urldef\tempurl%
\url{https://doi.org/10.1093/jamia/ocz161}
\showDOI{\tempurl}
\showeprint{https://academic.oup.com/jamia/article-pdf/27/1/99/34152076/ocz161.pdf}


\end{thebibliography}







\end{document}